\pdfoutput=1

\documentclass[10pt,twocolumn,twoside]{IEEEtran}

\usepackage{caption}

\usepackage{ifpdf}

\usepackage{cite}

\ifCLASSINFOpdf
  \usepackage[pdftex]{graphicx}
\else
  \usepackage[dvips]{graphicx}
\fi
\usepackage{amsmath}
\usepackage{amssymb}
\usepackage{algorithm}
\usepackage{algorithmic}
\ifCLASSOPTIONcompsoc
 \usepackage[caption=false,font=normalsize,labelfont=sf,textfont=sf]{subfig}
\else
 \usepackage[caption=false,font=footnotesize]{subfig}
\fi
\usepackage{fixltx2e}

\usepackage{stfloats}
\usepackage{url}

\begin{document}
%
\title{Lifelong Learning with Searchable Extension Units}

%

\author{Wenjin~Wang,
        Yunqing~Hu,
        and~Yin~Zhang*%
    \thanks{This work was supported by the NSFC (No. 61402403), Artificial Intelligence Research Foundation of Baidu Inc., Chinese Knowledge Center for Engineering Sciences and Technology, the Fundamental Research Funds for the Central Universities, and Engineering Research Center of Digital Library, Ministry of Education. (Corresponding author: Yin Zhang)}
    \thanks{W. Wang, Y. Hu, and Y. Zhang are with the College of Computer Science and Technology, Zhejiang University, Hangzhou 310027, China (e-mail: wangwenjin@zju.edu.cn; 11921141@zju.edu.cn; zhangyin98@zju.edu.cn).} 
}

%
%

\ifCLASSOPTIONpeerreview
    \markboth{Journal of \LaTeX\ Class Files,~Vol.~14, No.~8, August~2015}%
    {Shell \MakeLowercase{\textit{et al.}}: Bare Demo of IEEEtran.cls for IEEE Journals}
\fi
%



\maketitle

\begin{abstract}
    Lifelong learning remains an open problem. One of its main difficulties is catastrophic forgetting. Many dynamic expansion approaches have been proposed to address this problem, but they all use homogeneous models of predefined structure for all tasks. The common original model and expansion structures ignore the requirement of different model structures on different tasks, which leads to a less compact model for multiple tasks and causes the model size to increase rapidly as the number of tasks increases. Moreover, they can not perform best on all tasks. To solve those problems, in this paper, we propose a new lifelong learning framework named Searchable Extension Units (SEU) by introducing Neural Architecture Search into lifelong learning, which breaks down the need for a predefined original model and searches for specific extension units for different tasks, without compromising the performance of the model on different tasks. Our approach can obtain a much more compact model without catastrophic forgetting. The experimental results on the PMNIST, the split CIFAR10 dataset, the split CIFAR100 dataset, and the Mixture dataset empirically prove that our method can achieve higher accuracy with much smaller model, whose size is about 25-33 percentage of that of the state-of-the-art methods. 
\end{abstract}


%

\section{Introduction}
\IEEEPARstart{L}{ifelong} learning is an intrinsic ability of humans \cite{parisi_continual_2019, lifelong}. It is natural for human beings to constantly learn new knowledge and accumulate it. Moreover, humans can also effectively use past knowledge to help them learn new knowledge. For an agent in a changing environment, the lifelong learning ability is very meaningful. But in the field of deep learning, despite recent advances \cite{ResNet, AlexNet_2012}, lifelong learning remains an open question. A major problem with lifelong learning is catastrophic forgetting \cite{french_catastrophic_1999}. When learning a series of consecutive tasks, the knowledge learned from previous tasks is stored in weights of the deep neural network. Catastrophic forgetting means that the weights of the model are changed when learning a new task so that the knowledge of previous tasks will be forgotten \cite{french_catastrophic_1999}.

To alleviate the catastrophic forgetting, many methods have been proposed. Depending on whether the model size changes, these methods fall into two broad categories: \emph{fix methods} whose model structure is fixed, \emph{expansion methods} whose model structure is variable.

\emph{Fix methods} maintain the performance of model by retaining data information about previous tasks and the values of model parameters\cite{rebuffi_icarl:_2017, lopez-paz_gradient_2017,deep_generative_replay,incremental_classifier_learning_2018,adaption_2019,ewc, Synaptic_Intelligence,memory_aware_synapses,pathnet:_2017,hard_attention,adaption_2019}. In fact, due to limited capacity, methods with a fixed model structure are more likely to encounter a capacity bottleneck in continuous learning scenarios.

\begin{figure}
    \centering
    \includegraphics[width=1\linewidth]{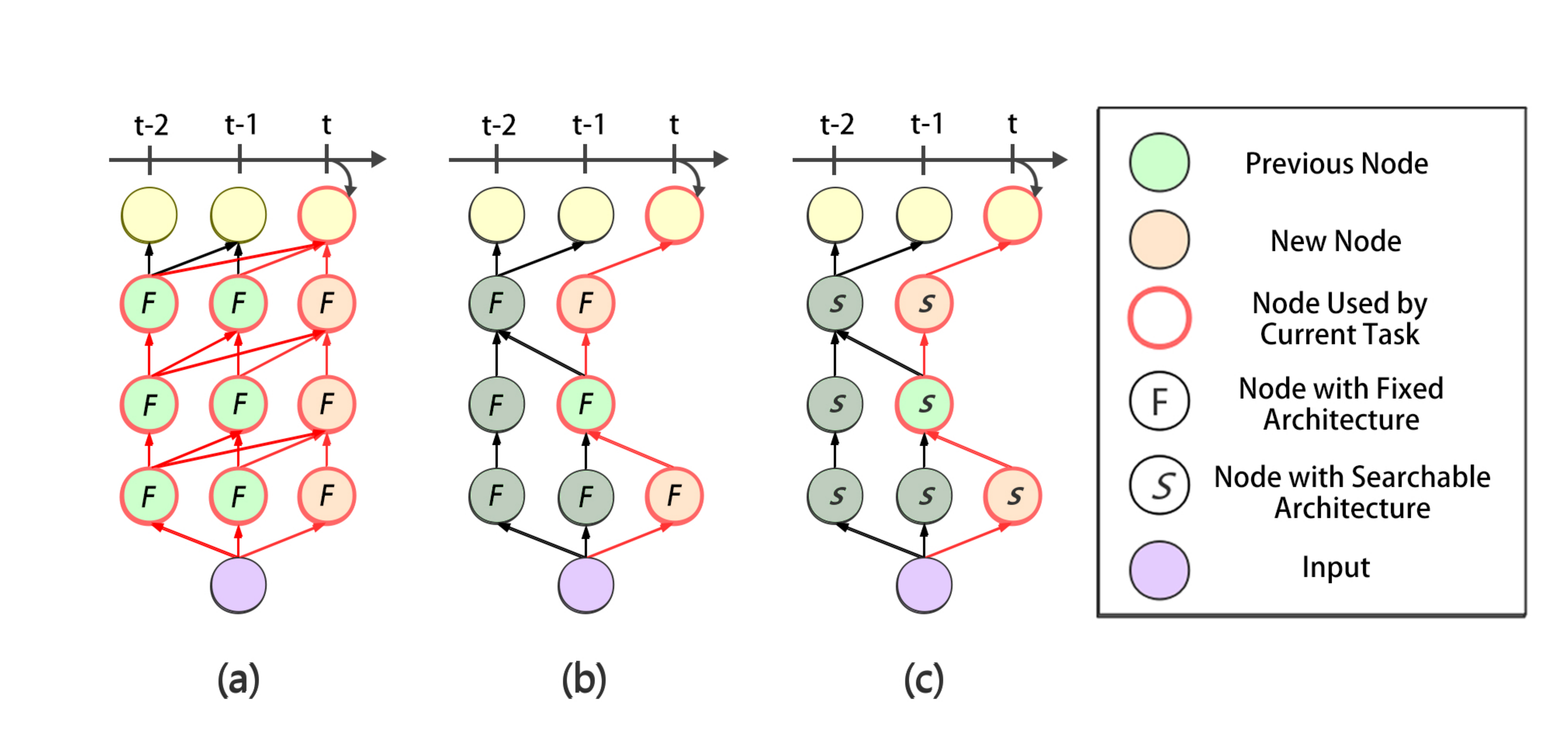}
    \caption{Types of lifelong Methods with expansion. (a) \textbf{Use the entire model} for current task. The architectures of all nodes are predefined and fixed. (b) \textbf{Use a sub-model} for current task. So, different models are used for different tasks. But the architectures of all nodes are also predefined and fixed. (c) \textbf{SEU}. Use a sub-model for current task and the architectures of all nodes are searchable.}
    \label{fig_types_of_methods}
\end{figure}

\emph{Expansion methods} overcome the capacity bottleneck by introducing new model resources, i.e parameters, to better deal with new tasks \cite{progressive_2016, DEN_2018, RCL_2018,ltg}. The number of parameters allocated to different layers is different and different methods have different allocation strategies. In general, newly allocated parameters are learnable, while old parameters are fixed or regularized. After new parameters are allocated, there are two ways to learn the current task model. One is to use the entire model for the current task (see \figurename{\ref{fig_types_of_methods}(a)}), and the other is to use only part of the entire model for the current task, namely the sub-model (see \figurename{\ref{fig_types_of_methods}(b)}). The latter takes into account the impact of the structure of sub-model on the current task to some extent.  

However, the number of model parameters in aforementioned methods tends to increase rapidly as the number of tasks increases. There are two main reasons. The first is that they all need a predefined original model to handle all tasks, which often leads to a bloated model. The second is that architectures of new resources assigned to new tasks are based on the same structure as the predefined model, which often leads to redundant unit parameters in the overall model. For example, if one layer of the original model is a $3 \times 3 conv$, then every time that the layer is added a $3 \times 3 conv$. The development of Neural Architecture Search (NAS) has shown that generic predefined original model and architecture of new resources cause: 1) too many redundant parameters. 2) limited performance on new tasks. To solve above problems and inspired by the Neural Architecture Search methods, in this paper, we propose a new concept called Extension Unit (EU). Based on the EU, we propose a new lifelong learning framework named Searchable Extension Units (SEU), in order to break down the need for a predefined original model and search for specific Extension Units for different tasks. Without compromising the performance of the model on different tasks, our approach can achieve a more compact model without catastrophic forgetting (see \figurename{\ref{fig_types_of_methods}(c)}).

Our primary contributions are as follows:

\begin{itemize}
    \item To the our best knowledge, we first propose to search different architectures of new neural networks for different tasks in Lifelong Learning problem by using the Neural Architecture Search methods.
    
    \item We present a new unit of new parameters in each layer of model called Extension Unit (EU) whose architecture is searchable. Based on the EU, we present a new lifelong learning framework named Searchable Extension Units, which can (1) search suitable architecture of EU for each layer of each task, (2) search suitable sub-model from the entire model for each task, (3) automatically construct the model from scratch.
    
    \item Our framework contains the following components: Extension Unit Searcher and Task Model Creator. Both component can adopt the appropriate methods. Therefore, our approach is flexible and expandable.
    
    \item We validate our framework on the PMNIST, the split CIFAR10 dataset, the split CIFAR100 dataset and the Mixture dataset. Our method can achieve higher accuracy with the significantly smaller entire model.
    
    \item We propose a new metric named mixed score (MS) that takes account both the accuracy and the number of parameters of the lifelong learning model. Our method achieves the highest mixed scores on all four datasets. 
    
    \item We release the source codes of experiments on github\footnote{\url{https://github.com/WenjinW/LLSEU}}.
\end{itemize}
    
The rest of this paper is organized as follows:  Section \ref{sec_SEU} introduces our new framework Searchable Extension Units (SEU). Section \ref{sec_exp} presents the evaluation protocol and results of our approach and examined ones. The related work of lifelong learning is introduced in Section \ref{sec_related}. Finally, we summarize our work and present future research directions in section \ref{sec_conclusion}.

\section{Lifelong Learning with Searchable Extension Units (SEU)}
\label{sec_SEU}
\begin{figure}[h]
    \begin{center}
    \includegraphics[width=1\linewidth]{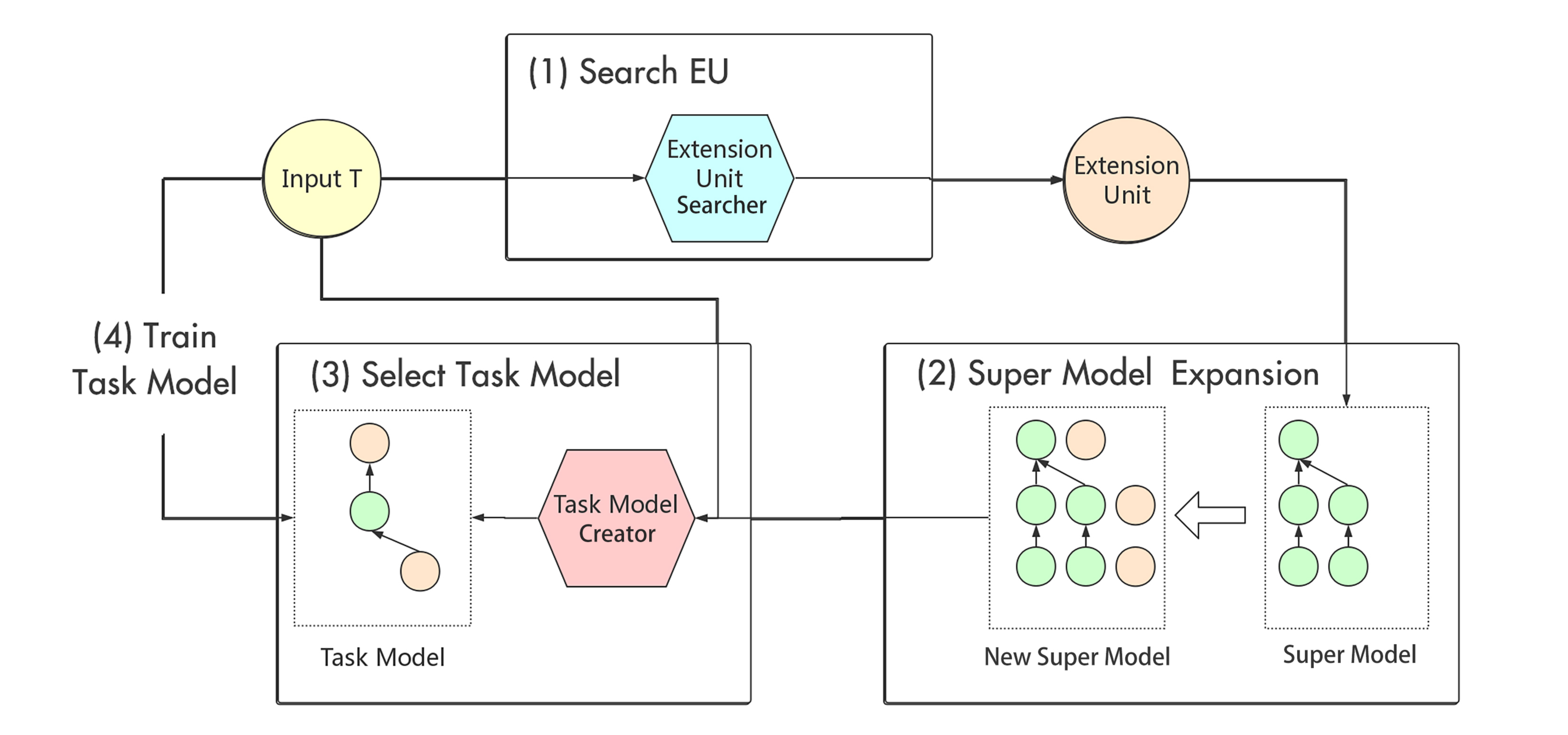}
    \end{center}
    \caption{\textbf{The basic framework of Searchable Extension Units.} \textbf{(1) Search EU:} Extension Unit Searcher search for an EU for current task according to task input. \textbf{(2) Expand Super Model:} Add an EU searched in (1) at each layer of Super Model. \textbf{(3) Create Task Model:} Select an EU from each layer of Super Model. These EUs make up the Task Model for current Task. New EUs that are not used will be deleted. \textbf{(4) Train Task Model:} Train the Task Model for current task.}
    \label{fig_framework}
\end{figure}

\subsection{Problem Formulation}
A formal definition of lifelong learning is as follows: The model needs to learn a sequence of tasks $T = \{T_1, T_2, ..., T_M\} $ one by one under two constraints. The first constraint is that the model can not forget the knowledge learned from previous tasks while learning a new task. And the second one is that only samples of the current task are available while learning the current task. Each task $T_m$ goes with a training dataset $D_m^{train} = \{(x_{m,i}, y_{m,i}); i = 1, 2, ..., n_m^{train}\}$, where $x_{m,i}$ is the sample, $y_{m,i}$ is the label and $n_m^{train}$ is the number of samples. Similarly, task $T_m$ goes with a validation dataset $D_m^{valid}=\{(x_{m,i}, y_{m,i}); i = 1, 2, ..., n_m^{valid}\}$. Then, the entire datasets are $D^{train}=\bigcup_{m=1}^MD_m^{train}$ and $D^{valid}=\bigcup_{m=1}^MD_m^{valid}$. The second constraint means only $D_m^{train}$ and $D_m^{valid}$, not $D^{train}$ and $D^{valid}$, are available when learning task $T_m$.

\subsection{Basic Concepts and Framework}
\begin{figure*}
\centering
\includegraphics[width=1\linewidth]{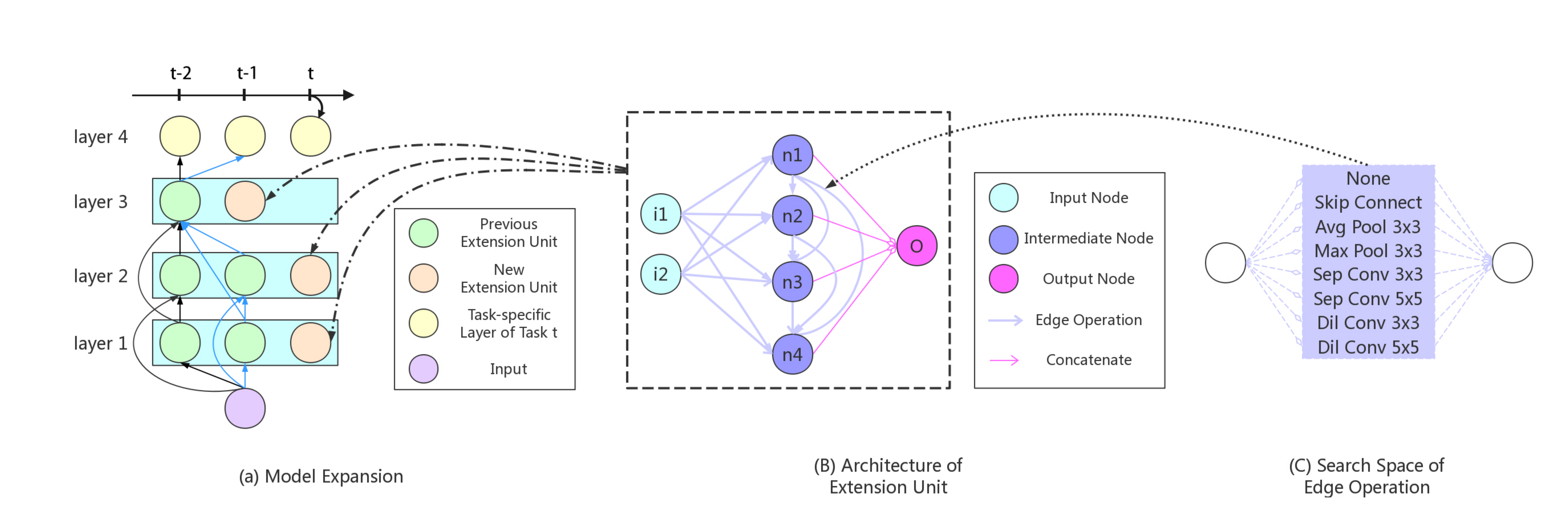}
\caption{\textbf{Search EU and model expansion.} \textbf{(a) Model Expansion} The current task is T. The model has 3 intermediate layers and 1 task-specific layer. Each intermediate layer already has multiple EUs. The model adds a new EU in each layer at first. Subsequently, the model will select an EU from each layer to form the sub-model dealing with the current task. \textbf{(b) Extension Unit (EU)} is the smallest unit of a model in SEU. An EU has 2 input nodes, 4 intermediate nodes and 1 output node. The value of an intermediate node is calculated by accumulating results of Edge Operations on the values of input nodes and intermediate nodes it depends on. The output node concatenates the values of 4 intermediate nodes. \textbf{(3) Search Space of Edge Operation}. An Edge Operation between two nodes is sampled from a search space containing 8 operations.}
\label{fig_search_extension}
\end{figure*}

In this section, we present the terms used in this paper and briefly describe the basic components and workflow of SEU. The definitions of basic terms are as follows:

\begin{itemize}
    \item \textbf{Super Model} is the entire multi-head model to handle all $M$ tasks and it is empty at the beginning. It has $N$ intermediate layers and $1$ task-specific output layer.
    \item \textbf{Task Model} is part of Super Model and is responsible for a specific task $T_m$. All task models have the same number of layers in this paper. Each intermediate layer of Task Model has only one EU that either is newly created for the current task or reuses the existing one.
    \item \textbf{Task-specific Layer} is the last output layer of the Super Model and contains $M$ output heads (FC layers). Each Task Model has a corresponding output head.
    \item \textbf{Intermediate Layer}. All layers in the Super Model except the task-specific output layer are intermediate layers. Each intermediate layer is made up of multiple EUs. On each layer, each task has a specific EU and multiple tasks can share one EU.
    \item \textbf{Extension Unit (EU)} is the smallest model unit in this paper. The architecture of an Extension Unit is searched by Neural Architecture Search methods (see \figurename{\ref{fig_search_extension}}). Each EU contains 2 input nodes, 4 intermediate nodes and 1 output nodes. The operation of the input edge of the intermediate node is searched out from a search space containing 8 operations.
\end{itemize}

\figurename{\ref{fig_framework}} shows the workflow of our approach, containing 2 main components: \emph{Extension Unit Searcher} and \emph{Task Model Creator}. The functions of these components are as follows: 

\begin{itemize}
    \item \textbf{Extension Unit Searcher} is used to search the most suitable architecture of EU for current task $T_m$. Several methods\cite{darts, Block_Wise_nas_2018, MDL_2019} that focus on searching a micro-architecture can be used here. We adopt the Multinomial Distribution Learning method in this paper due to its high-efficiency \cite{MDL_2019}.
    \item \textbf{Task Model Creator} is used to create Task Model for current task $T_m$. It will select an EU out of candidate ones, i.e., the newly-created EU for new knowledge and existing EUs for past knowledge in each layer of Super Model. 
\end{itemize}

For each task, the workflow of our approach consists of four phases (see \figurename{\ref{fig_framework}}): \textbf{(1) Search EU.} At first, \emph{Extension Unit Searcher} will search for a suitable architecture of EU for the current task from search space. \textbf{(2) Expand Super Model.} Based on the selected EU, the Super Model will expand a column, i.e., adding a new EU in each intermediate layer. \textbf{(3) Create Task Model.} Then, the \emph{Task Model Creator} will select the best EU for current task in each intermediate layer of Super Model. With the corresponding output head, these EUs make up the Task Model. At the same time, new EUs introduced in second phase will be deleted except those that are selceted. Note that different Task Models may share the same EU in some layers. \textbf{(4) Train Task Model.} Finally, the Task Model will be trained and reserved for later evaluation. 

\subsection{Algorithm:SEU}
Suppose current task is $T_t$, we denote the parameters of Task Model for task $(1,2,...,t)$ as $(\theta_1,\theta_2,...,\theta_t)$. So, our goal is to minimize cumulative loss of the model on tasks $(T_1, T_2, ..., T_t)$. The total loss $L_{total}$ is
\begin{equation}
\label{eq_loss_total}
L_{total} = \Sigma_{m=1}^{t}\Sigma_{i=1}^{n_m^{train}}loss_m(f_{\theta_m}(x_{m,i}), y_{m,i})
\end{equation} where $loss_m$ is the loss function of task $T_m$ and $f_{\theta_m}$ is the Task Model of task $T_m$. The problem is, limited by the constraint that only $D_t^{train}$ is available, the loss of data about previous tasks $L_{pre}$ is not directly calculable.
\begin{equation}
    L_{total} = L_{pre} + L_{t}
\end{equation}
\begin{equation}
\label{eq_loss_pre}
    L_{pre} = \Sigma_{m=1}^{t-1}\Sigma_{i=1}^{n_m^{train}}loss_m(f_{\theta_m}(x_{m,i}), y_{m,i})
\end{equation}
\begin{equation}
\label{eq_loss_t}
    L_{t} = \Sigma_{i=1}^{n_t^{train}}loss_t(f_{\theta_t}(x_{t,i}), y_{t,i})
\end{equation}
Obviously, parameters $(\theta_1, \theta_2, ..., \theta_{t-1})$ have been optimized before learning task $T_t$. We can denote $\Theta_{t-1}=\bigcup_{m=1}^{t-1}\theta_m$. Then, by freezing $\Theta_{t-1}$, we can keep $L_{pre}$ as a constant $C$ which is already a lower value. Then, we get an approximate optimization target
\begin{equation}
    \min_{\Theta_t\backslash\Theta_{t-1}}L_{total}=\min_{\Theta_t\backslash\Theta_{t-1}}C+\Sigma_{i=1}^{n_t^{train}}loss_t(f_{\theta_t}(x_{t,i}), y_{t,i})
\end{equation} where $\Theta_t=\bigcup_{m=1}^t\theta_m$. It is important to note that $\theta_i\cap\theta_j$ where $i\neq j$ might not be an empty set. Model parameters can be shared between different tasks.

In order to better deal with task $T_t$, we need to introduce new parameters $\theta_{t,n}$ for task $T_t$ into the super model, while employing the previous knowledge in the existing parameters $\theta_{t,p}$. Then the parameters of task $T_t$ is $\theta_t=\theta_{t,n}\cup\theta_{t,p}$ where $\theta_{t,p}=\theta_t\cap\Theta_{t-1}$. If $\theta_t \in \Theta_{t-1}$, the optimization does not happen because of $\Theta_t\backslash\Theta_{t-1}=\varnothing$.   

The acquisition of $\theta_t$ is divided into two stages. At first, we search for a suitable architecture $\theta^e$ of EU for task $T_t$ and add $\theta^{e,l}=\theta^e$ to each intermediate layer $l$ of Super Model. $\theta^{new}=\bigcup_{l=1}^N\theta^{e,l}$ where $N$ is the number of intermediate layers. Leave the details of producing $\theta^e$ in section \ref{sec_search_expand}, then we can select $\theta_t$ from $\Theta_t^*=\Theta_{t-1}\cup\theta^{new}$. The selected parameters for task $T_t$ is $\theta_t=\theta_{t,n}\cup\theta_{t,p}$ where $\theta_{t,n}\in\theta^{new}, \theta_{t,p}\in\Theta_{t-1}$ and will be trained. More details are presented in section \ref{sec_select_train} and we summarize the workflow as algorithm \ref{alg_total}.

\begin{algorithm}
    \caption{SEU}
    \label{alg_total}

    \textbf{Input} $D^{train}=\{D_1^{train}, D_2^{train}, ..., D_M^{train}\}$
    
    \textbf{Require} $\Theta_0 = \varnothing$ \# initialize Super Model
    
    \begin{algorithmic}[1]
        \FOR{$t=1$ to $M$}
            \STATE $\theta^e = ExtensionUnitSearcher(D_t^{train})$
            \STATE $\Theta_t^* = ExpandSuperModel(\theta^e, \Theta_{t-1})$
            \STATE $\Theta_t, \theta_t = TaskModelCreator(D_t^{train}, \Theta_t^*)$
            \STATE $\theta_t = TrainTaskModel(D_t^{train}, \theta_t)$
        \ENDFOR
    \end{algorithmic}
\end{algorithm}

\subsection{Search and Expansion}
\label{sec_search_expand}
To introduce $\theta^{new}$, SEU needs to allocate new EUs for task $T_t$. Before doing so, SEU first needs to determine the architecture of these new EUs. So, the Extension Unit Searcher should search for an $\theta^e$ according to the training dataset of current task (see \figurename{\ref{fig_search_extension}b,\ref{fig_search_extension}c}). We directly adopt the Multinomial Distribution Learning \cite{MDL_2019} as the Extension Unit Searcher in this paper. The search space of an edge operation $o^{i,j}$ between node $i$ and $j$ contains $L$($=8$)  operations. So, we have
\begin{equation}
\label{eq_search_operation}
	o^{i,j} = \begin{cases}
	o_1\ &with\ probability\ p_1^{i,j} \\
	o_2\ &with\ probability\ p_2^{i,j} \\
	... &...\\
	o_L\ &with\ probability\ p_L^{i,j}
	\end{cases}
\end{equation} where $j\in(3,...,6)$ and $i\in(1,...,j-1)$. The accuracy record and epoch record of operations between node $i$ and $j$ are denoted as $\{A_l^{i,j};l=1,2,...,L\}$ and $\{E_l^{i,j};l=1,2,...,L\}$. All of accuracy records and epoch records are initialized to zero and the probability of each operation in one edge is initialized with $\frac{1}{L}$. So,
\begin{equation}
\label{eq_search_initialize_AE}
    A=\{A_l^{i,j}=0\};E=\{E_l^{i,j}=0\}    
\end{equation}
\begin{equation}
\label{eq_search_initialize_p}
    P=\{p_l^{i,j}=\frac{1}{L}\}
  \end{equation} where $j\in(3,...,6)$, $i\in(1,...,j-1)$, and $l\in(1,...,L)$. To update $P$, then, we will take $K$ samples. After each sample, the sampled network will be trained with one epoch where $s_{i,j}$ is the ID of selected operation in edge between node $i$ and $j$. Suppose the evaluation accuracy after training is $A^*$, the update rules of $A, E, P$ are as follows:
\begin{equation}
\label{eq_search_update_ae}
    A_{s_{i,j}}^{i,j}=A^*;E_{s_{i,j}}^{i,j}=E_{s_{i,j}}^{i,j}+1
\end{equation}
\begin{equation}
\label{eq_search_reward}
    reward_{s_{i,j}}^{i,j}=\Sigma_{l=1}^L\mathbb{I}(A_{s_{i,j}}^{i,j}>A_l^{i,j},E_{s_{i,j}}^{i,j}<E_l^{i,j})
\end{equation}
\begin{equation}
\label{eq_search_penalty}
    penalty_{s_{i,j}}^{i,j}=\Sigma_{l=1}^L\mathbb{I}(A_{s_{i,j}}^{i,j}<A_l^{i,j},E_{s_{i,j}}^{i,j}>E_l^{i,j})
\end{equation}
\begin{equation}
\label{eq_search_update_p}
    p_{s_{i,j}}^{i,j}=p_{s_{i,j}}^{i,j}+\alpha(reward_{s_{i,j}}^{i,j}-penalty_{s_{i,j}}^{i,j})
\end{equation}In order to make sure that the sum of probabilities is $1$, we have to do softmax on the updated probability.
\begin{equation}
\label{eq_search_softmax}
    p_l^{i,j}=\frac{p_l^{i,j}}{\Sigma_{l=1}^{L}p_l^{i,j}}
\end{equation}

After all samples have been taken, the operation of each side is selected as the one with the highest probability whose ID is $l^*_{i,j}$ and the selected architecture of EU is
\begin{equation}
    \label{eq_theta_e}
    \theta^e=\{o^{i,j}=o_{l^*_{i,j}};i\in(1,...,j-1),j\in(3,...,6)\}
\end{equation}
The whole process is summarized in algorithm \ref{alg_EU_searcher}.
\begin{algorithm}
    \caption{ExtensionUnitSearcher}
    \label{alg_EU_searcher}
    
    \textbf{Input} $D_t^{train}=\{(x_{t,i},y_{t,i});i=1,2,...,n_t^{train})\}$
    
    \textbf{Require} $K$ \# the number of samples;
    
    \begin{algorithmic}[1]
    \STATE Initialize $P,A,E$ according to Equation (\ref{eq_search_initialize_AE}) and (\ref{eq_search_initialize_p});
    \FOR{$k=1$ to $K$}
    \STATE Sample the network according to Equation (\ref{eq_search_operation});
    \STATE Train the network with 1 epoch and evaluate it;
    \STATE Update $A,E$ according to Equation (\ref{eq_search_update_ae});
    \STATE Update $P$ according to Equation (\ref{eq_search_reward}),(\ref{eq_search_penalty}),(\ref{eq_search_update_p}),(\ref{eq_search_softmax});
    \ENDFOR
    \STATE Generate $\theta^e$ according to Equation (\ref{eq_theta_e});
    \end{algorithmic}
    \textbf{Return} $\theta^e$
\end{algorithm}

Now, with $\theta^e$, each layer of the Super Model will be expanded. We can denote the number of EUs that the Super Model already has in layer $j$ as $ne_j$ and parameters of EUs in layer $j$ as $\beta_{j,1}, \beta_{j,2}, ..., \beta_{j,ne_j}$. Then we have
\begin{equation}
    \Theta_{t-1} = \bigcup_{j=1}^N\bigcup_{k=1}^{ne_j}\beta_{j,k}
\end{equation}
In each layer $j$, a new EU is added whose parameters are contained in $\beta_{j,ne_j+1}$. So, we have
\begin{equation}
\label{eq_theta_n}
    \theta^{new}=\{\beta_{j,ne_j+1}=\theta^e;j=1,2,...,N\}
\end{equation}
\begin{equation}
\label{eq_expand}
    \Theta_t^*=\Theta_{t-1}\cup\theta^{new}= \Theta_{t-1}\cup\bigcup_{j=1}^{N}\beta_{j,ne_j+1}
\end{equation} And the expansion process is summarized in algorithm \ref{alg_expand_super_model}.
\begin{algorithm}
    \caption{ExpandSuperModel}
    \label{alg_expand_super_model}

    \textbf{Input} $\theta^e, \Theta_{t-1}$
    
    \begin{algorithmic}[1]
    \STATE Generate $\theta^{new}$ according to Equation (\ref{eq_theta_n});
    \STATE Expand $\Theta_{t-1}$ according to Equation (\ref{eq_expand});
    \end{algorithmic}
    \textbf{Return} $\Theta_t^*$
\end{algorithm}

\subsection{Creation and Training}
\label{sec_select_train}

\begin{algorithm}[t]
    \caption{TaskModelCreator}
    \label{alg_task_creator}
    \textbf{Input} $\Theta_t^*$
    
    \textbf{Input} $D_t^{train} = \{(x_i, y_i); i = 1, 2, ..., n\}$
    
    \textbf{Require} $K$ \# the number of training epochs;
    
    \begin{algorithmic}[1]
        \STATE Initialize $P, A, E$ according to Equation (\ref{eq_initial_ae}), (\ref{eq_initial_p});
        \FOR{$k=1$ to $K$}
        
        \STATE sample $\theta_t^*$ according to Equation (\ref{eq_sample});
        \STATE Train $\theta_t^*$ according to Equation (\ref{eq_train_grad}), (\ref{eq_train_update});
        \STATE Compute $A^*$ and Update $A, E$ according to Equation (\ref{eq_eval}), (\ref{eq_update_ae});
        \STATE Update $P$ according to Equation (\ref{eq_reward}), (\ref{eq_penalty}), (\ref{eq_update_p}),
        (\ref{eq_softmax});
        \ENDFOR
        \STATE Create $\theta_t$ according to Equation (\ref{eq_select_theta});
        \STATE Update the Super Model according to Equation (\ref{eq_EU_preserved}), (\ref{eq_EU_deleted});
    \end{algorithmic}
    \textbf{Return} $\Theta_t, \theta_t$
\end{algorithm}

\begin{figure*}
\centering
\includegraphics[width=1\linewidth]{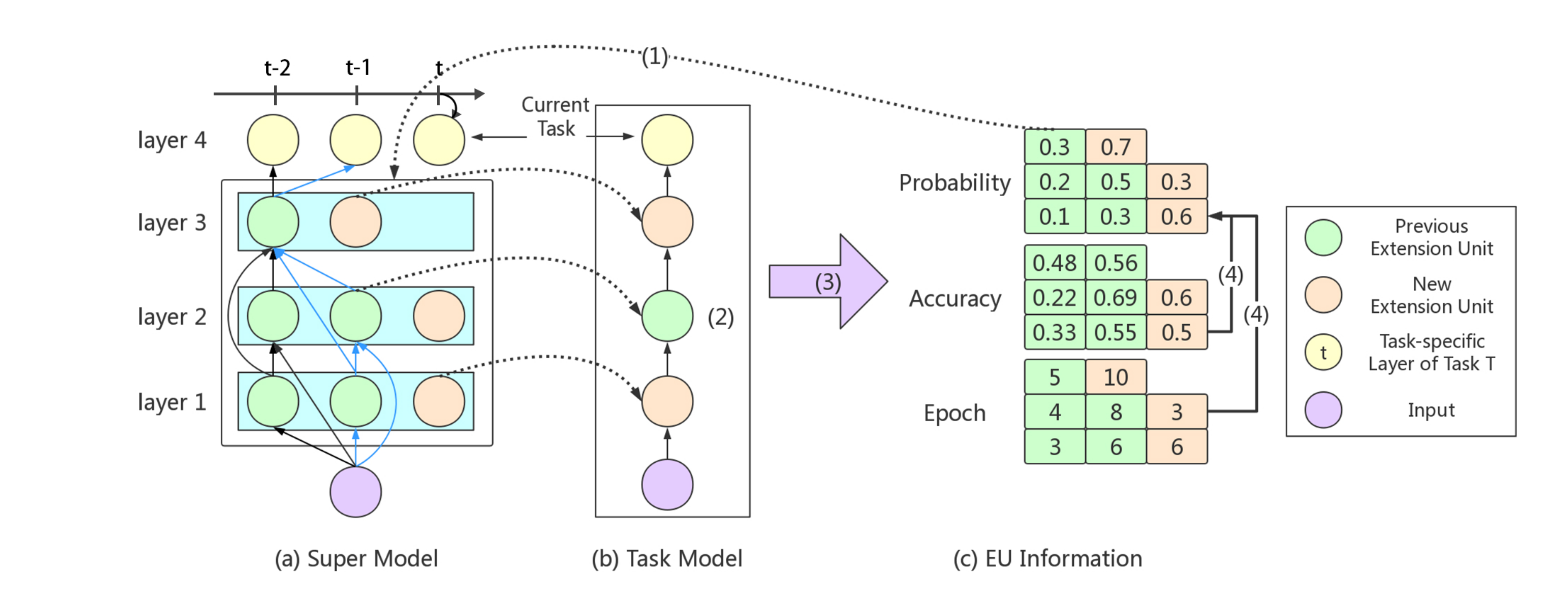}
\caption{\textbf{Create $t$-th Task Model.} \textbf{(c)} contains 3 types of information records associated with each EU in Super Model. Probability contains the sampling probability of each EU in Super Model. Accuracy contains the evaluation performance of each EU after it was last sampled and Epoch contains the total number it was sampled. The following loop is executed multiple times.  \textbf{(1)} Sample an EU from each intermediate layer according to sampling probability of EUs in (c). Together with the task-specific layer, these EUs make up the Task Model. \textbf{(2)} Train the Task Model sampled on the current task data with an epoch. \textbf{(3)} Record the number of epochs each EU is selected and accuracy of the most recent training of each EU. (4) Update the Probability according to the Accuracy and Epoch in (c).}

\label{fig_select_task_model}
\end{figure*}
After the model is expanded, SEU needs to create the Task Model for current task, i.e. select $\theta_t$ from $\Theta_t^*$ and only one EU will be used in each intermediate layer. There are two types of EUs in each layer, one that has been used and one that is new. If the selected EU has been used in the past, it will be frozen while learning task $T_t$. The new EUs have no such restriction.

We adopt the same idea in the Multinomial Distribution Learning method to search for appropriate EUs \cite{MDL_2019}. We assign each EU in the Super Model a probability of being selected and all EUs in the same layer start with the same probability. Each EU also has an accuracy record and a epoch record. The accuracy record represents the EU's evaluation performance after it was last selected and the epoch record represents the total number it was selected (see in \figurename{\ref{fig_select_task_model}c}). All of accuracy and epoch records are initialized to zero. 
\begin{equation}
\label{eq_initial_ae}
    A=\{A_{j,k}=0\}; E=\{E_{j,k}=0\}
\end{equation}
\begin{equation}
\label{eq_initial_p}
    P=\{p_{j,k}=\frac{1}{ne_j+1}\}
\end{equation} where $j\in(1,2,...,N)$ and $k\in(1,2,...,ne_j+1)$.
The probability of each EU will be updated according to accuracy and epoch record during sampling. Then, multiple samples are repeated (see \figurename{ \ref{fig_select_task_model}}). In each sample, the selected EUs form $t$-th Task Model which is trained one epoch. We can denote the selection operation in Super Model as $O(\Theta_t)$ and the selection result in layer $j$ as $o_j$. So we have \begin{equation}
	o_j = \begin{cases}
	\beta_{j,1}\ &with\ probability\ p_{j,1} \\
	\beta_{j,2}\ &with\ probability\ p_{j,2} \\
	... &...\\
	\beta_{j,ne_j+1}\ &with\ probability\ p_{j, ne_j+1}
		   \end{cases}
\label{eq_selection_layer}    
\end{equation}
\begin{equation}
\label{eq_sample}
    \theta_t^* = O(\Theta_t^*) = \bigcup_{j=1}^No_j
\end{equation}

Suppose the sampling result is $\theta_t^*=\bigcup_{j=1}^N\beta_{j,q_j}$. Then, we can train the $\theta_t^*$ one epoch and only update new parameters.
\begin{equation}
    \label{eq_train_grad}
    \Delta(\beta_{1,q_1}, \beta_{2,q_2}, ..., \beta_{N,q_N})=\frac{\partial\Sigma_{i=1}^{n/2}loss_t(f_{\theta_t^*}(x_i),y_i)}{\partial \theta_t^*}
\end{equation}
\begin{equation}
    \label{eq_train_update}
    \beta_{j,q_j} = \begin{cases}
        \beta_{j,q_j} - \mu\Delta\beta_{j,q_j}, & \text{if } q_j = ne_j+1\\
        \beta_{j,q_j}, & \text{otherwise}
    \end{cases},j\in(1,N)
\end{equation}
After training, the $t$-th Task Model will be evaluated and the performance is $A^*$. 
\begin{equation}
\label{eq_eval}
    A^*=\Sigma_{i=\frac{n}{2}+1}^{n}\mathbb{I}(f_{\theta_t^*}(x_h)==y_h)
\end{equation}
The update rules of $A, E, P$ are as follows:
\begin{equation}
\label{eq_update_ae}
    A_{j,q_j} = A^*; E_{j,q_j} = E_{j,q_j} + 1
\end{equation}
\begin{equation}
\label{eq_reward}
    reward = \Sigma_{z=1}^{ne_j+1} \mathbb{I}(A_{j,q_j}>A_{j,z},E_{j,q_j}<E_{j,z})
\end{equation}
\begin{equation}
\label{eq_penalty}
    penalty = \Sigma_{z=1}^{ne_j+1} \mathbb{I}(A_{j,q_j}<A_{j,z},E_{j,q_j}>E_{j,z})
\end{equation}
\begin{equation}
\label{eq_update_p}
    p_{j,q_j} = p_{j,q_j}+\alpha(reward-penalty)
\end{equation} where $j \in (1, 2, ..., N)$. In order to make sure that the sum of probabilities is 1, we need to do softmax on the updated probability.
\begin{equation}
\label{eq_softmax}
    p_{j,k} = \frac{p_{j,k}}{\Sigma_{k=1}^{ne_j+1}p_{j,k}}
\end{equation} where $j\in(1,2,...,N)$ and $k\in(1,2,...,ne_j+1)$.
After all epochs, we select the EU with the highest probability in each layer.
\begin{equation}
\label{eq_select_theta}
\theta_t = \bigcup_{j=1}^N\beta_{j,q^*}
\end{equation} where $p_{j,q^*}=max(p_{j,1}, p_{j,2}, ..., p_{j,ne_j+1})$. 

Then, the selected new EUs introduced in expansion stage will remain, and the others will be deleted. If $p_{j,q^*}=p_{j,ne_j+1}$, where $j\in(1,N)$ which means the new EU is selected, then the EU will be preserved. 
\begin{equation}
    \label{eq_EU_preserved}
    ne_j=ne_j+1 \text{ if } p_{j,q^*}=p_{j,ne_j+1}
\end{equation}Otherwise, it will be deleted.
\begin{equation}
    \label{eq_EU_deleted}
    \Theta_t=\Theta_t^*\backslash\{\beta_{j,ne_j+1}|p_{j,q^*}\neq p_{j,ne_j+1},j\in(1,N)\}
\end{equation} The whole process is shown in Algorithm \ref{alg_task_creator}.

After the $t$-th Task Model is created, it will be trained on the training data of task $t$. The Task Model contains two types of EUs: existing ones and new ones. The parameters of EUs used by previous tasks will be frozen and will not be updated during training. We denote the parameters of Task Model that contains $N$ intermediate layers as $\theta_t={\beta_j;j=1,2,...,N}$ and the training rules are as follows:
\begin{equation}
    \label{eq_train_task_grad}
    \Delta(\beta_1, \beta_2, ..., \beta_N)=\frac{\partial\Sigma_{i=1}^nloss_t(f_{\theta_t}(x_{t,i}),y_{t,i})}{\partial\theta_t}
\end{equation}
\begin{equation}
    \label{eq_train_task_update}
    \beta_j=\begin{cases}
        \beta_j-\mu\Delta\beta_j, & \text{if } \beta_j\not\in\Theta_{t-1} \\
        \beta_j, &\text{otherwise}
    \end{cases},j\in(1,N)
\end{equation}The training process is summarized in algorithm \ref{alg_train_task_model}.

\begin{algorithm}[h]
    \caption{TrainTaskModel}
    \label{alg_train_task_model}

    \textbf{Input} $D_t^{train}=\{(x_{t,i},y_{t,i};i=1,2,...,n_t^{train})\}$
    
    \textbf{Input} $\theta_t=\{\beta_j;j=1,2,...,N\}$ 
    
    \textbf{Require} K \# the number of training epochs;
    
    \begin{algorithmic}[1]
    \FOR{$k=1$ to $K$}
    \STATE Update parameters according to Equation (\ref{eq_train_task_grad}), (\ref{eq_train_task_update});
    \ENDFOR
    \end{algorithmic}
    \textbf{Return} $\theta_t$
\end{algorithm}
\section{Experiments}
\label{sec_exp}
\subsection{Baselines, Datasets, and Implementation Details}
We compare our method(SEU) with SGD, EWC\cite{ewc}, Learn to Grow(LTG)\cite{ltg}, and Progressive Network(PN)\cite{progressive_2016}. The SGD trains the entire model when learning a new task. We conduct experiments on 4 different datasets: permuted MNIST(PMNIST), split CIFAR10, split CIFAR100, and Mixture dataset. The PMNIST comes from the MNIST\cite{MNIST} dataset. A unique fixed random permutation is used to shuffle the pixels of each sample to generate a task. The split CIFAR10 is split from CIFAR10\cite{CIFAR} and contains 5 2-class classification tasks. Similarly, the split CIFAR100 contains 5 20-class classification tasks. The Mixture contains 5 different task: EMNIST\cite{EMNIST}, MNIST\cite{MNIST}, Fashion MNIST\cite{FMNIST}, SVHN\cite{SVHN}, and KMNIST\cite{KMNIST}. Each experiment is repeated 5 times in order to obtain stable results.

For SGD, EWC, PN, and LTG, we adopt the Alexnet as the initial model in all experiments. SGD, EWC, and PN only contain training stage. LTG is divided into training stage and search stage. SEU contains 3 stages: EU Searching, Task Model Creation, and trainging stage.

In the training stage of all methods, we adopt the Stochastic Gradient Descent as optimizer and it contains 5 hyper-parameters: an initial learning rate $\alpha=0.025$ which is annealed to $0$ following a cosine schedule, a momentum $\beta=0.9$, a weight decay $\lambda$, a number of training epochs $e=50$, and a batch size $b=128$. The value of $\lambda$ is selected from $\{0.0003,0.001,0.003\}$, which yields the best experimental result. EWC contains one more hyper-parameter $\lambda_e$ to adjust the proportion of penalty item in loss function. The value of $\lambda_e$ is $50000$ on the split CIFAR10 and $20000$ on others.

In the search stage of LTG, there are two optimizers. One for parameters of model (Stochastic Gradient Descent), and the other for parameters of architecture
(Adam). The former contains an initial learning rate $\alpha_m=0.025$, a momentum $\beta_m=0.9$, and a weight decay $\lambda_m=0.0003$. The latter contains an initial learning rate $\alpha_a=0.025$, a momentum $(\beta_1=0.5,\beta_2=0.999)$ and a weight decay $\lambda_a=0.001$. The number of epochs $e_a=50$ and the batch size $b_a=128$. The coefficient of the penalty term for model size $\lambda_{size}$ is selected from $\{0, 0.001, 0.01, 0.1\}$. 

In SEU, EU Searching uses a optimizer (Stochastic Gradient Descent) that contains an initial learning rate $\alpha_{es}=0.025$, a momentum $\beta_{es}=0.9$, a weight decay $\lambda_{es}$ (same as $\lambda$). The coefficient of probability update is $\mu_{es}=0.01$ and the number of layers of model is $l_{es}=4$. The number of epochs is $e_{es}=100$ and the batch size is $b_{es}=512$. The optimizer of Task Model Creation is the same as the one in EU Searching. The number of epochs $e_{ts}=100$ and the batch size $b_{ts}=128$. The number of layers of model $l_{ts}$ is $5$ on the PMNIST and is $6$ on others. The coefficient of probability update is $\mu_{ts}=0.01$. 

\subsection{Ability to Resist Forgetting}
\begin{figure*}[ht]
    \subfloat[The accuracy curve of each task from the beginning learning to the completion of all tasks on the split CIFAR10.]{
        \begin{minipage}[]{0.2\linewidth}
        \centering
        \includegraphics[width=1\linewidth]{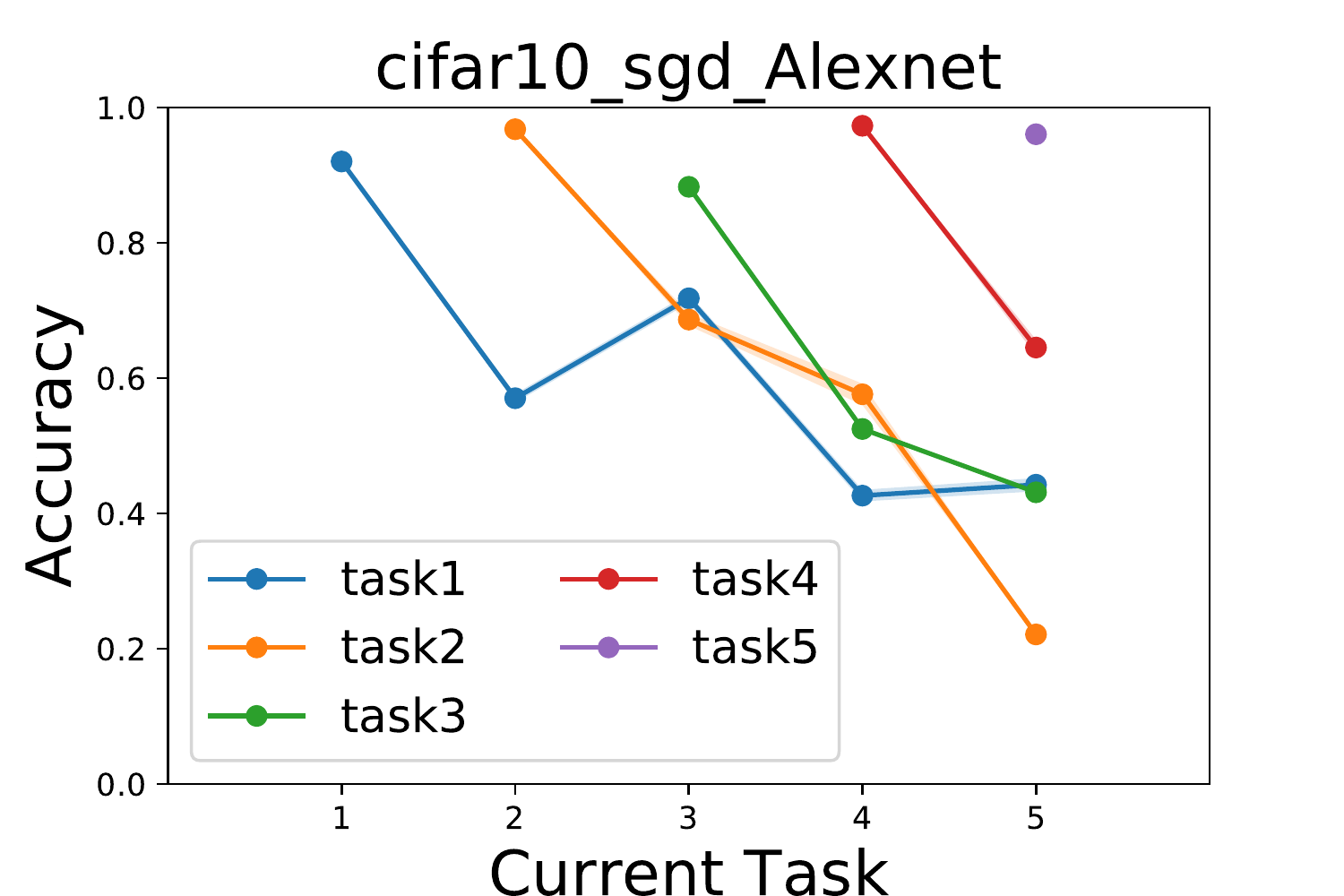}
        \end{minipage}
        \begin{minipage}[]{0.2\linewidth}
        \centering
        \includegraphics[width=1\linewidth]{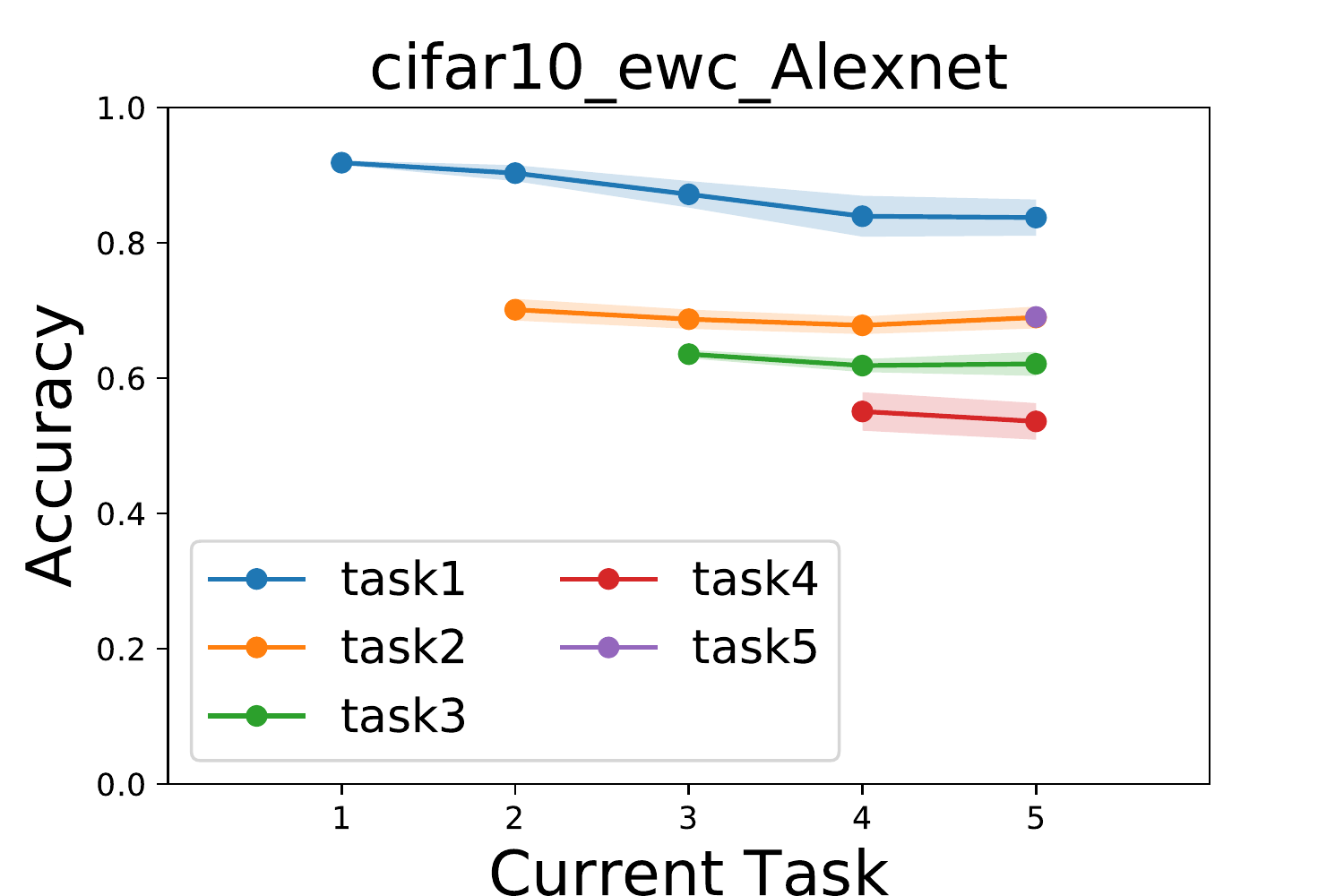}
        \end{minipage}%
        \begin{minipage}[]{0.2\linewidth}
        \centering
        \includegraphics[width=1\linewidth]{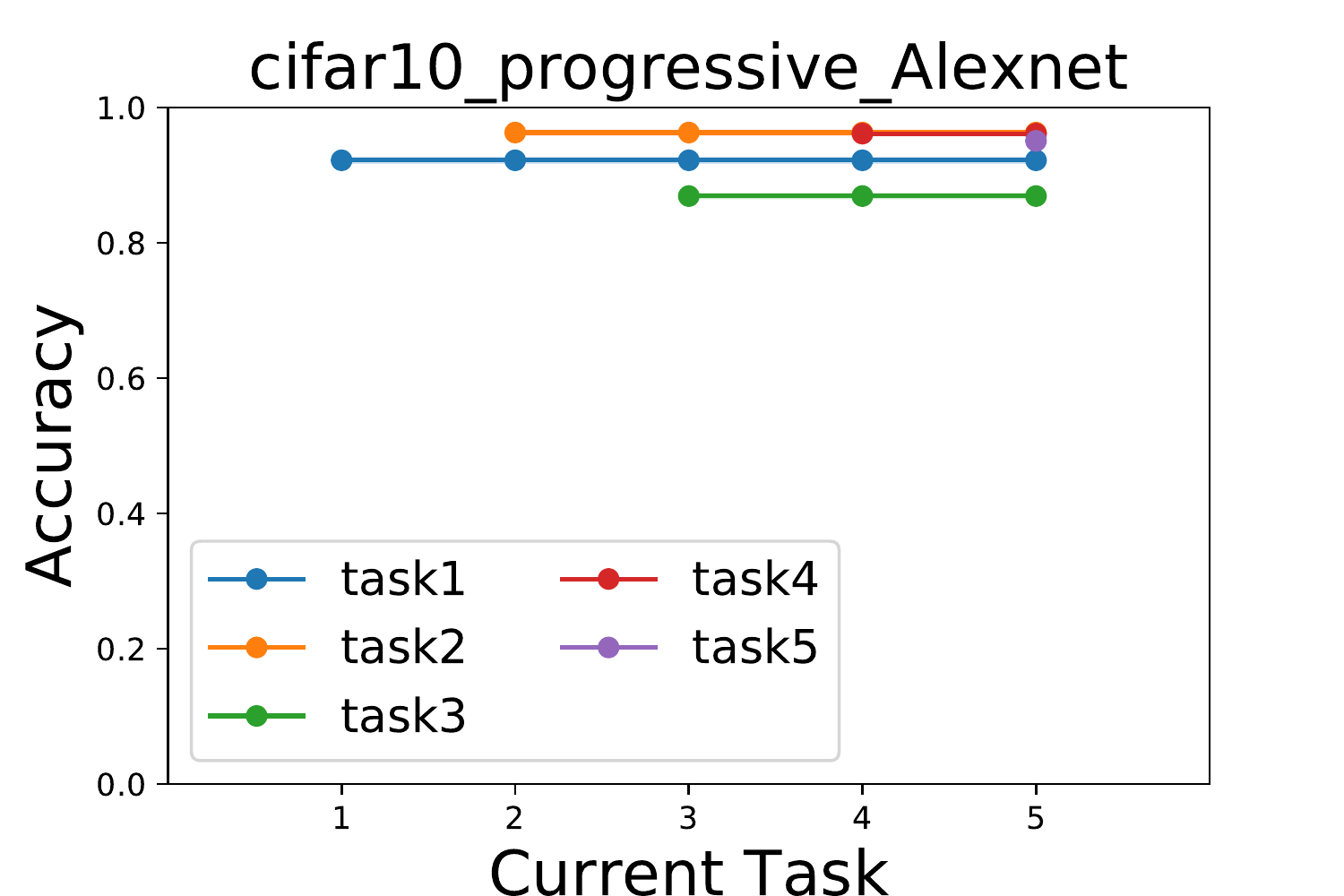}
        \end{minipage}
        \begin{minipage}[]{0.2\linewidth}
        \centering
        \includegraphics[width=1\linewidth]{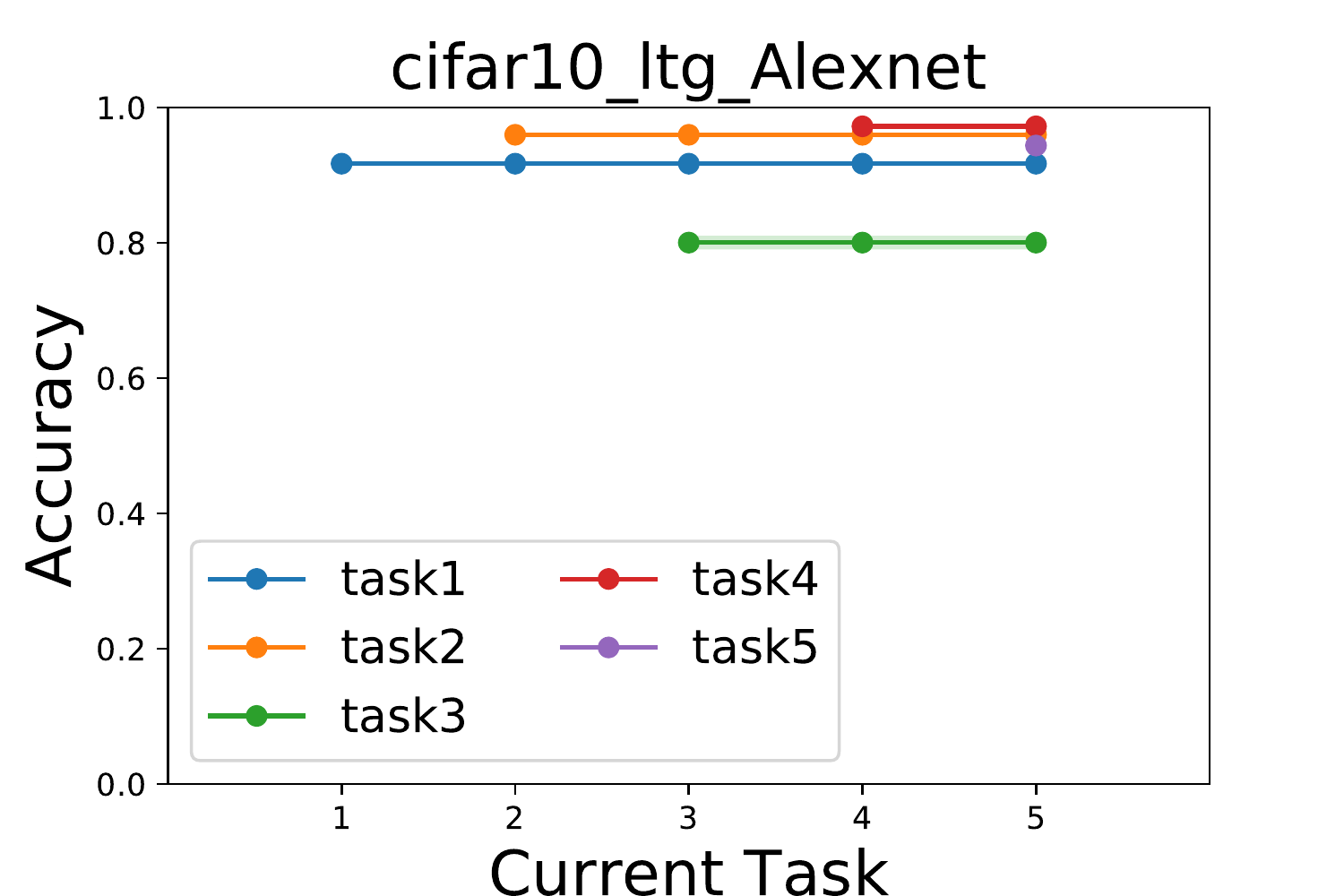}
        \end{minipage}
        \begin{minipage}[]{0.2\linewidth}
        \centering
        \includegraphics[width=1\linewidth]{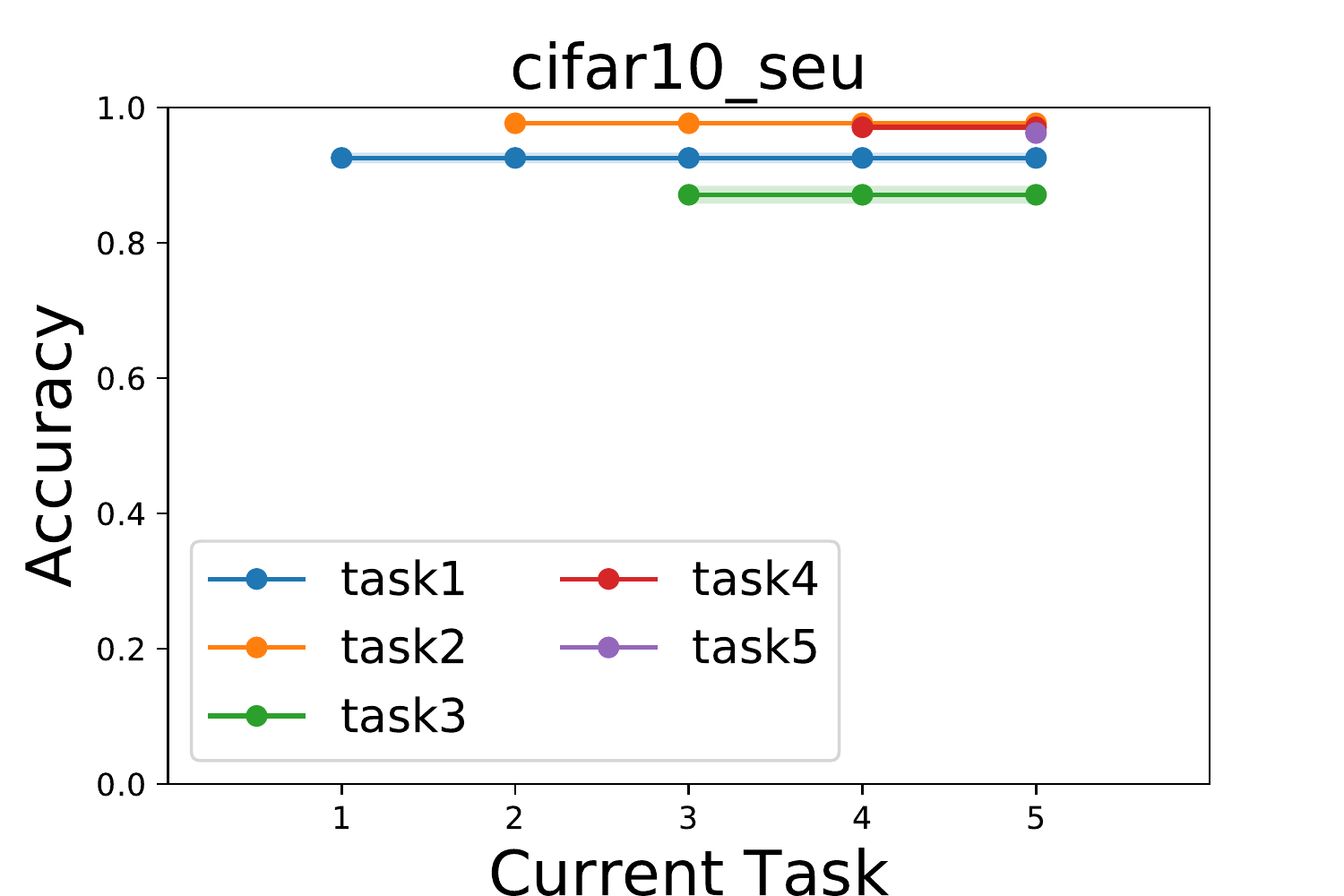}
        \end{minipage}
        \label{exp_acc_curve_cifar10}
    }
    
    \subfloat[The original accuracy at the beginning and the retained accuracy after completing all tasks of each task on the split CIFAR10.]{
        \begin{minipage}[]{0.2\linewidth}
        \centering
        \includegraphics[width=1\linewidth]{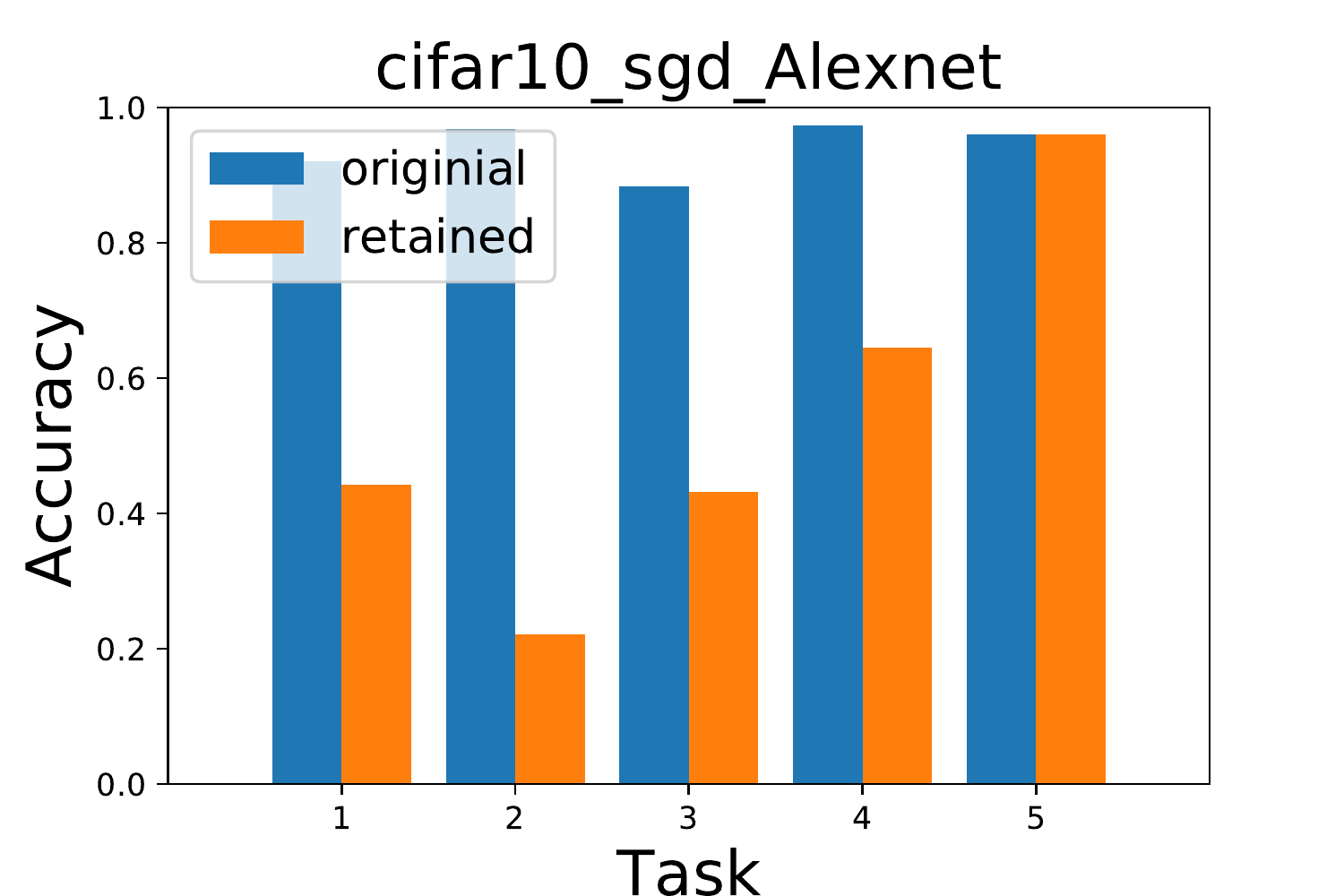}
        \end{minipage}
        \begin{minipage}[]{0.2\linewidth}
        \centering
        \includegraphics[width=1\linewidth]{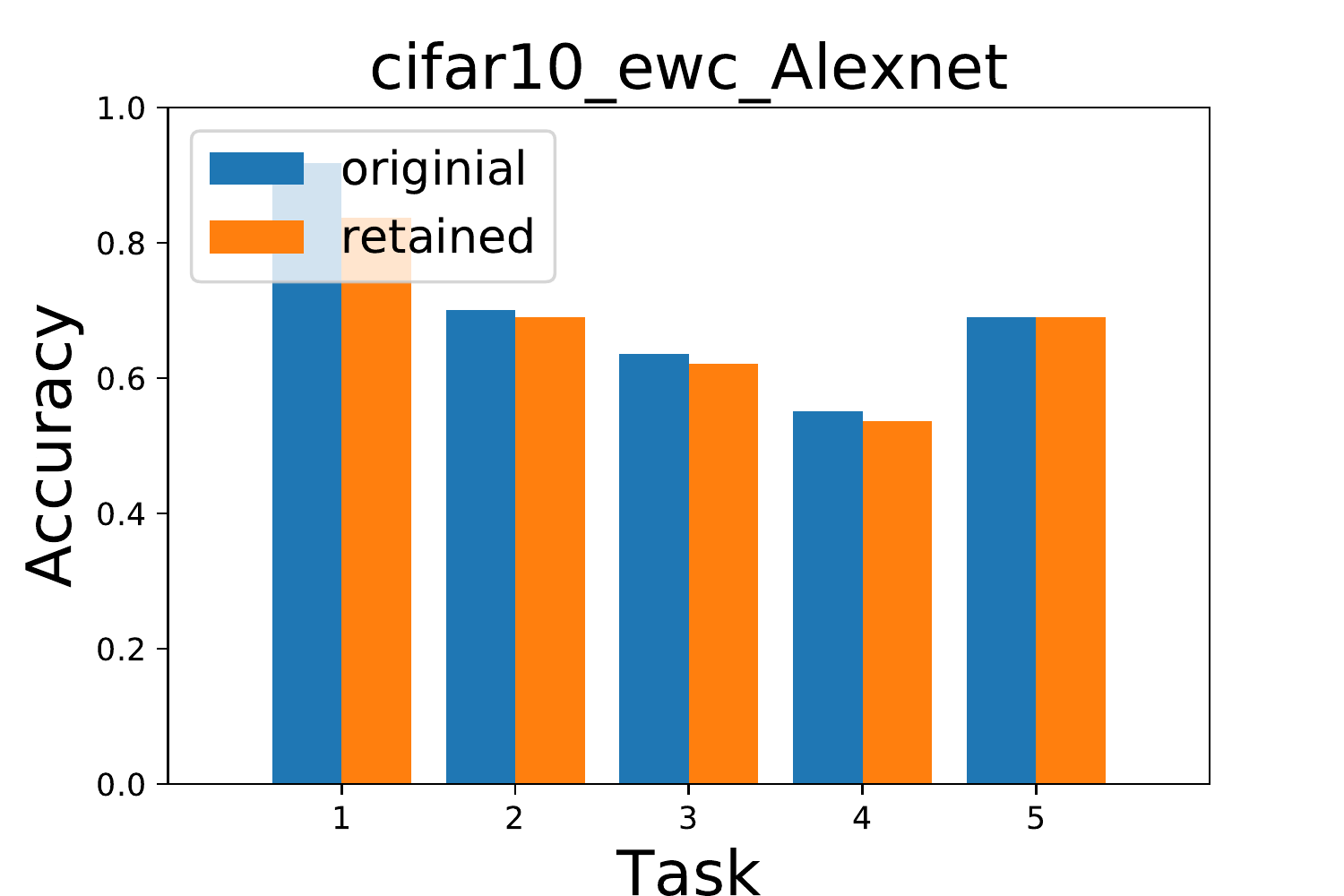}
        \end{minipage}%
        \begin{minipage}[]{0.2\linewidth}
        \centering
        \includegraphics[width=1\linewidth]{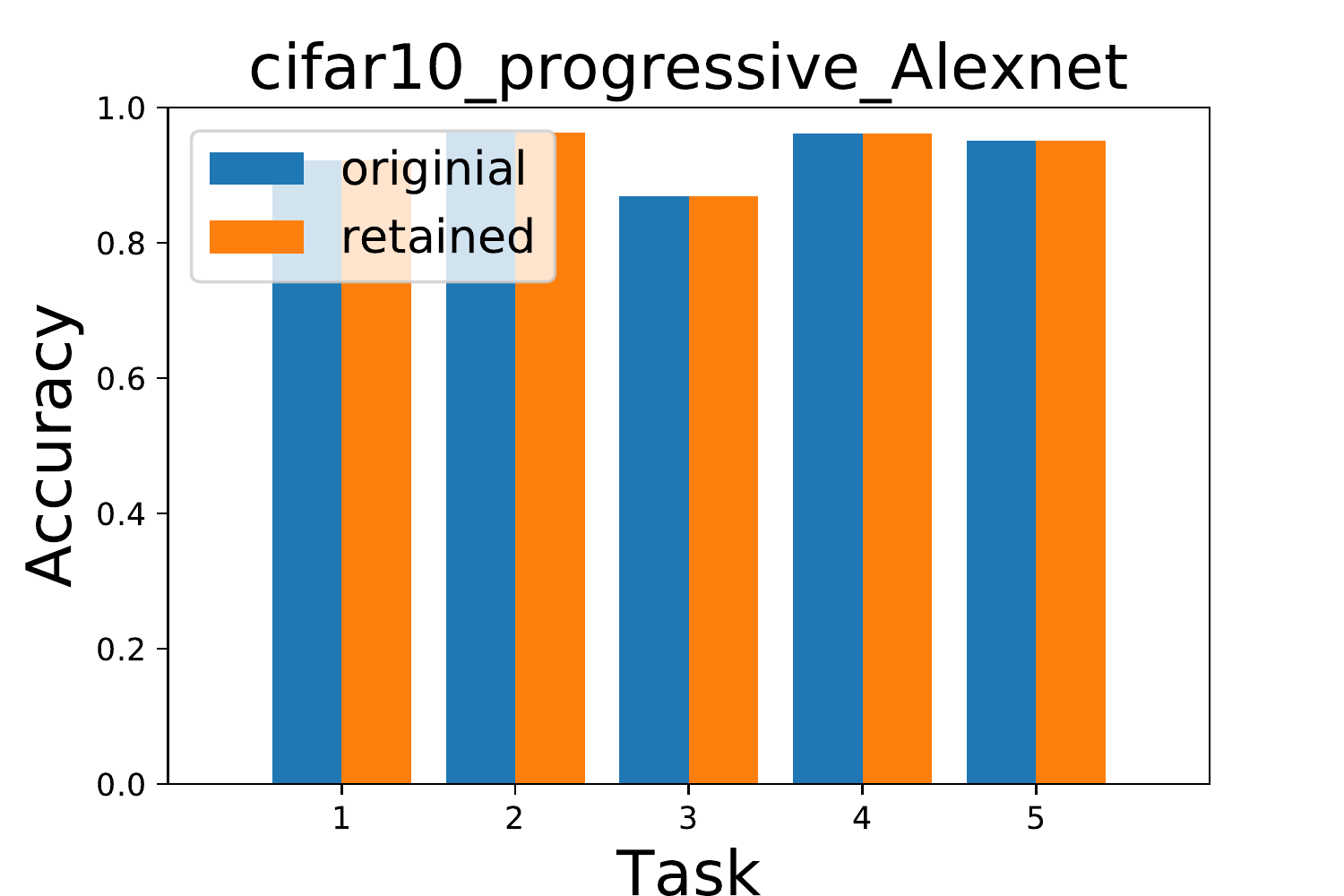}
        \end{minipage}
        \begin{minipage}[]{0.2\linewidth}
        \centering
        \includegraphics[width=1\linewidth]{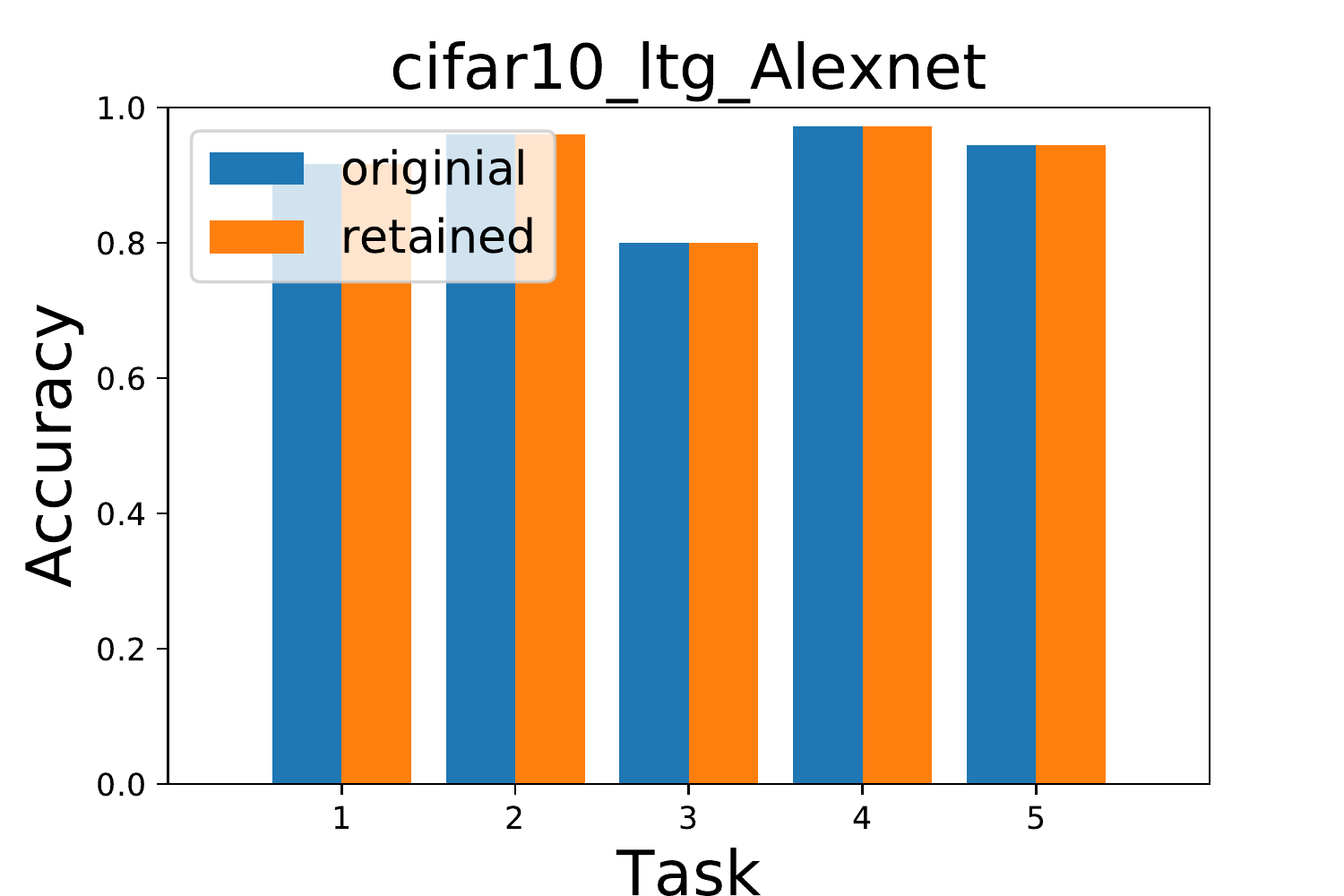}
        \end{minipage}
        \begin{minipage}[]{0.2\linewidth}
        \centering
        \includegraphics[width=1\linewidth]{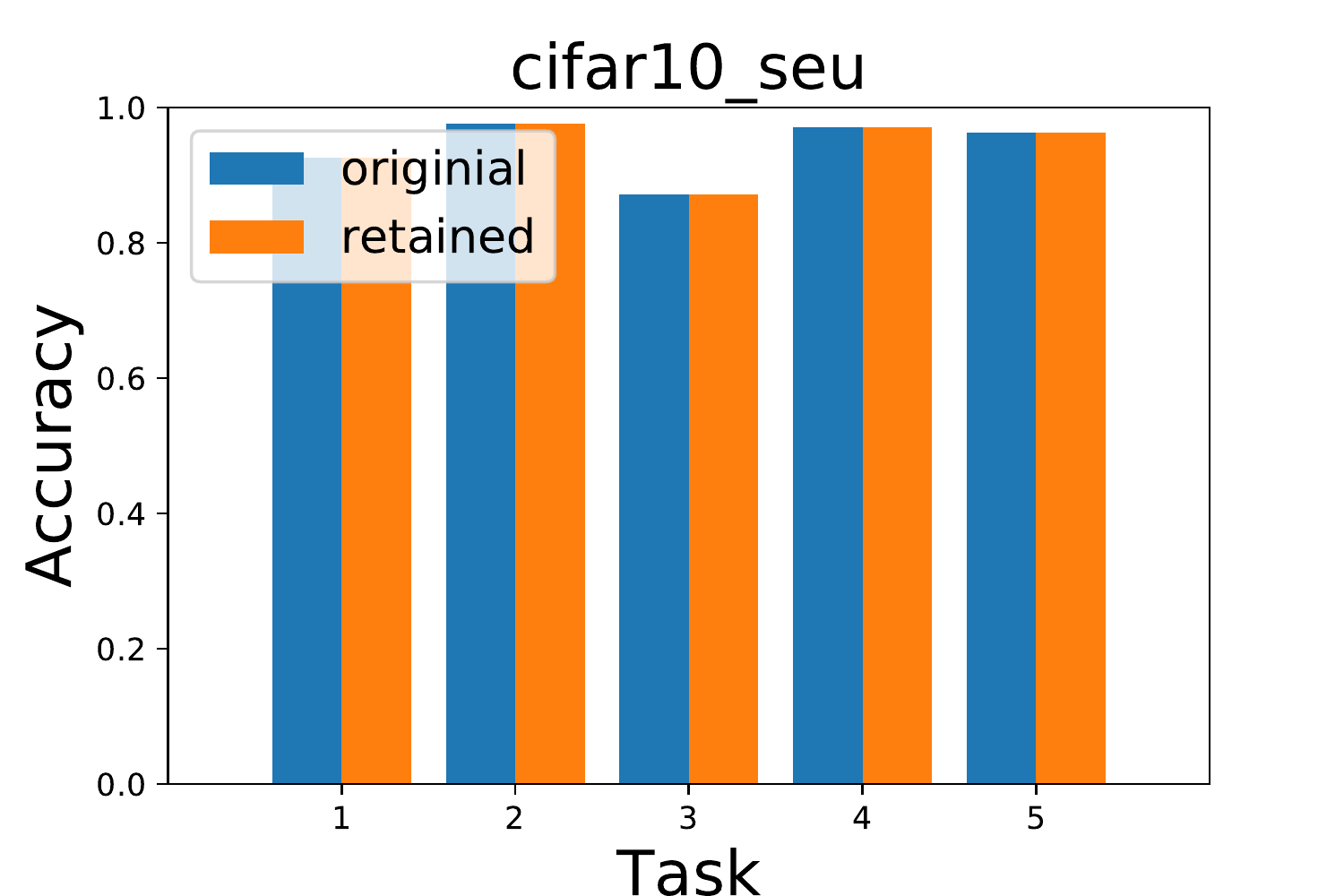}
        \end{minipage}
        \label{exp_acc_change_cifar10}
    }
    
    \subfloat[The accuracy curve of each task from the beginning learning to the completion of all tasks on the split CIFAR100.]{
        \begin{minipage}[]{0.2\linewidth}
        \centering
        \includegraphics[width=1\linewidth]{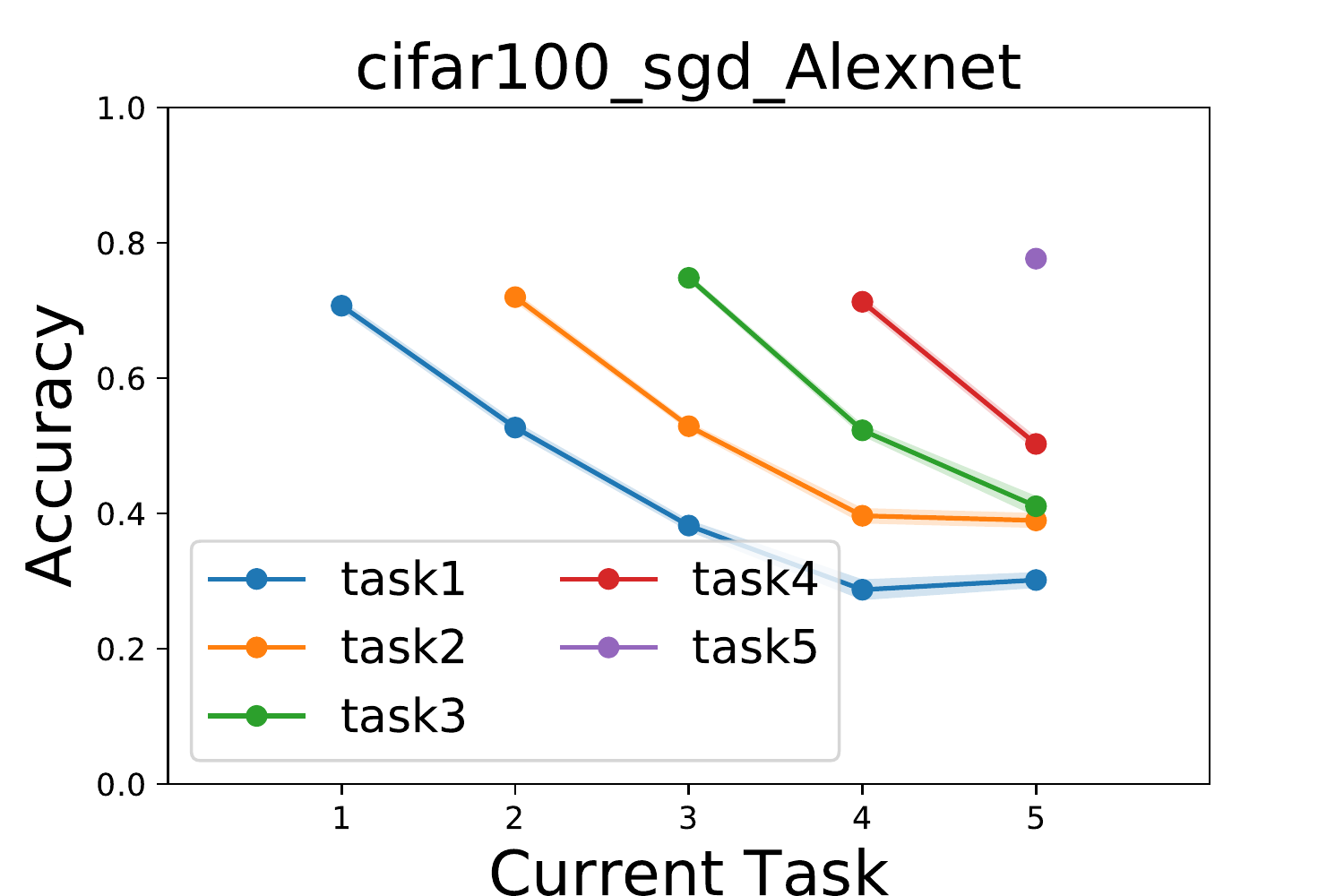}
        \end{minipage}
        \begin{minipage}[]{0.2\linewidth}
        \centering
        \includegraphics[width=1\linewidth]{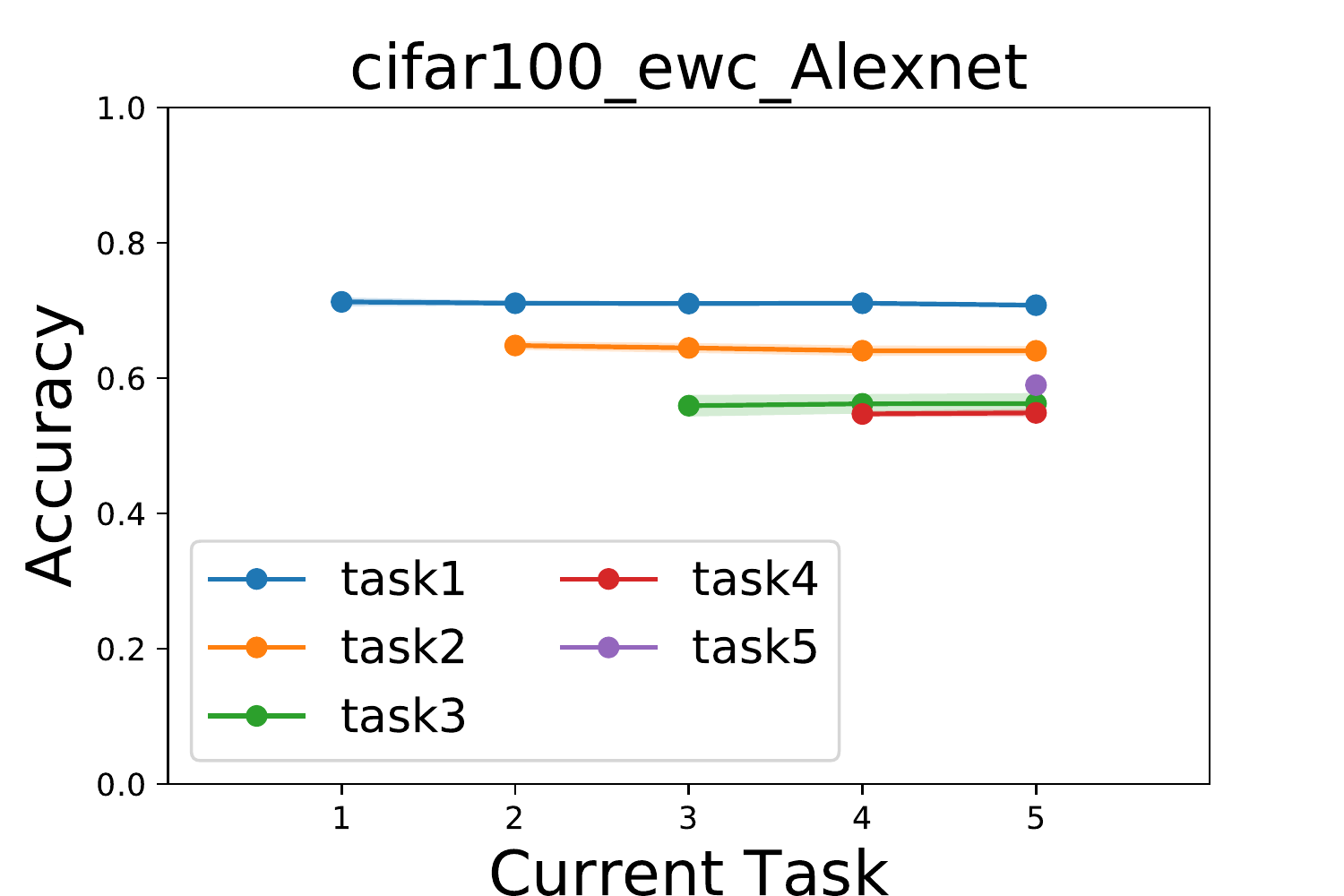}
        \end{minipage}%
        \begin{minipage}[]{0.2\linewidth}
        \centering
        \includegraphics[width=1\linewidth]{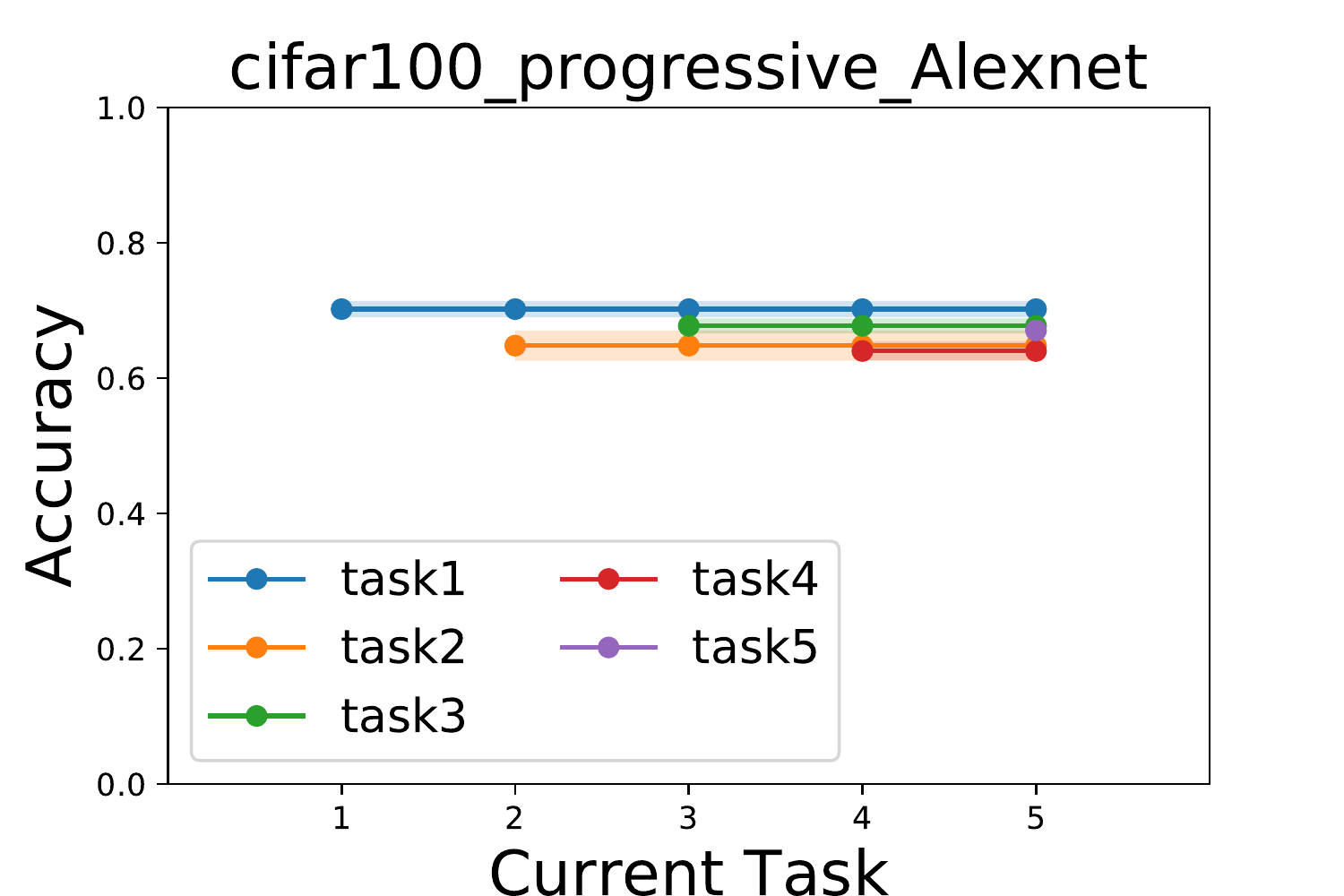}
        \end{minipage}
        \begin{minipage}[]{0.2\linewidth}
        \centering
        \includegraphics[width=1\linewidth]{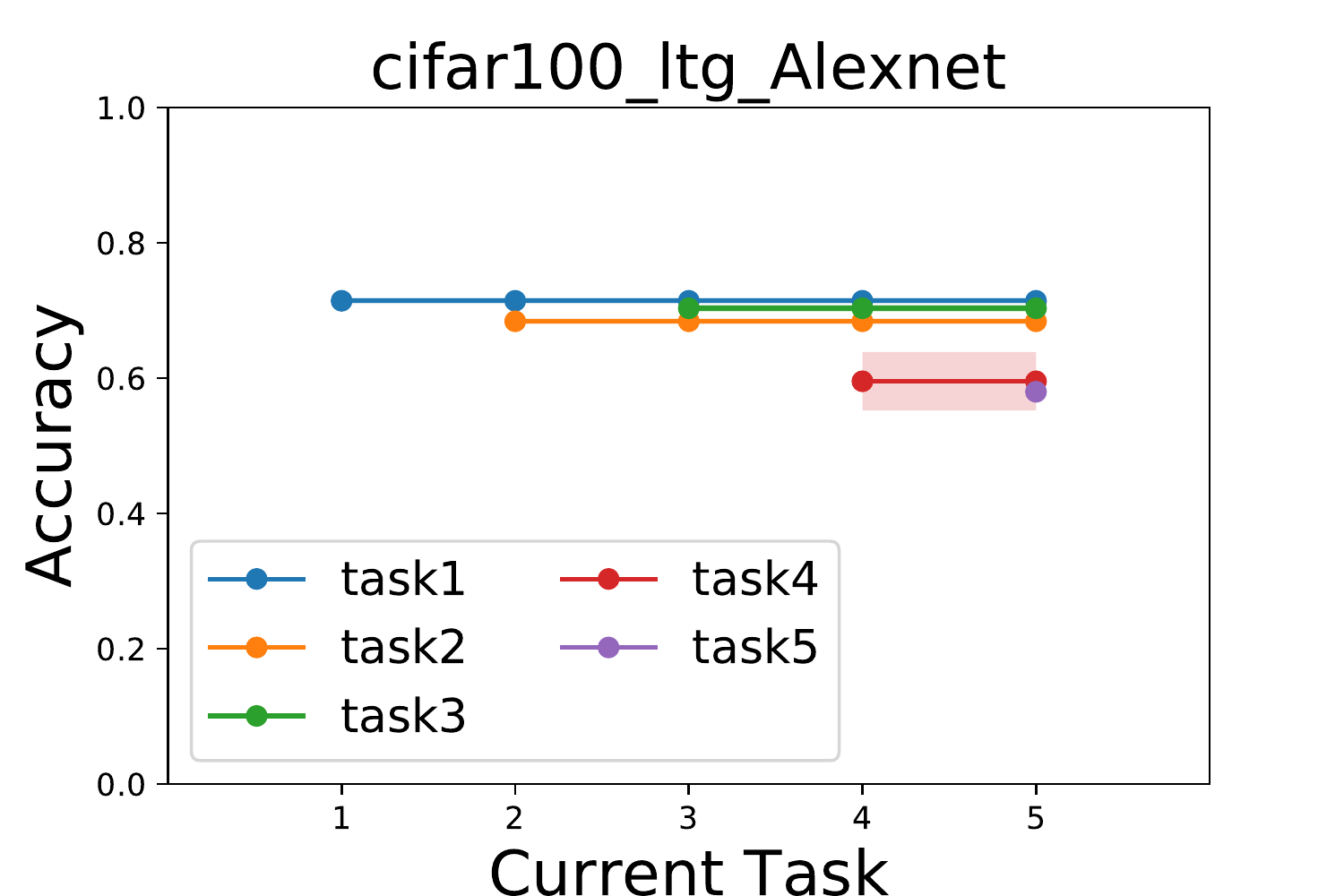}
        \end{minipage}
        \begin{minipage}[]{0.2\linewidth}
        \centering
        \includegraphics[width=1\linewidth]{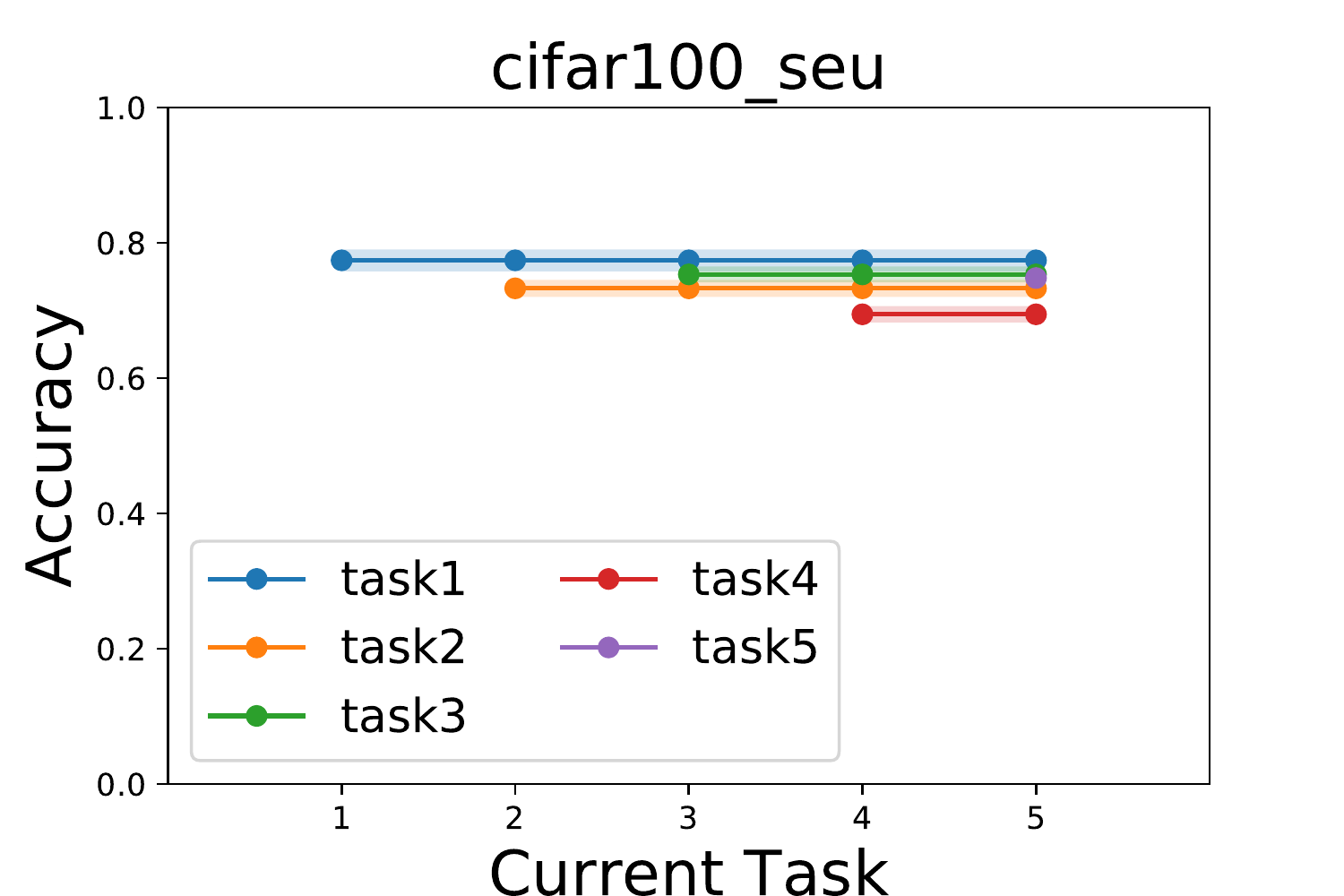}
        \end{minipage}
        \label{exp_acc_curve_cifar100}
    }
    
    \subfloat[The original accuracy at the beginning and the retained accuracy after completing all tasks of each task on the split CIFAR100.]{
        \begin{minipage}[]{0.2\linewidth}
        \centering
        \includegraphics[width=1\linewidth]{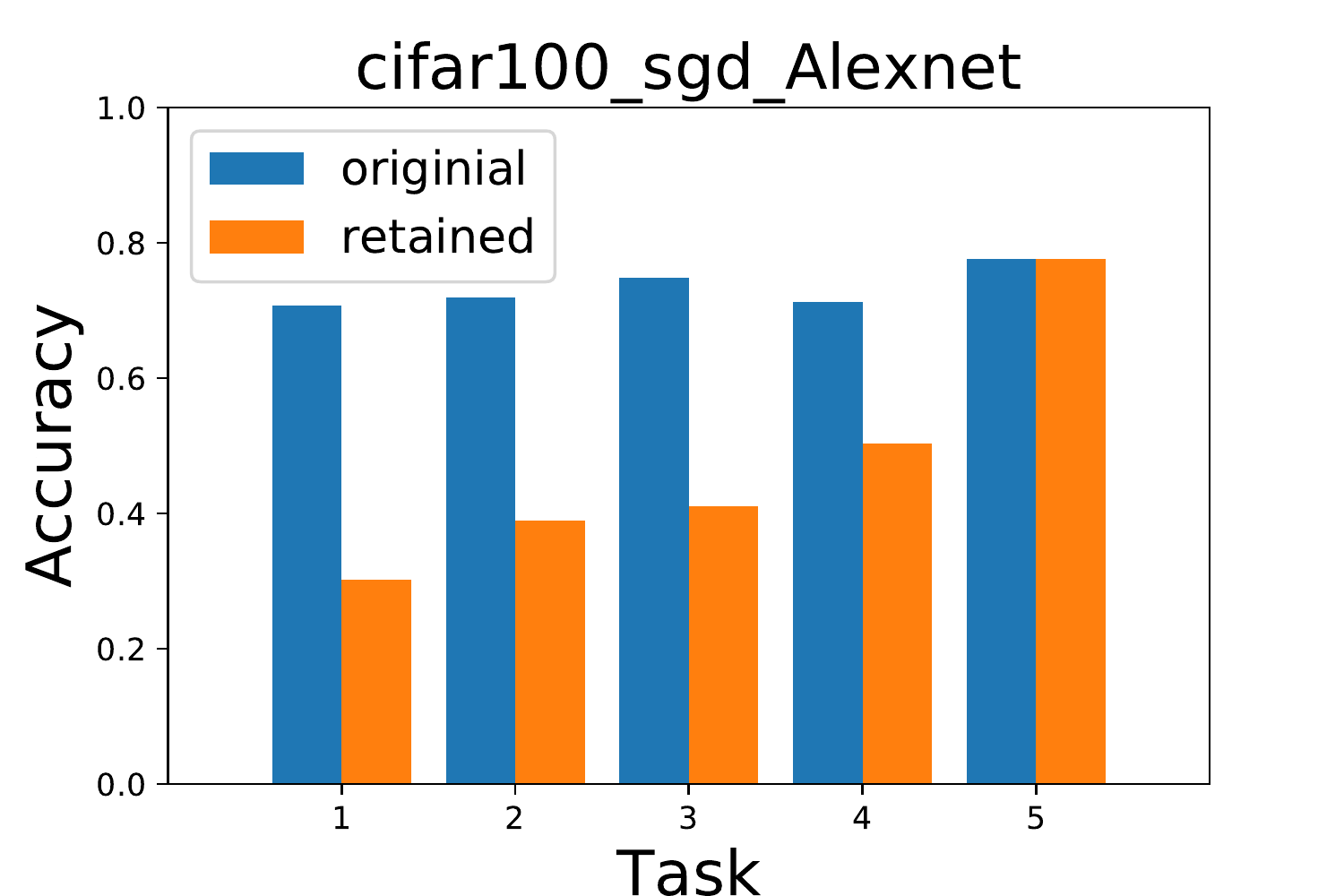}
        \end{minipage}
        \begin{minipage}[]{0.2\linewidth}
        \centering
        \includegraphics[width=1\linewidth]{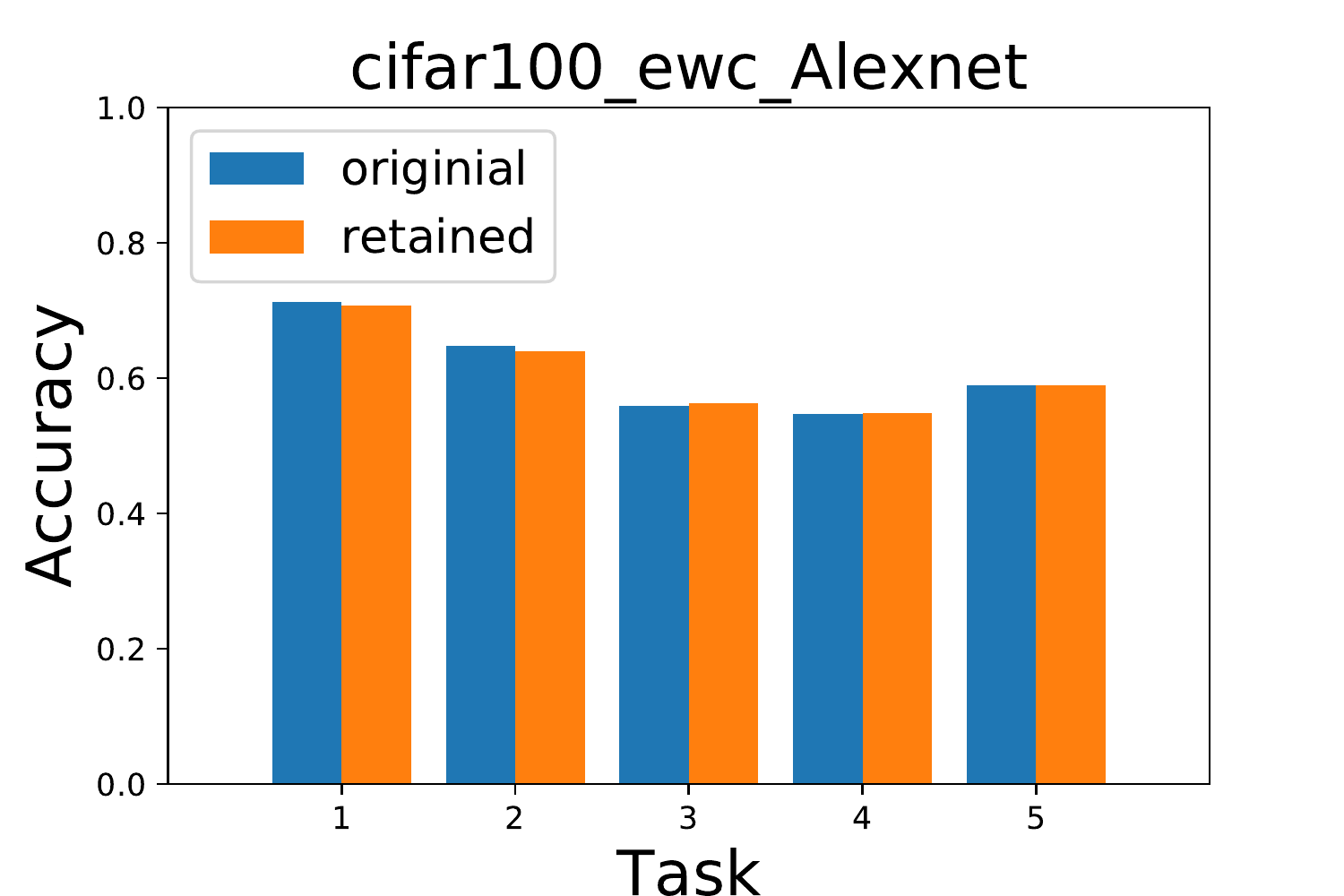}
        \end{minipage}%
        \begin{minipage}[]{0.2\linewidth}
        \centering
        \includegraphics[width=1\linewidth]{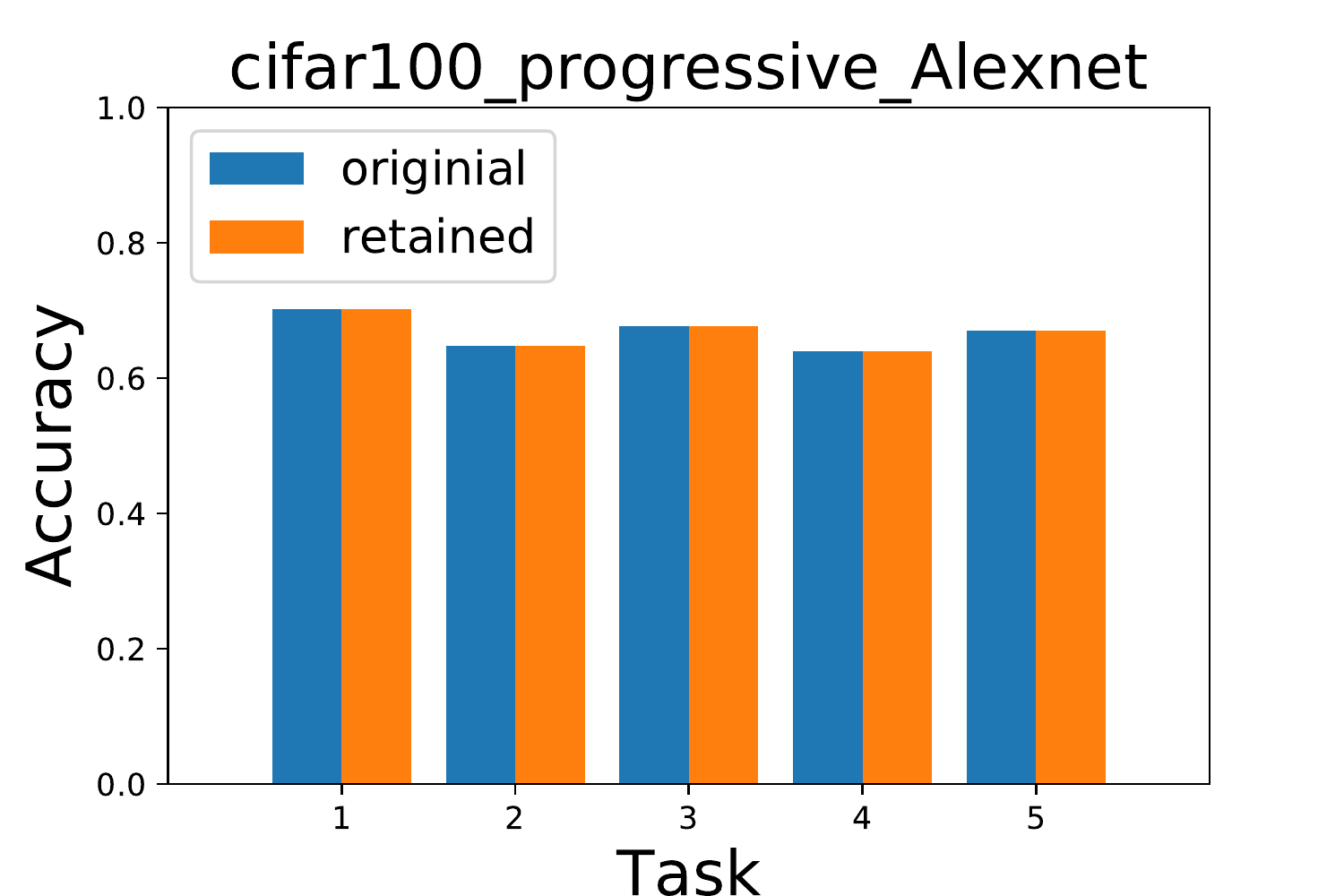}
        \end{minipage}
        \begin{minipage}[]{0.2\linewidth}
        \centering
        \includegraphics[width=1\linewidth]{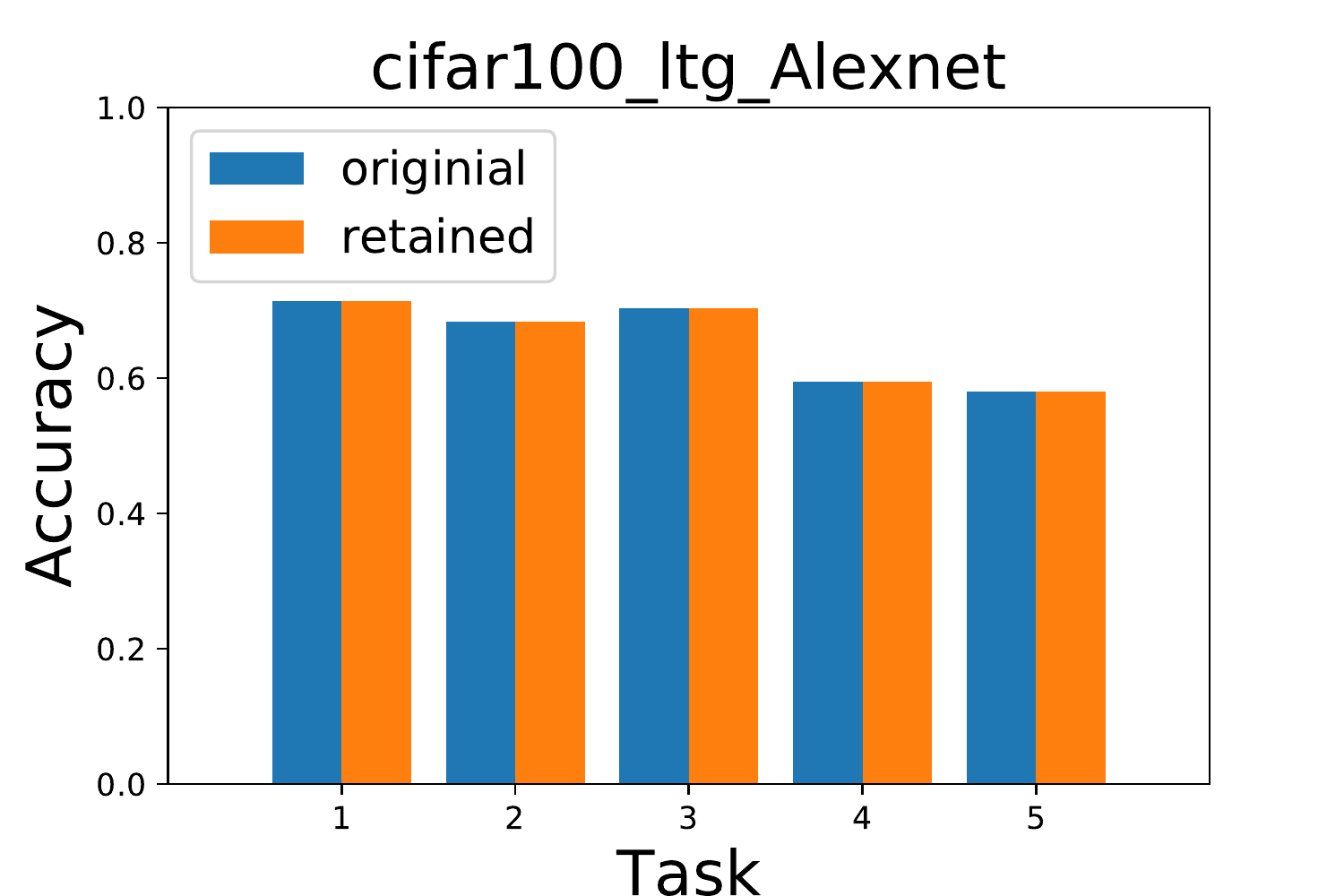}
        \end{minipage}
        \begin{minipage}[]{0.2\linewidth}
        \centering
        \includegraphics[width=1\linewidth]{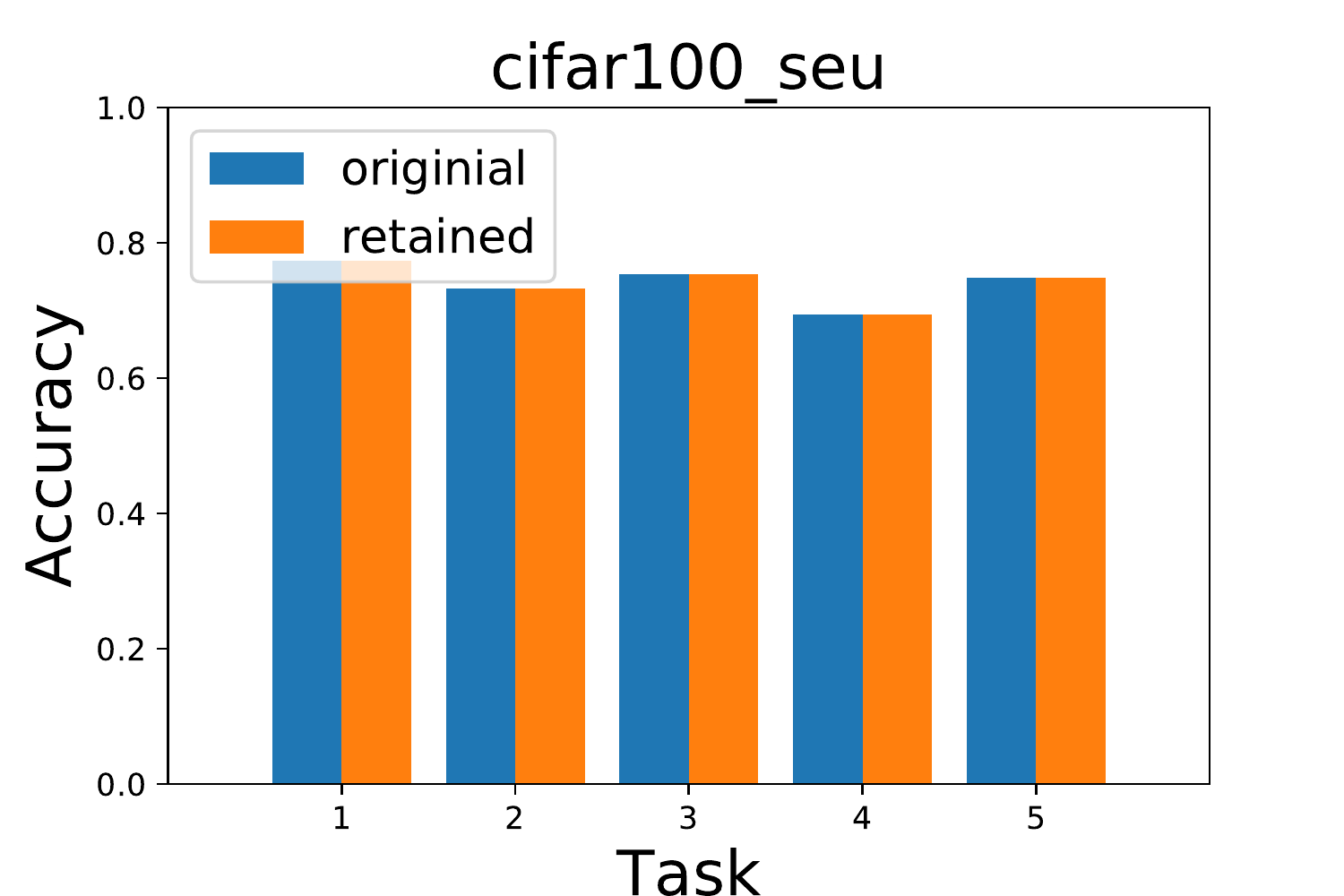}
        \end{minipage}
        \label{exp_acc_change_cifar100}
    }
    
    \centering
    \caption{Accuracy curve and accuracy change on the split CIFAR10 and CIFAR100.}
    \label{exp_acc_cifar}
\end{figure*}

\begin{figure*}[ht]
    \subfloat[The accuracy curve of each task from the beginning learning to the completion of all tasks on the PMNIST.]{
    \begin{minipage}[]{0.2\linewidth}
    \centering
    \includegraphics[width=1\linewidth]{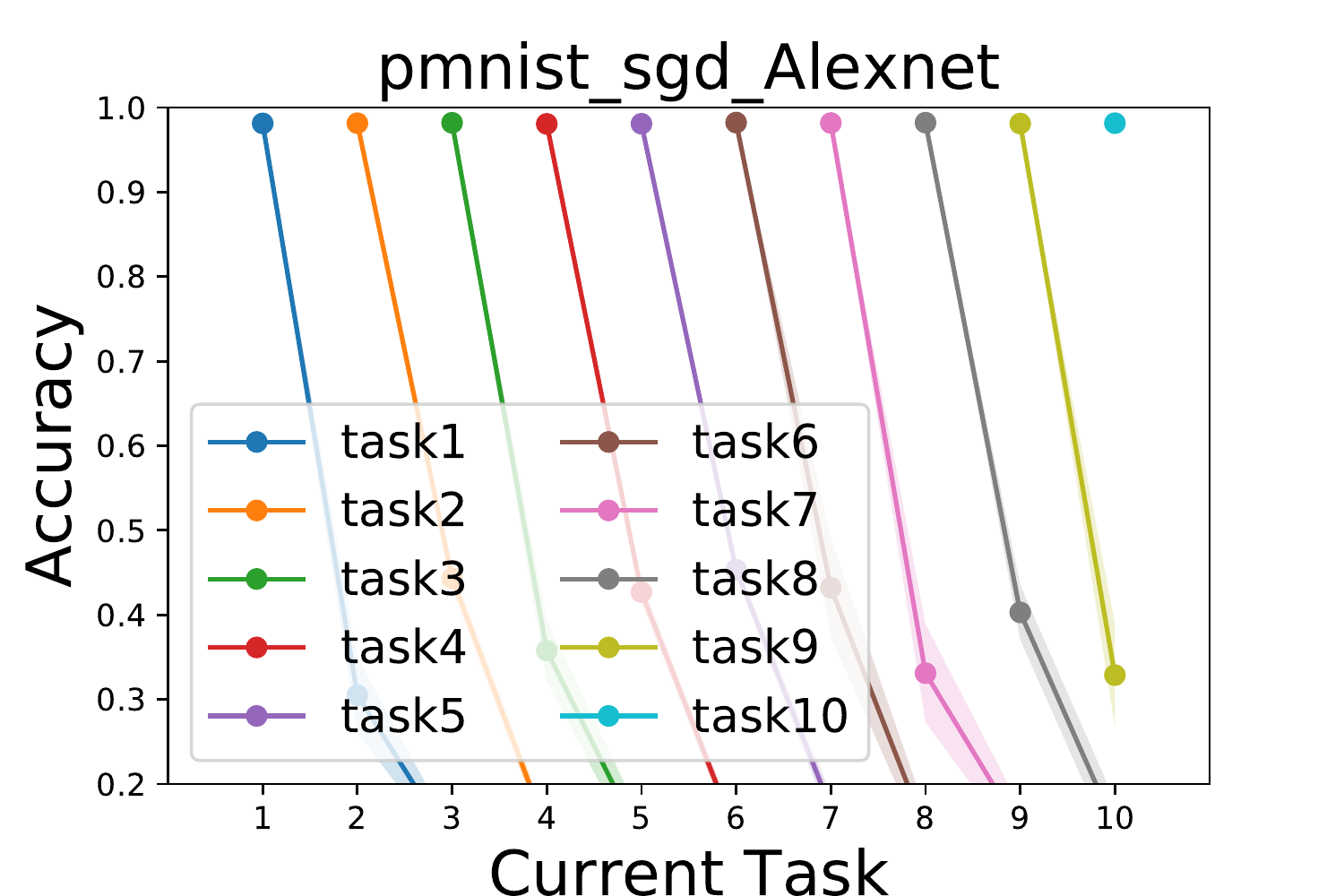}
    \end{minipage}
    \begin{minipage}[]{0.2\linewidth}
    \centering
    \includegraphics[width=1\linewidth]{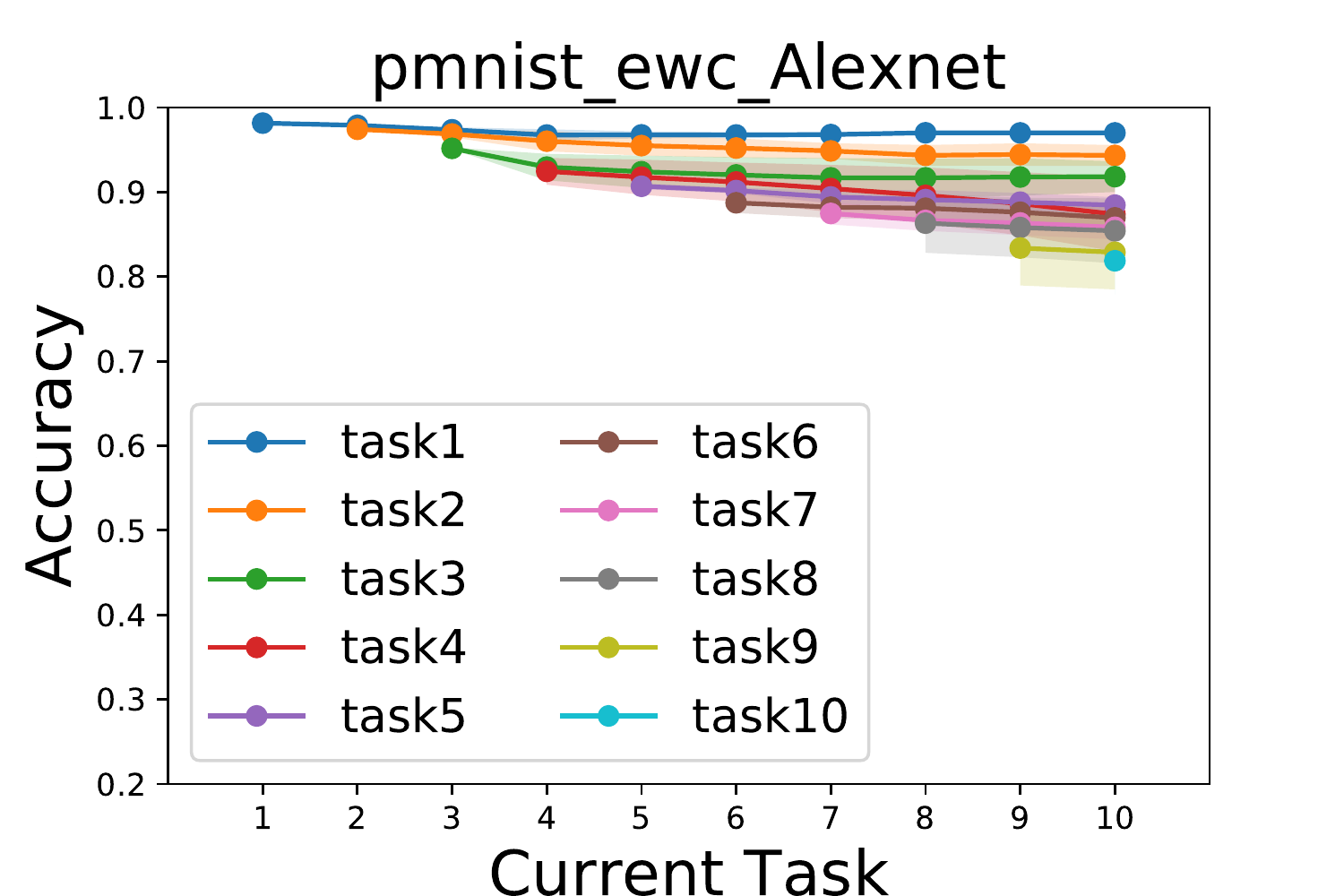}
    \end{minipage}%
    \begin{minipage}[]{0.2\linewidth}
    \centering
    \includegraphics[width=1\linewidth]{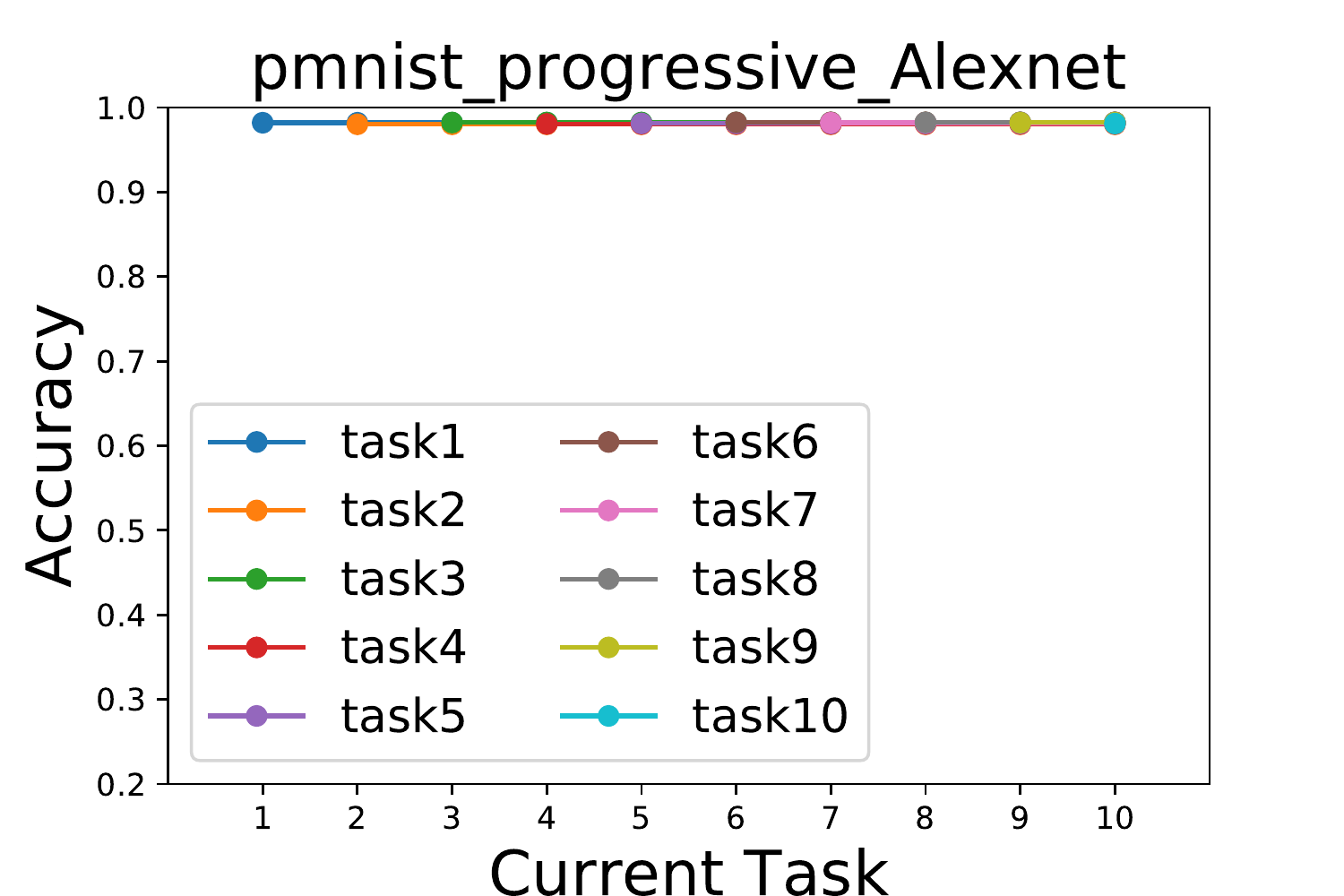}
    \end{minipage}
    \begin{minipage}[]{0.2\linewidth}
    \centering
    \includegraphics[width=1\linewidth]{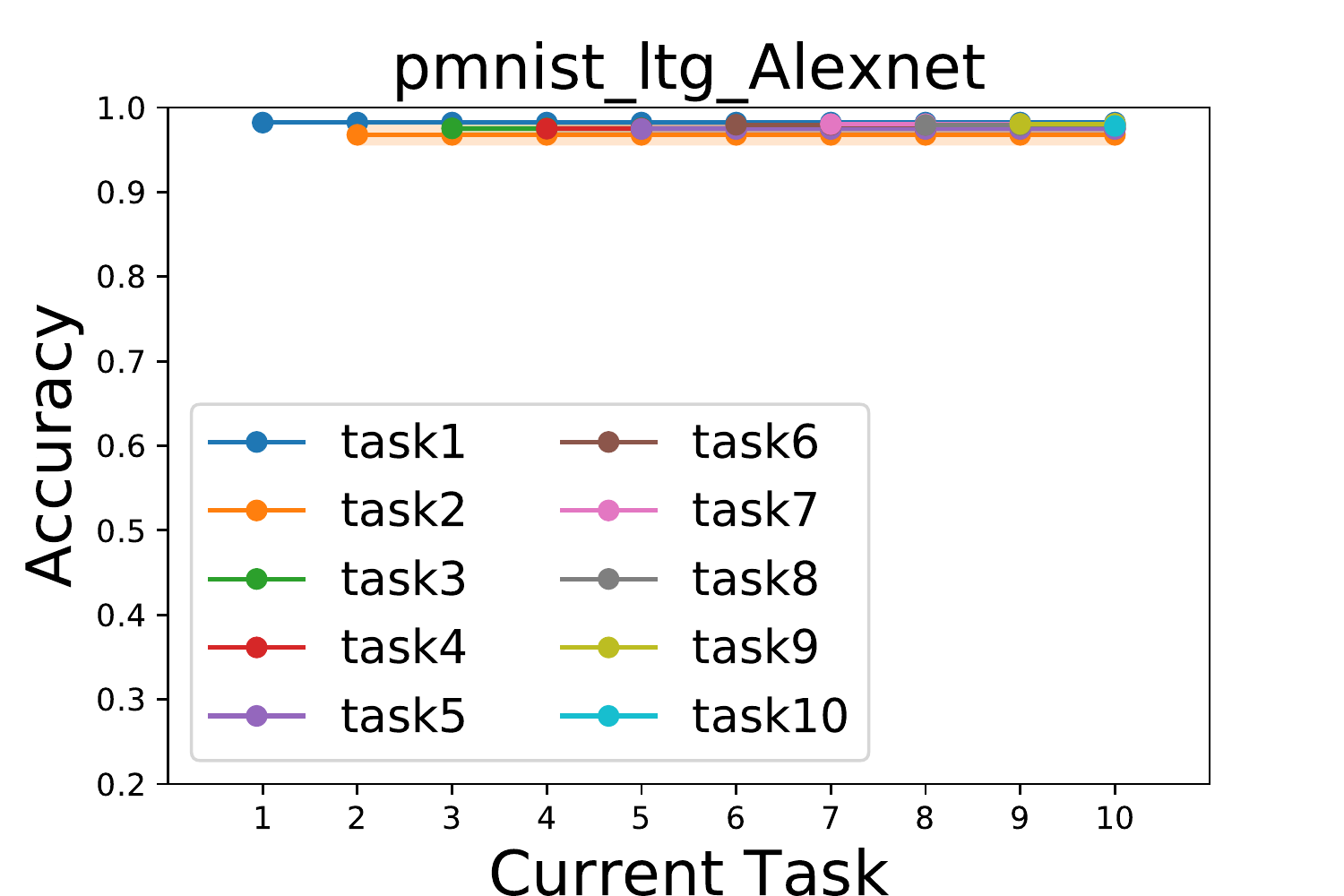}
    \end{minipage}
    \begin{minipage}[]{0.2\linewidth}
    \centering
    \includegraphics[width=1\linewidth]{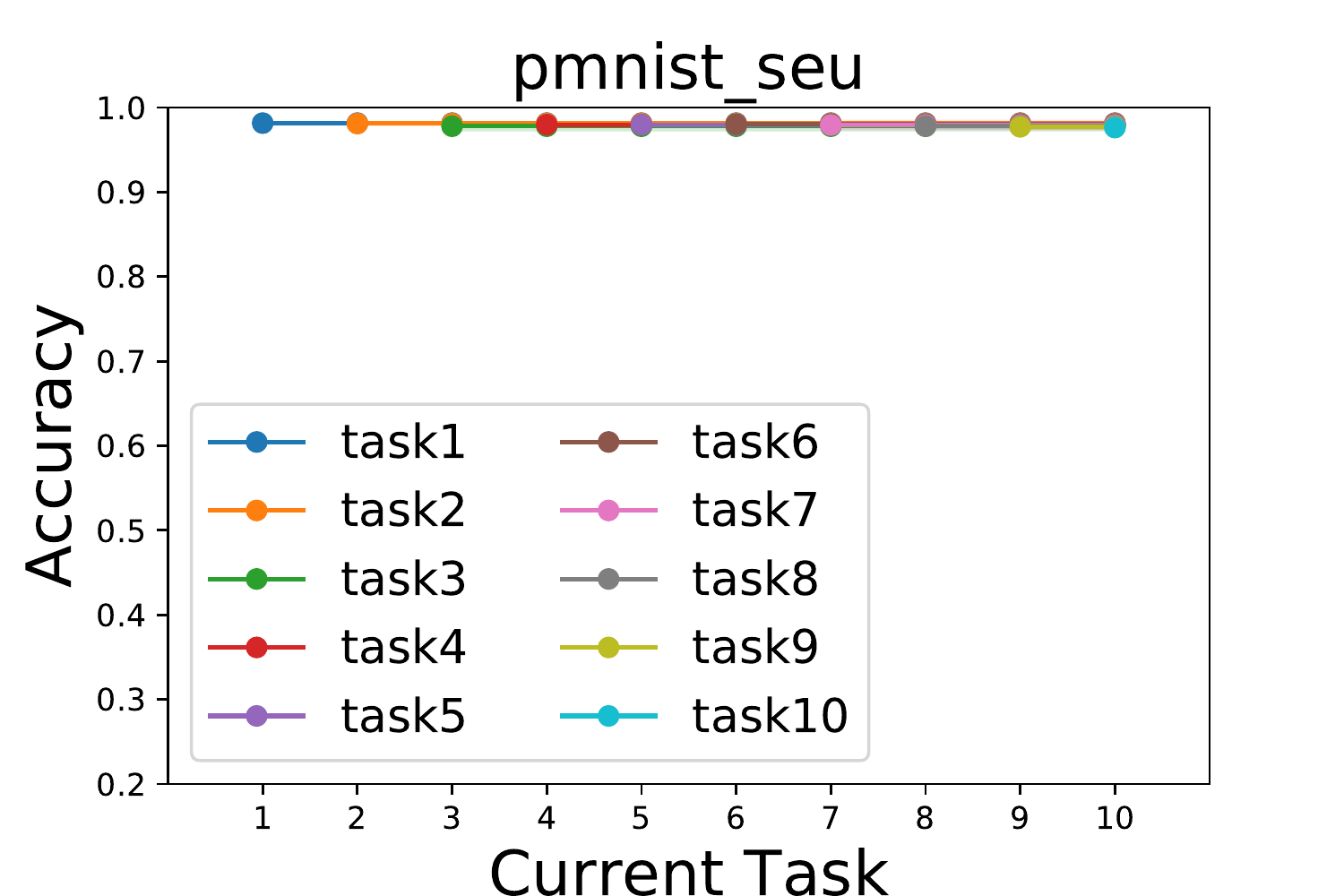}
    \end{minipage}
    \label{exp_acc_curve_pmnist}
}

    \subfloat[The original accuracy at the beginning and the retained accuracy after completing all tasks of each task on the PMNIST.]{
    \begin{minipage}[]{0.2\linewidth}
    \centering
    \includegraphics[width=1\linewidth]{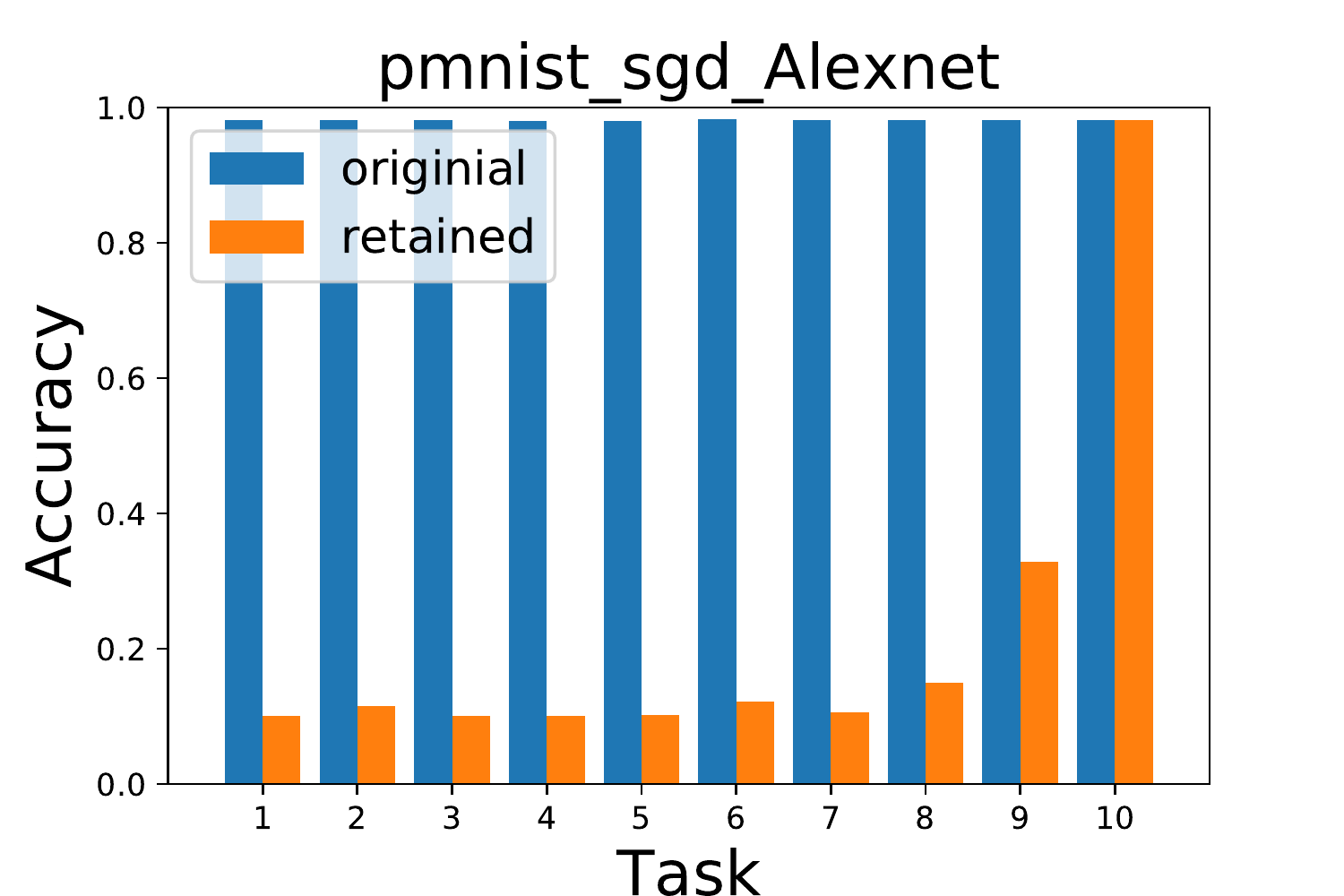}
    \end{minipage}
    \begin{minipage}[]{0.2\linewidth}
    \centering
    \includegraphics[width=1\linewidth]{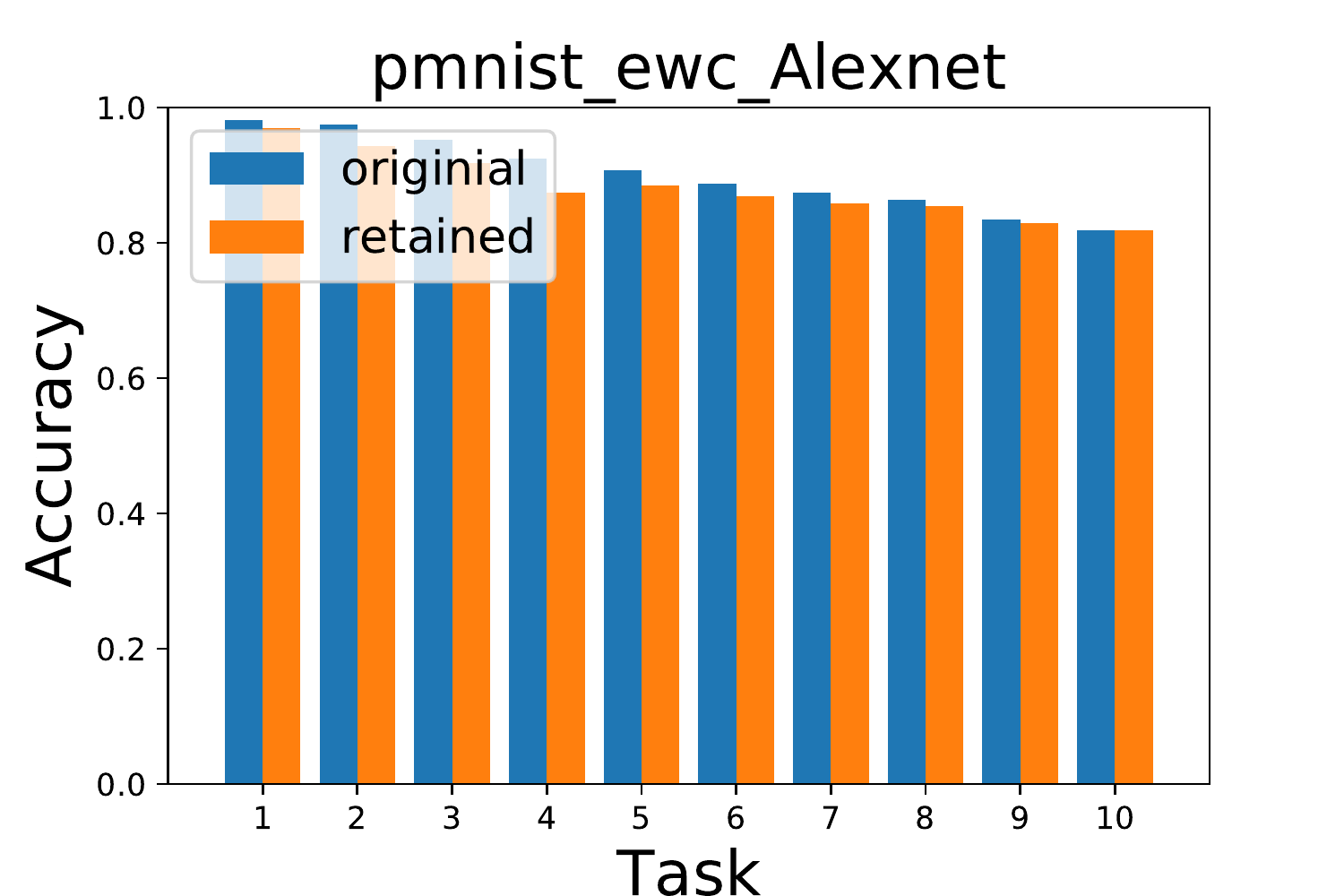}
    \end{minipage}%
    \begin{minipage}[]{0.2\linewidth}
    \centering
    \includegraphics[width=1\linewidth]{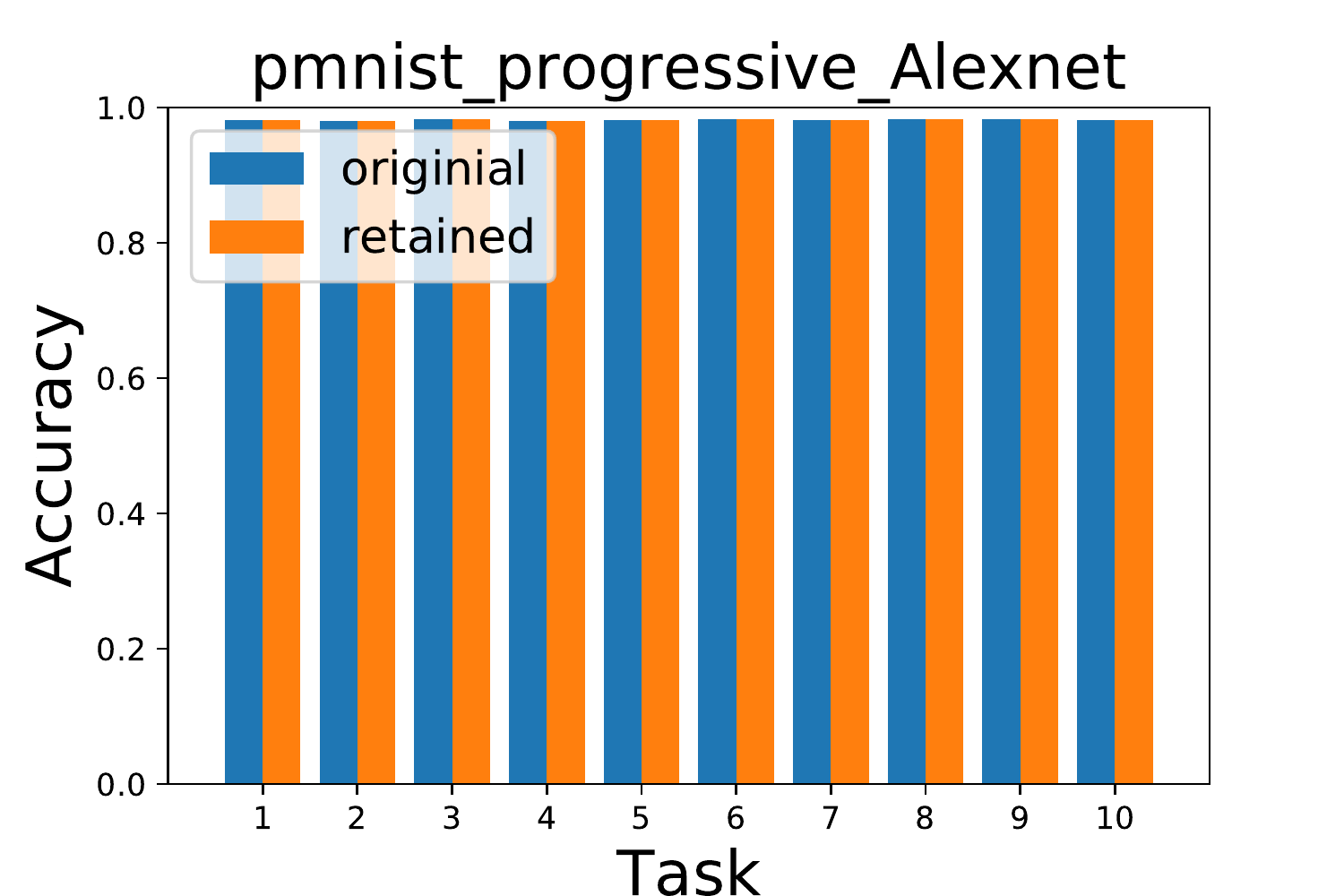}
    \end{minipage}
    \begin{minipage}[]{0.2\linewidth}
    \centering
    \includegraphics[width=1\linewidth]{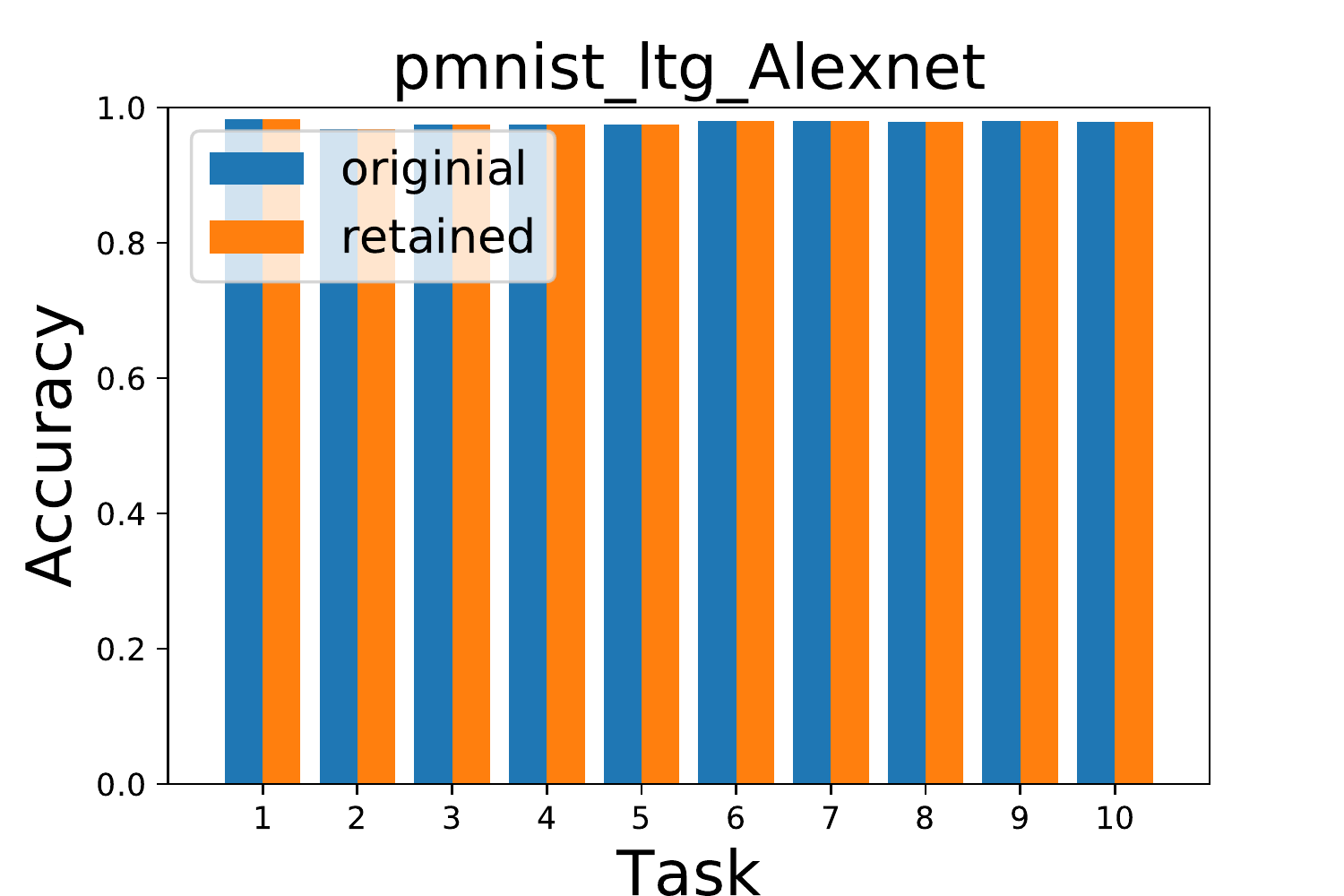}
    \end{minipage}
    \begin{minipage}[]{0.2\linewidth}
    \centering
    \includegraphics[width=1\linewidth]{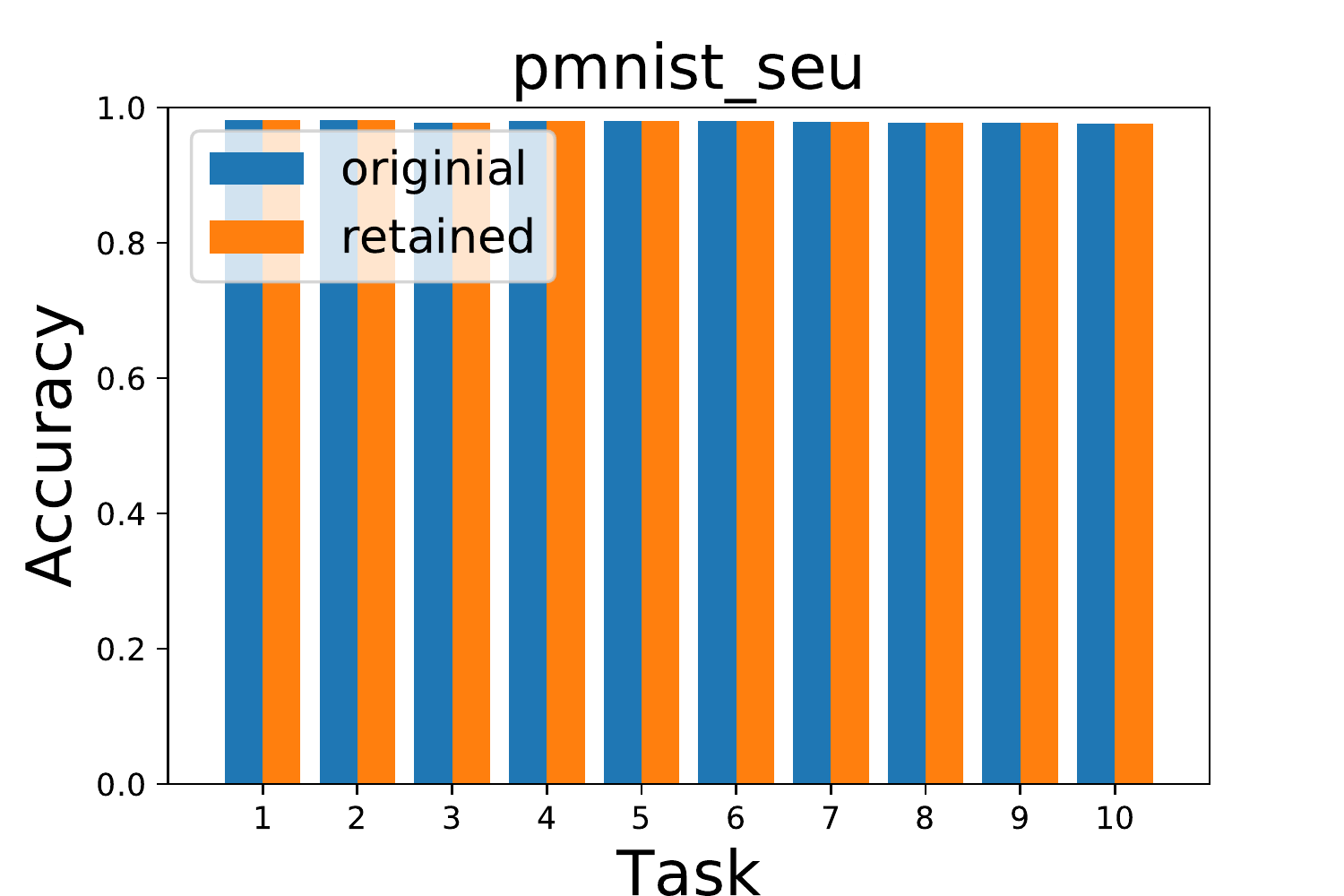}
    \end{minipage}
    \label{exp_acc_change_pmnist}
}

    \subfloat[The accuracy curve of each task from the beginning learning to the completion of all tasks on the Mixture.]{
    \begin{minipage}[]{0.2\linewidth}
    \centering
    \includegraphics[width=1\linewidth]{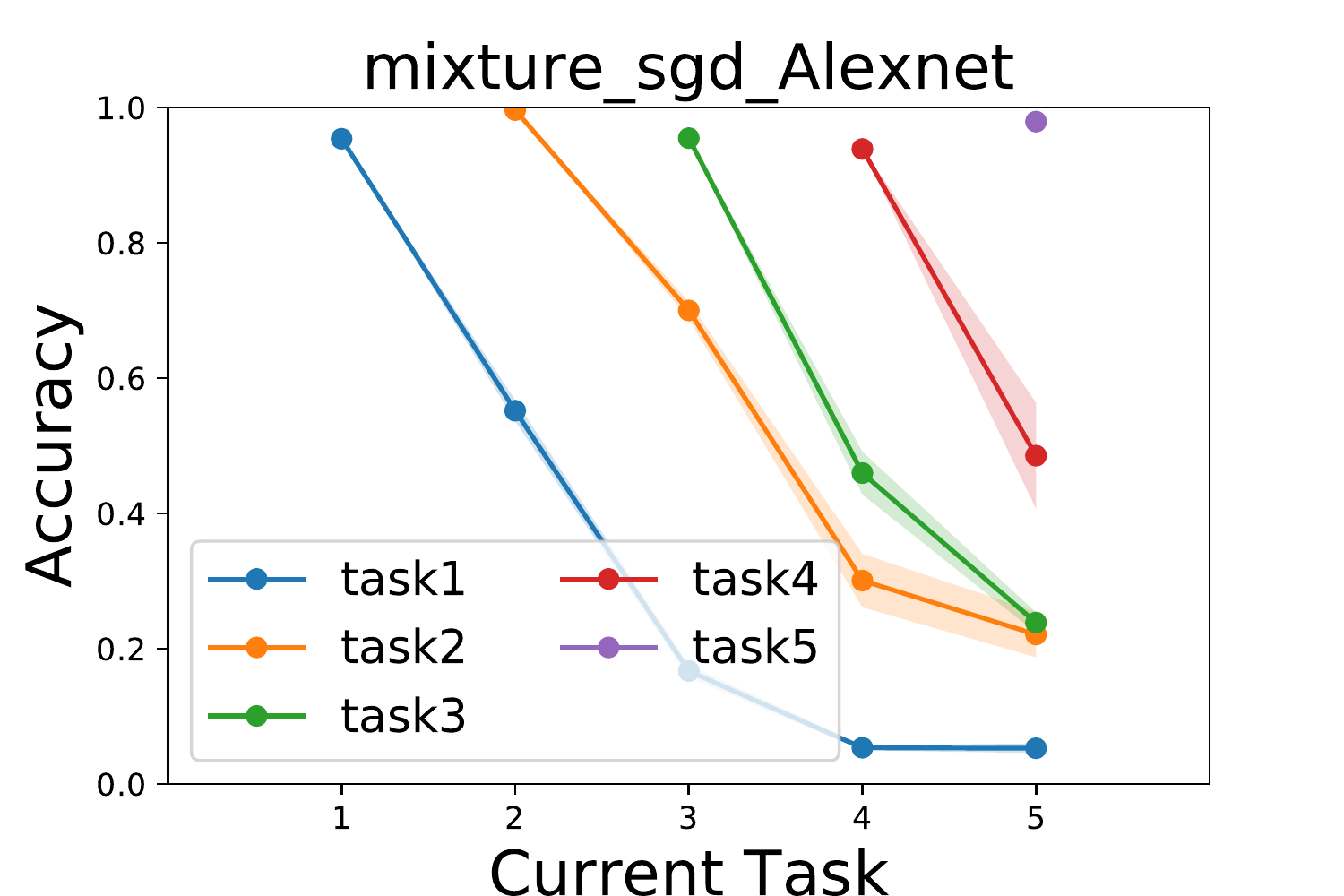}
    \end{minipage}
    \begin{minipage}[]{0.2\linewidth}
    \centering
    \includegraphics[width=1\linewidth]{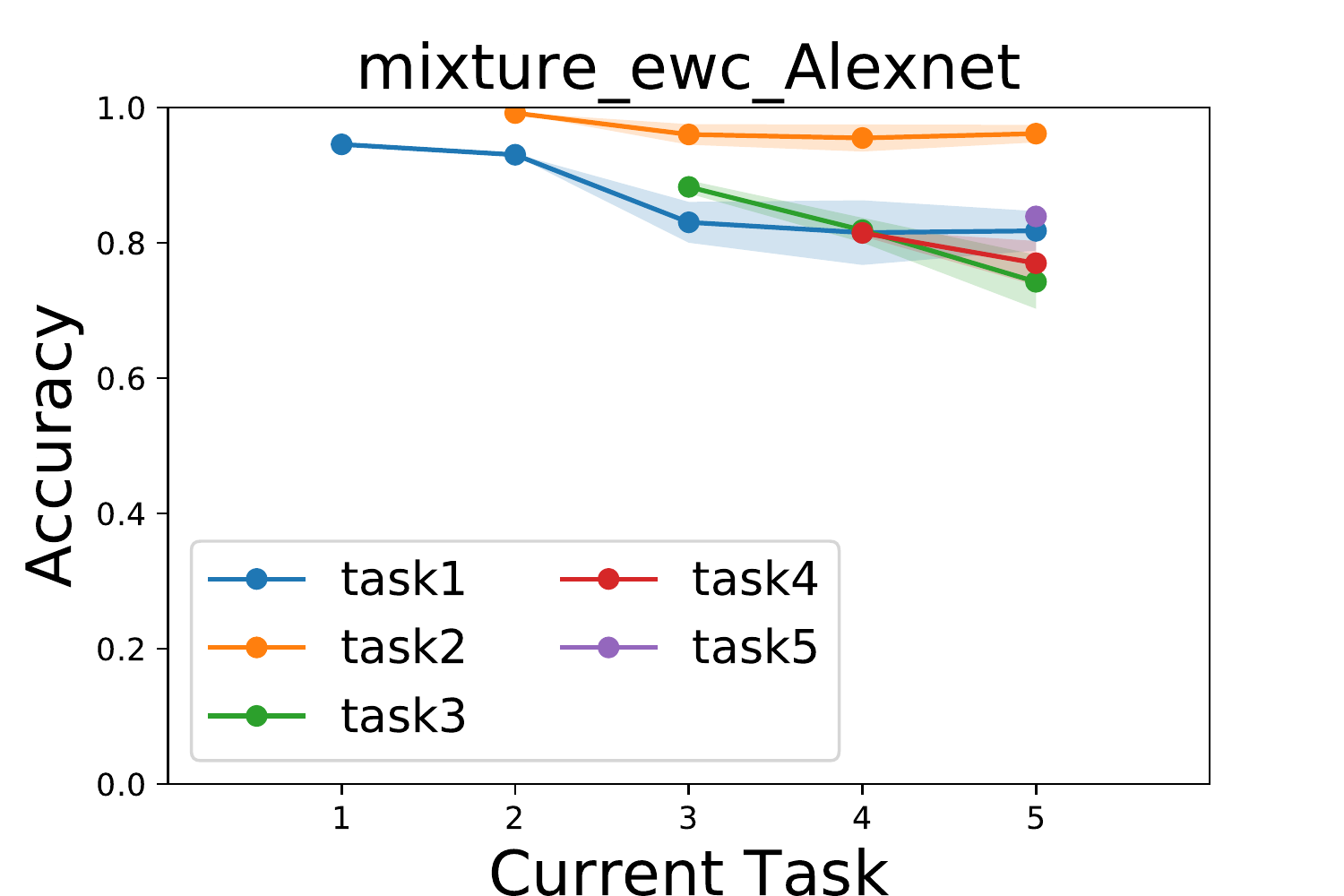}
    \end{minipage}%
    \begin{minipage}[]{0.2\linewidth}
    \centering
    \includegraphics[width=1\linewidth]{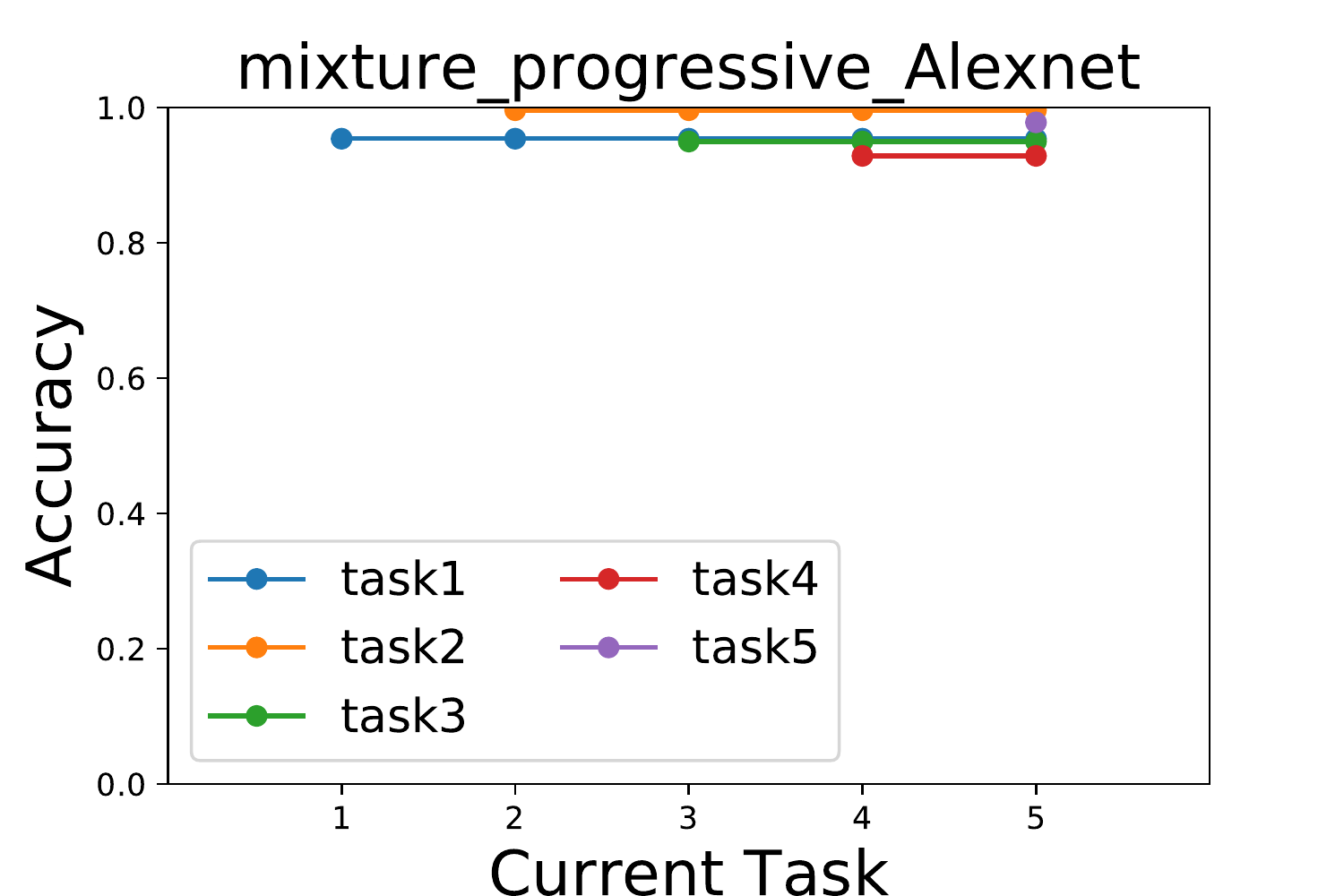}
    \end{minipage}
    \begin{minipage}[]{0.2\linewidth}
    \centering
    \includegraphics[width=1\linewidth]{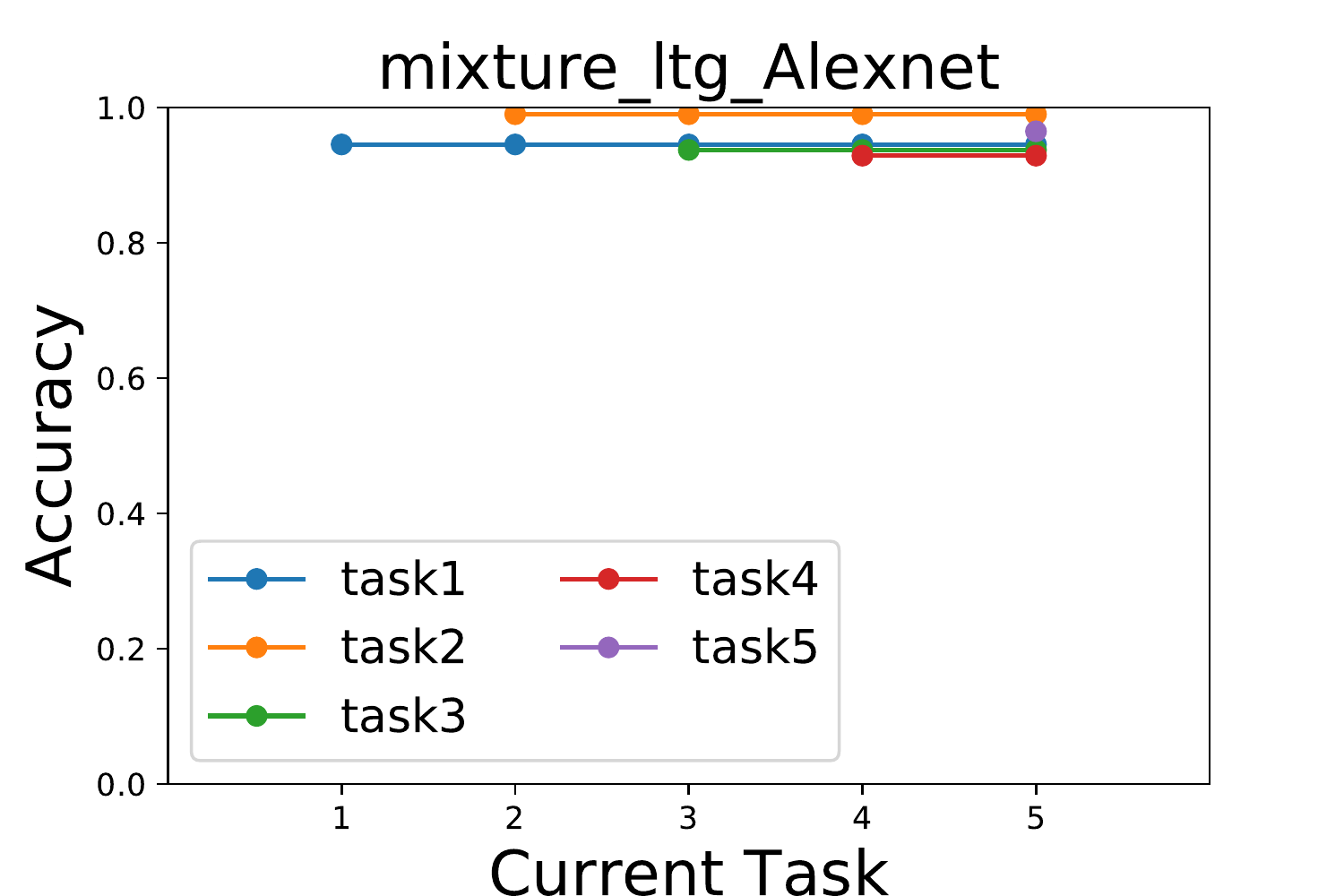}
    \end{minipage}
    \begin{minipage}[]{0.2\linewidth}
    \centering
    \includegraphics[width=1\linewidth]{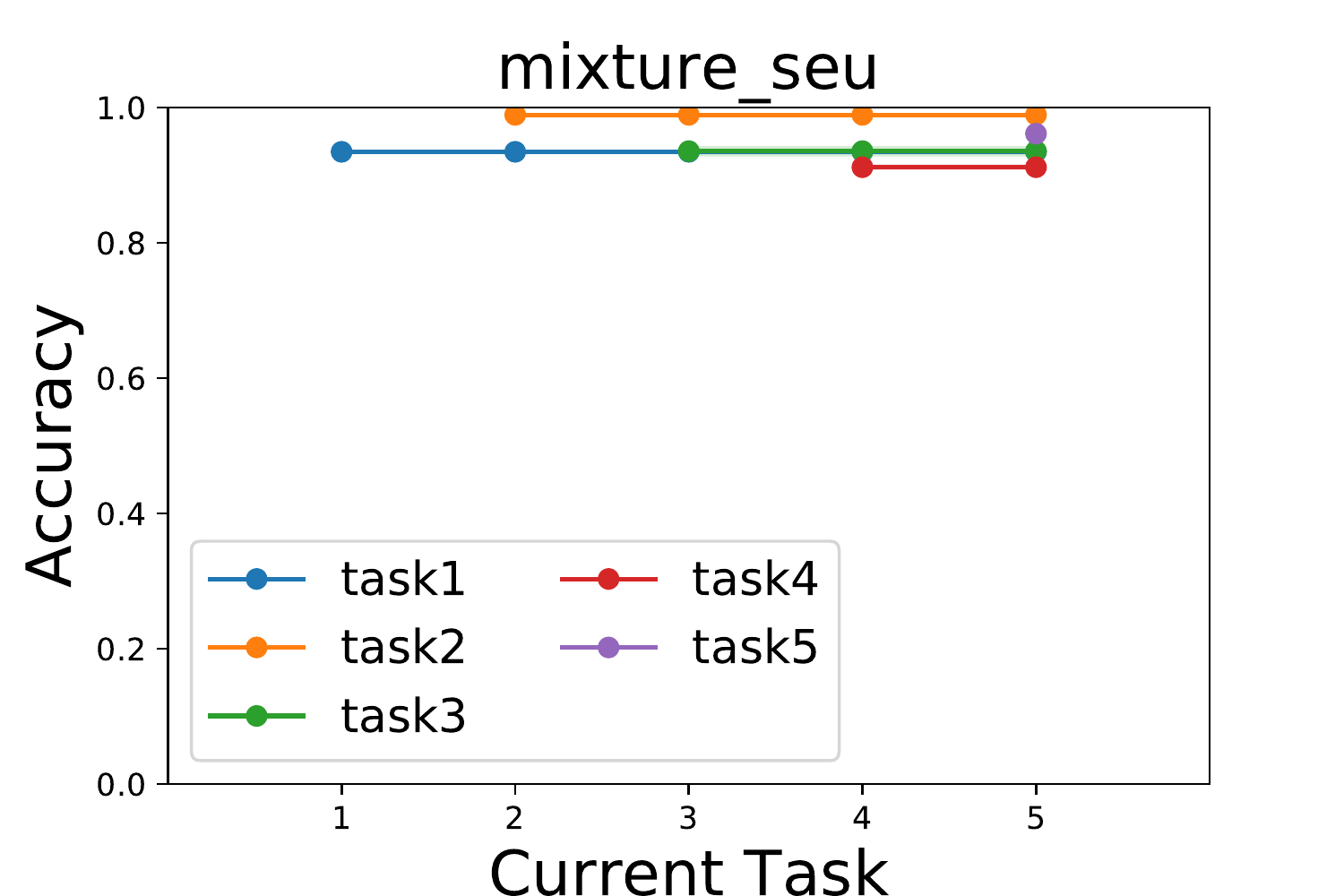}
    \end{minipage}
    \label{exp_acc_curve_mixture}
}

    \subfloat[The original accuracy at the beginning and the retained accuracy after completing all tasks of each task on the Mixture.]{
    \begin{minipage}[]{0.2\linewidth}
    \centering
    \includegraphics[width=1\linewidth]{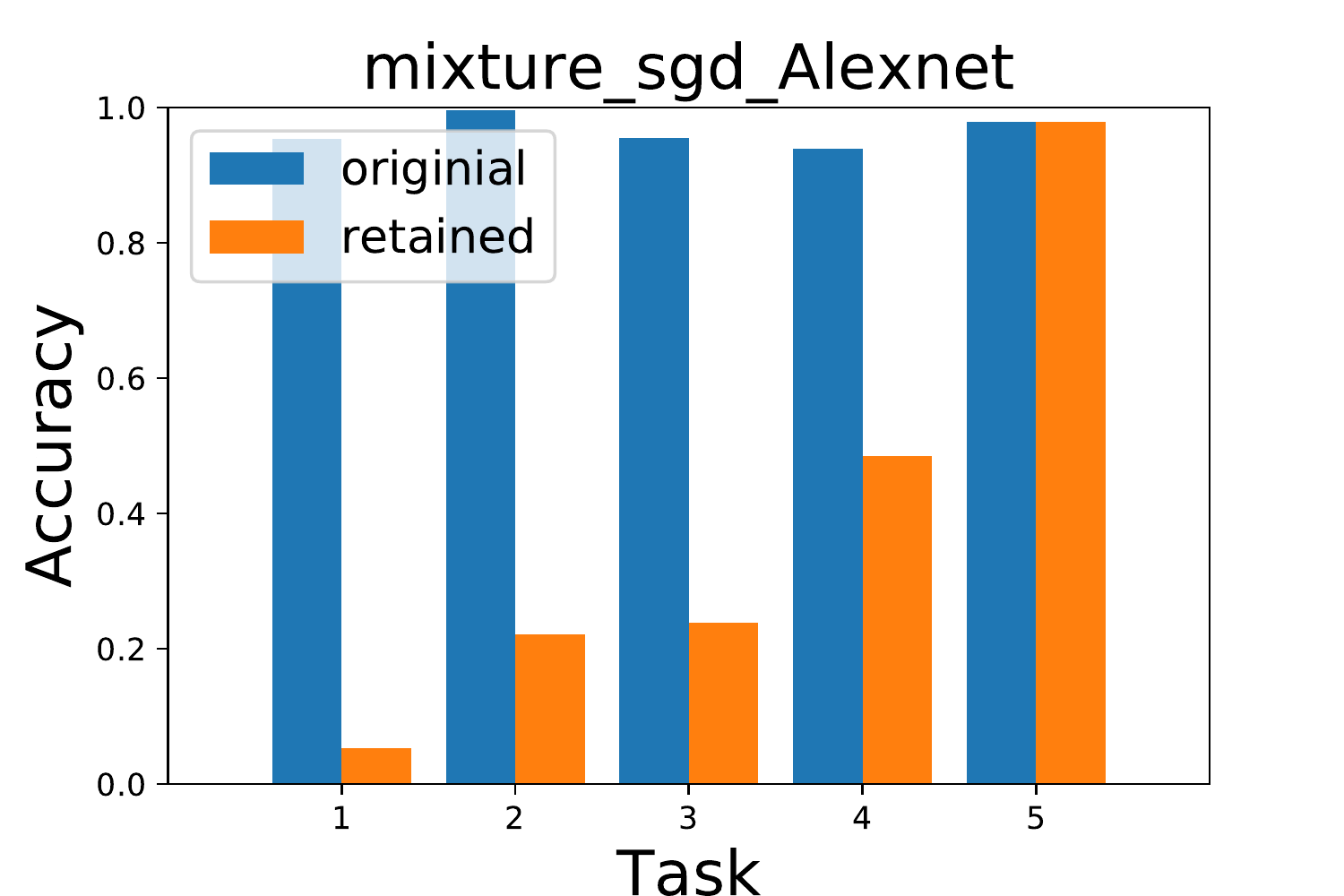}
    \end{minipage}
    \begin{minipage}[]{0.2\linewidth}
    \centering
    \includegraphics[width=1\linewidth]{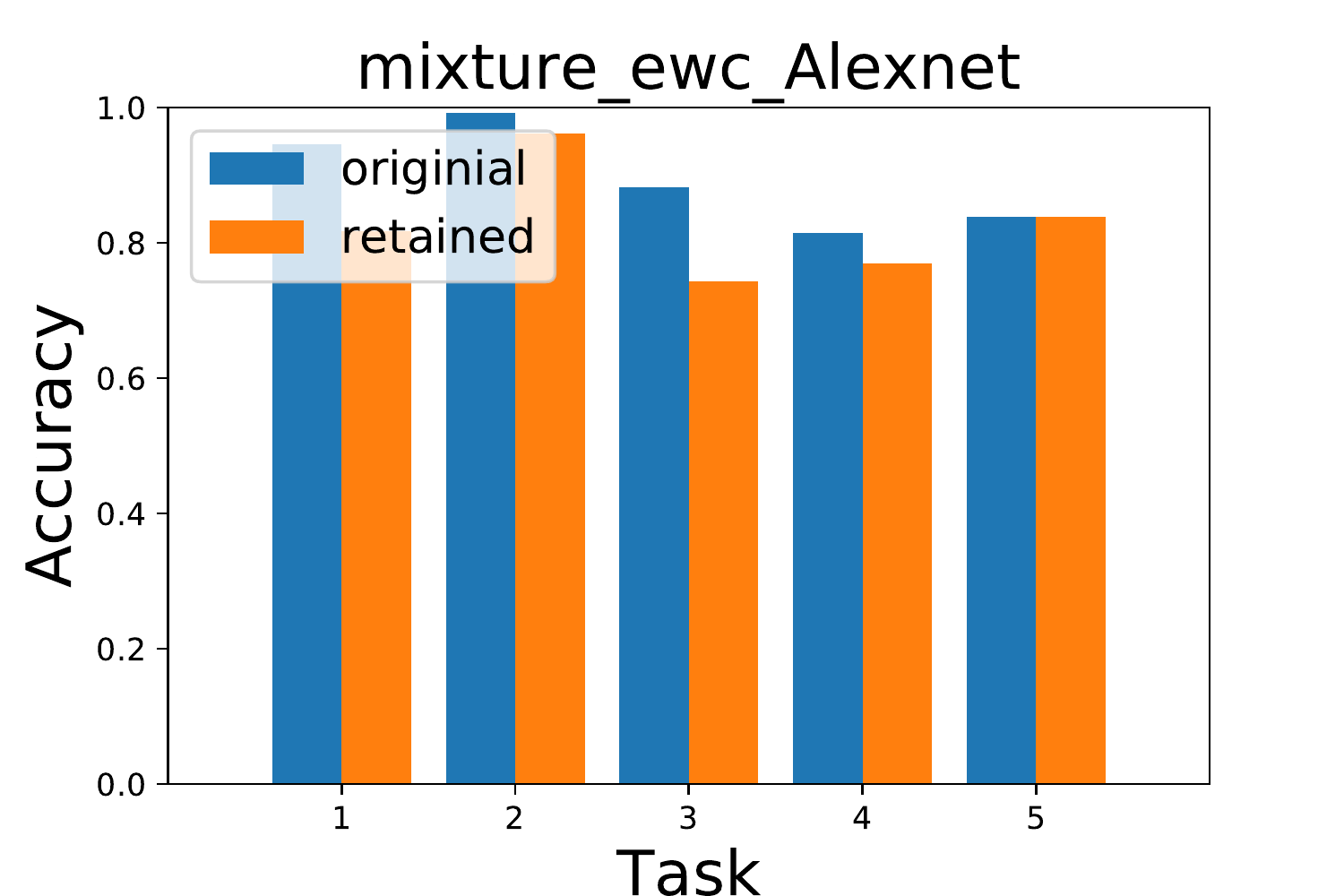}
    \end{minipage}%
    \begin{minipage}[]{0.2\linewidth}
    \centering
    \includegraphics[width=1\linewidth]{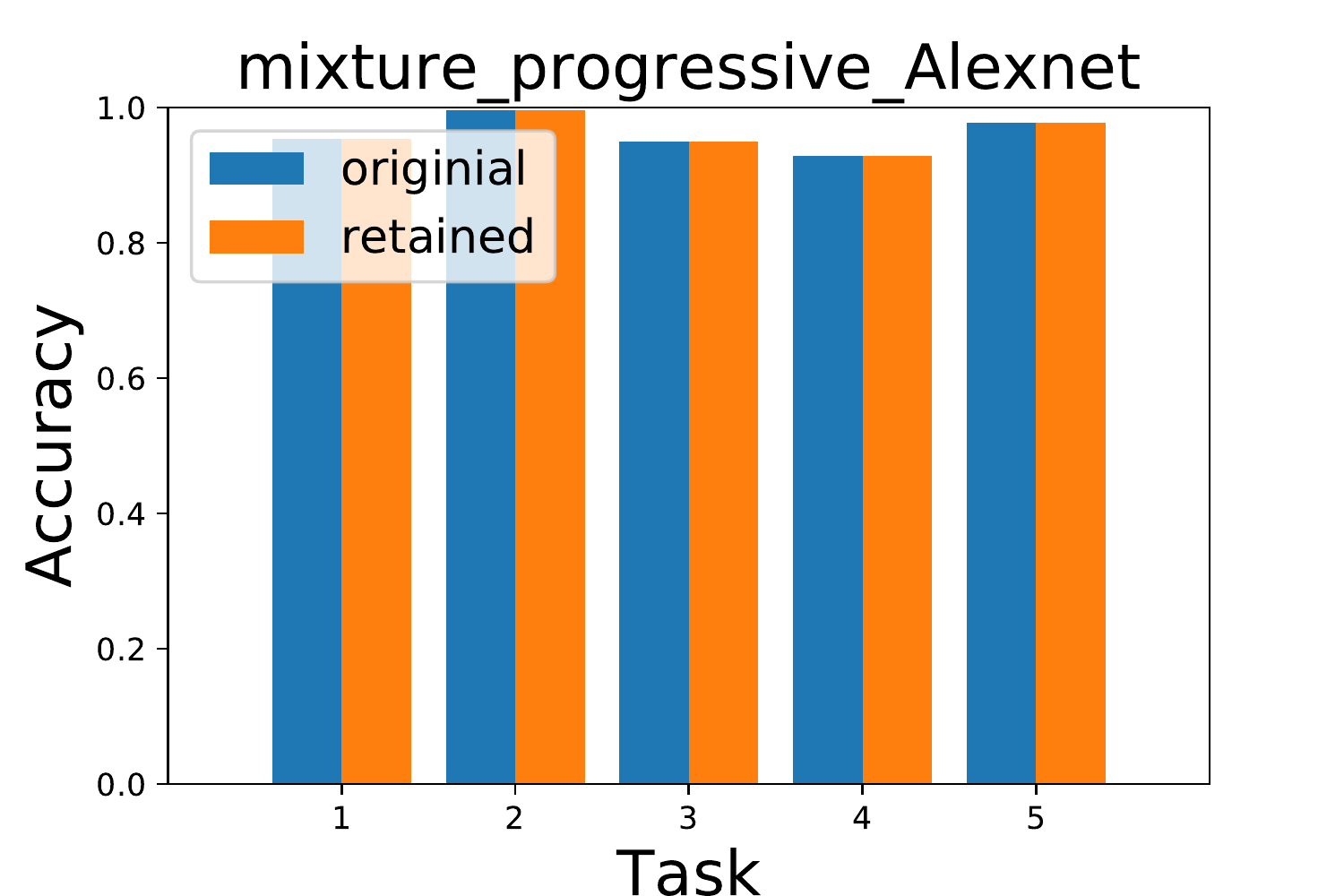}
    \end{minipage}
    \begin{minipage}[]{0.2\linewidth}
    \centering
    \includegraphics[width=1\linewidth]{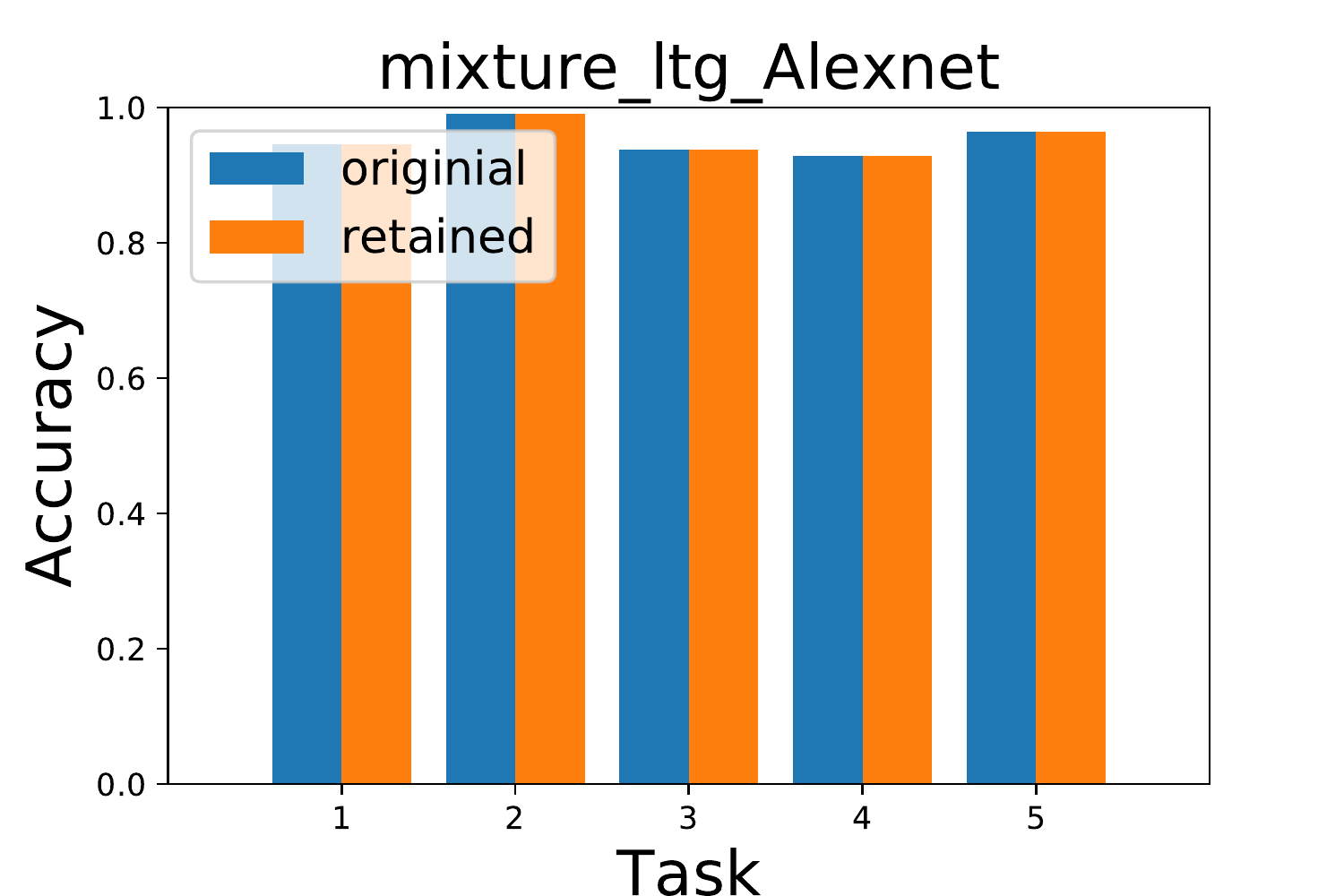}
    \end{minipage}
    \begin{minipage}[]{0.2\linewidth}
    \centering
    \includegraphics[width=1\linewidth]{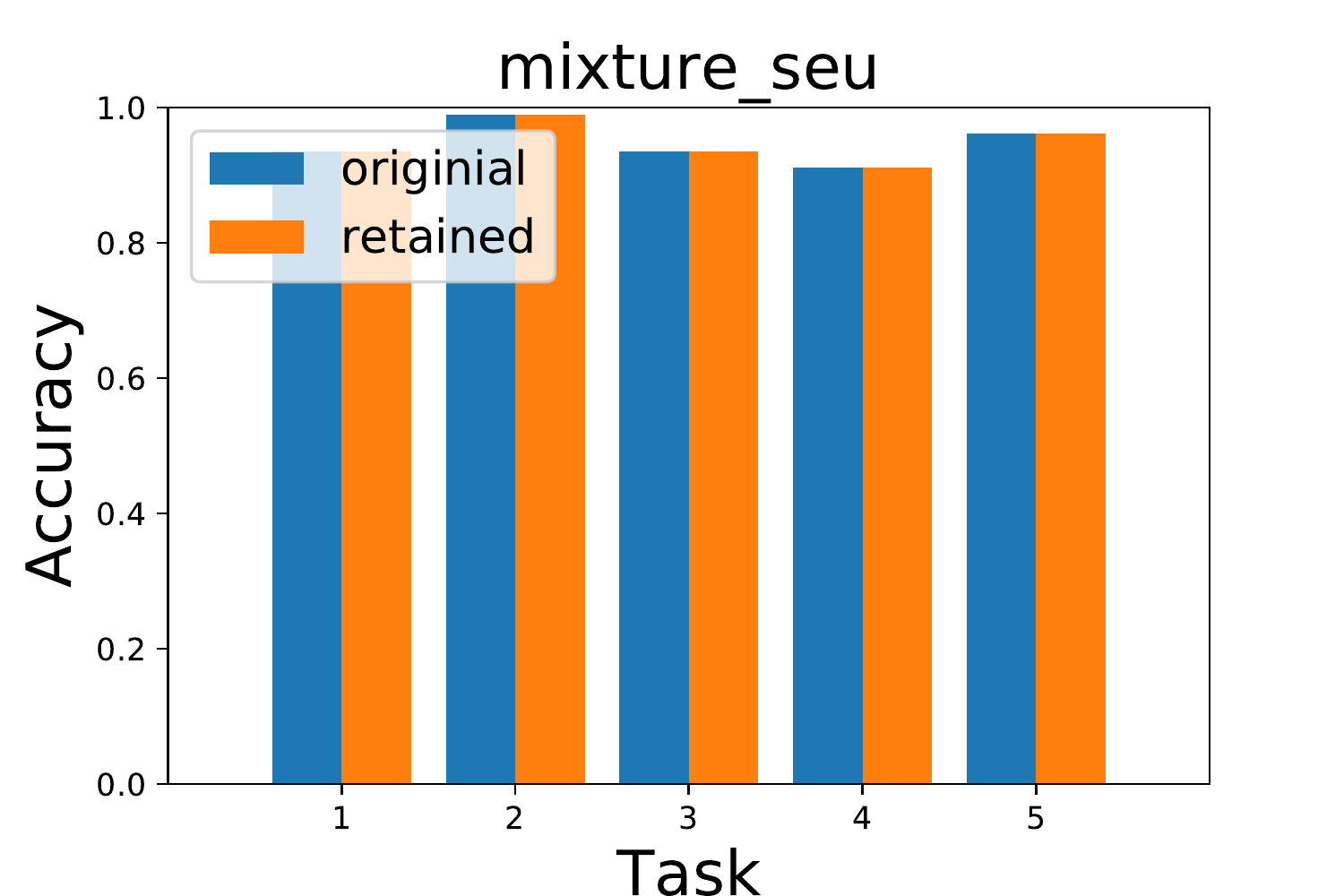}
    \end{minipage}
    \label{exp_acc_change_mixture}
}
    \centering
    \caption{Accuracy curve and accuracy change on the PMNIST and Mixture.}
    \label{exp_acc_pmnist_mixture}
\end{figure*}

To evaluate the ability of our method in overcoming catastrophic forgetting, we observe the model performance on all learned tasks after learning a new task each time.

We first look at performance on the split CIFAR10 (see \figurename{\ref{exp_acc_curve_cifar10},\ref{exp_acc_change_cifar10}}). After the first learning, SGD on task 1 is 92.02\%. But after learning task 2, the accuracy on task 1 is only 57.03\%, although the accuracy on task 2 is 96.79\%. Similarly, the accuracy on task 2 drops to 68.65\% after learning task 3. This show that without any constraints, when learning a new task, the model quickly loses a lot of the knowledge learned from previous tasks. What's more, as new tasks coming along, the loss of knowledge learned from previous tasks will accumulate. After completing all tasks, the accuracy on task 1 drops from 92.02\% to 44.24\% and on task 2 drops from 96.79\% to 22.11\%. EWC mitigates the rapid loss of knowledge learned from previous task. After learning task 2, the accuracy on task 1 drops from 91.84\% to 90.30\%. And it drops to 87.17\% after learning task 3. After completing all tasks, the accuracy on task 1 drops from 91.84\% to 83.73\% and on task 2 drops from 70.11\% to 68.96\%. SEU, ltg, and Progressive Network all succeed in completely avoiding the catastrophic forgetting problem. After learning a new task, the accuracy of the model on previous tasks do not change and all the knowledge is retained. 

The performance on the split CIAFR100 is similar (see \figurename{\ref{exp_acc_curve_cifar100},\ref{exp_acc_change_cifar100}}). The accuracy of SGD drops from 70.70\% to 30.15\% on task 1 and from 71.97\% to 38.96\% on task 2 after completing all tasks. SEU, ltg, and Progressive Network also implement zero forgetting whose accuracy curves are flat. EWC almost avoids forgetting and its accuracy curve is already relatively flat. We also perform the same experiment on PMNIST and Mixture datasets, and achieve similar performances. The experimental results are shown in \figurename{\ref{exp_acc_pmnist_mixture}}.

\subsection{Average Performance}
\begin{figure*}[!t]
    \subfloat[]{\includegraphics[width=0.25\linewidth]{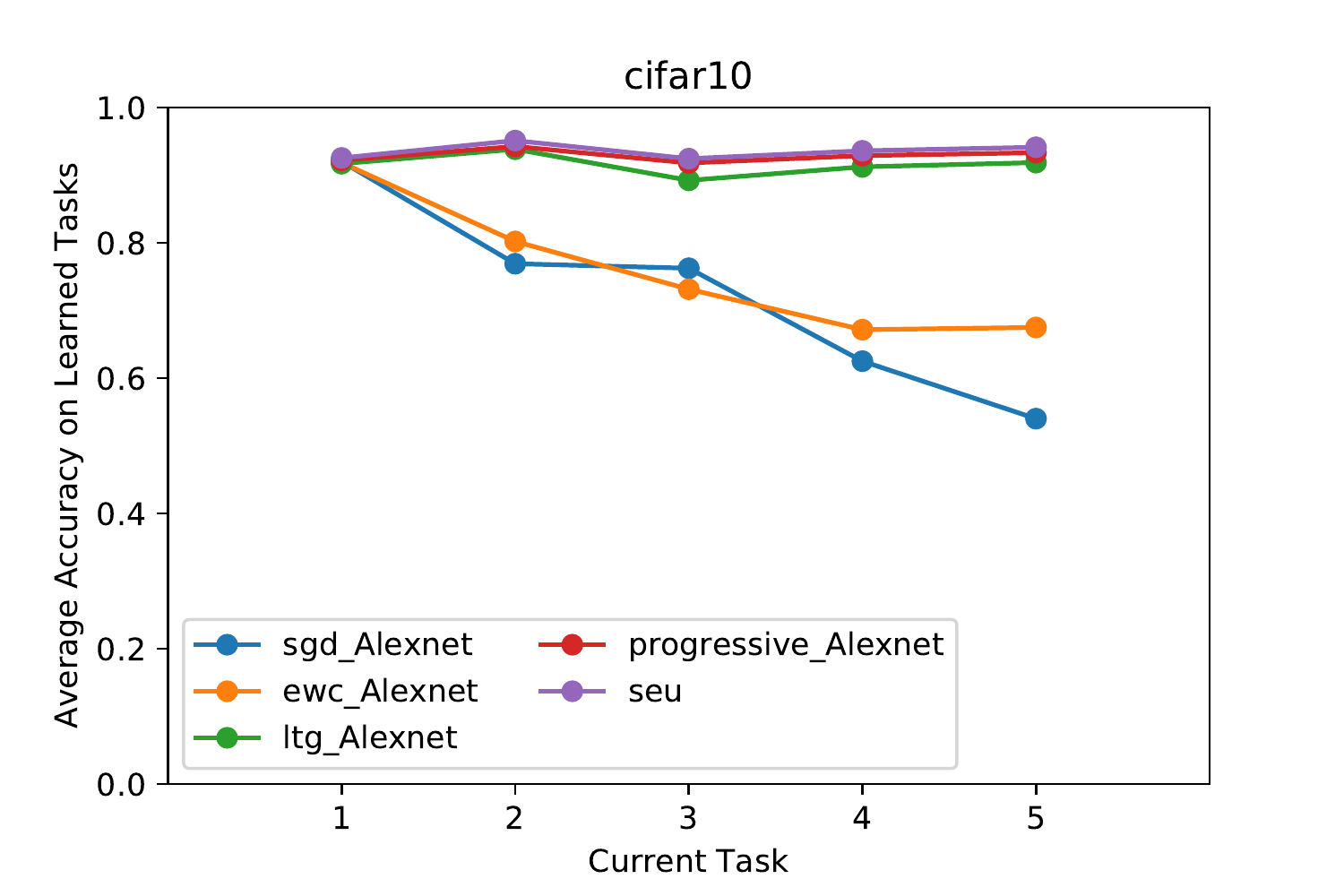}%
    \label{exp_avg_acc_cifar10}}
    \hfil
    \subfloat[]{\includegraphics[width=0.25\linewidth]{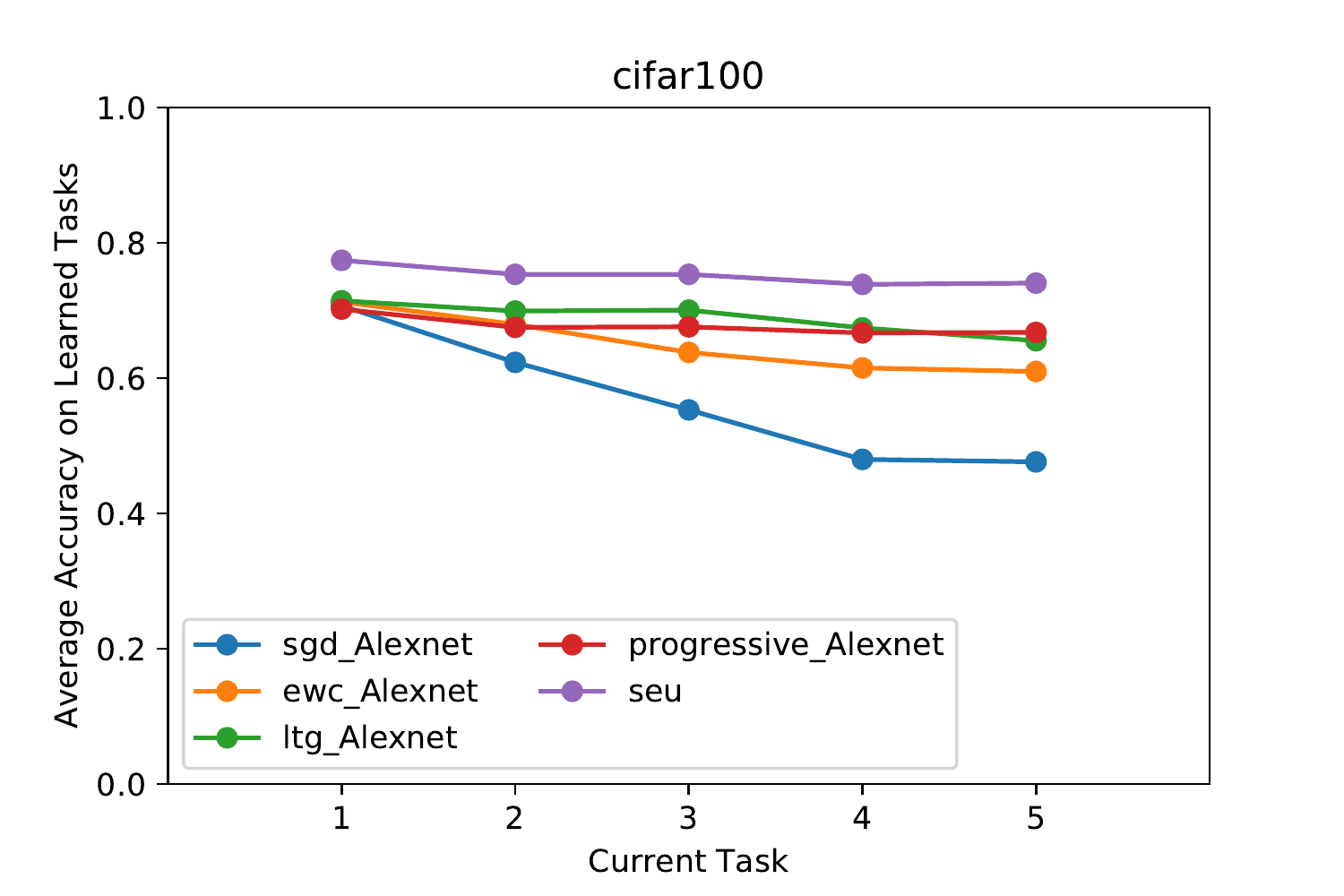}%
    \label{exp_avg_acc_cifar100}}
    \hfil
    \subfloat[]{\includegraphics[width=0.25\linewidth]{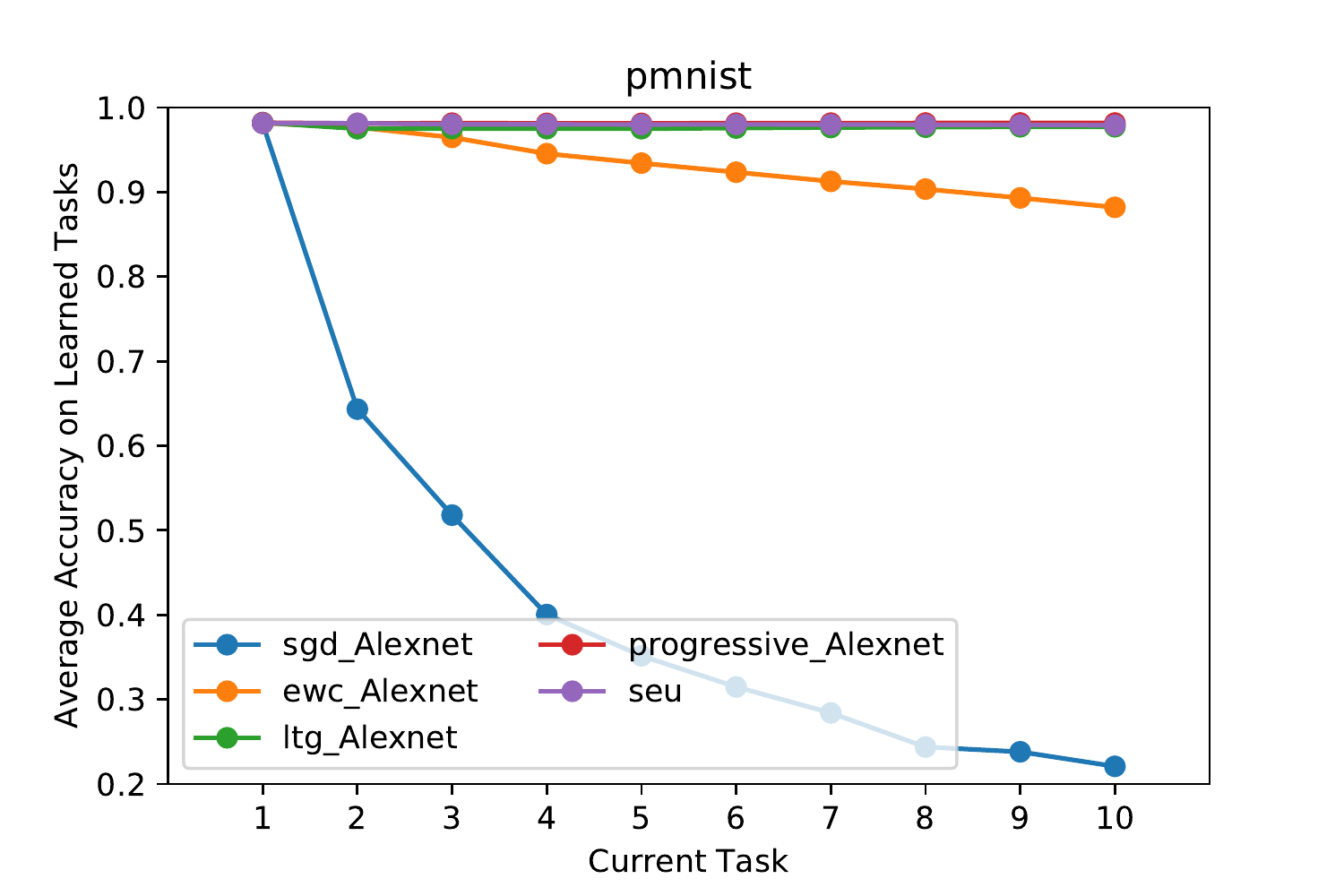}%
    \label{exp_avg_acc_pmnist}}
    \hfil
    \subfloat[]{\includegraphics[width=0.25\linewidth]{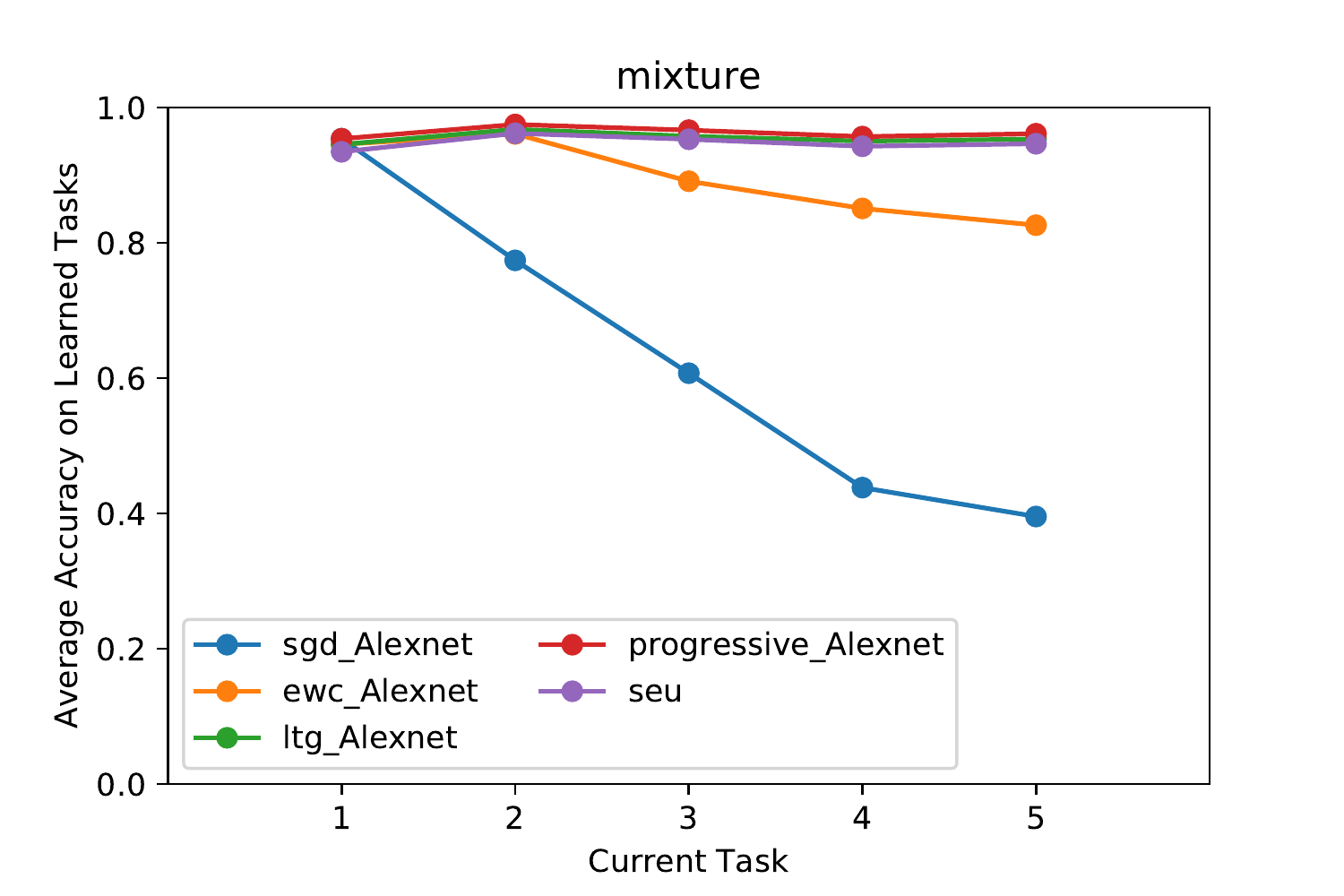}%
    \label{exp_avg_acc_mixture}}

    \centering
    \caption{Average accuracy of all learned tasks after learning current task. (a) CIFAR10 (b) CIFAR100 (c) PMNIST (d) Mixture}
    \label{exp_avg_acc}
\end{figure*}

\begin{figure*}[!t]
    \subfloat[]{
    \begin{minipage}[]{0.2\linewidth}
    \centering
    \includegraphics[width=1\linewidth]{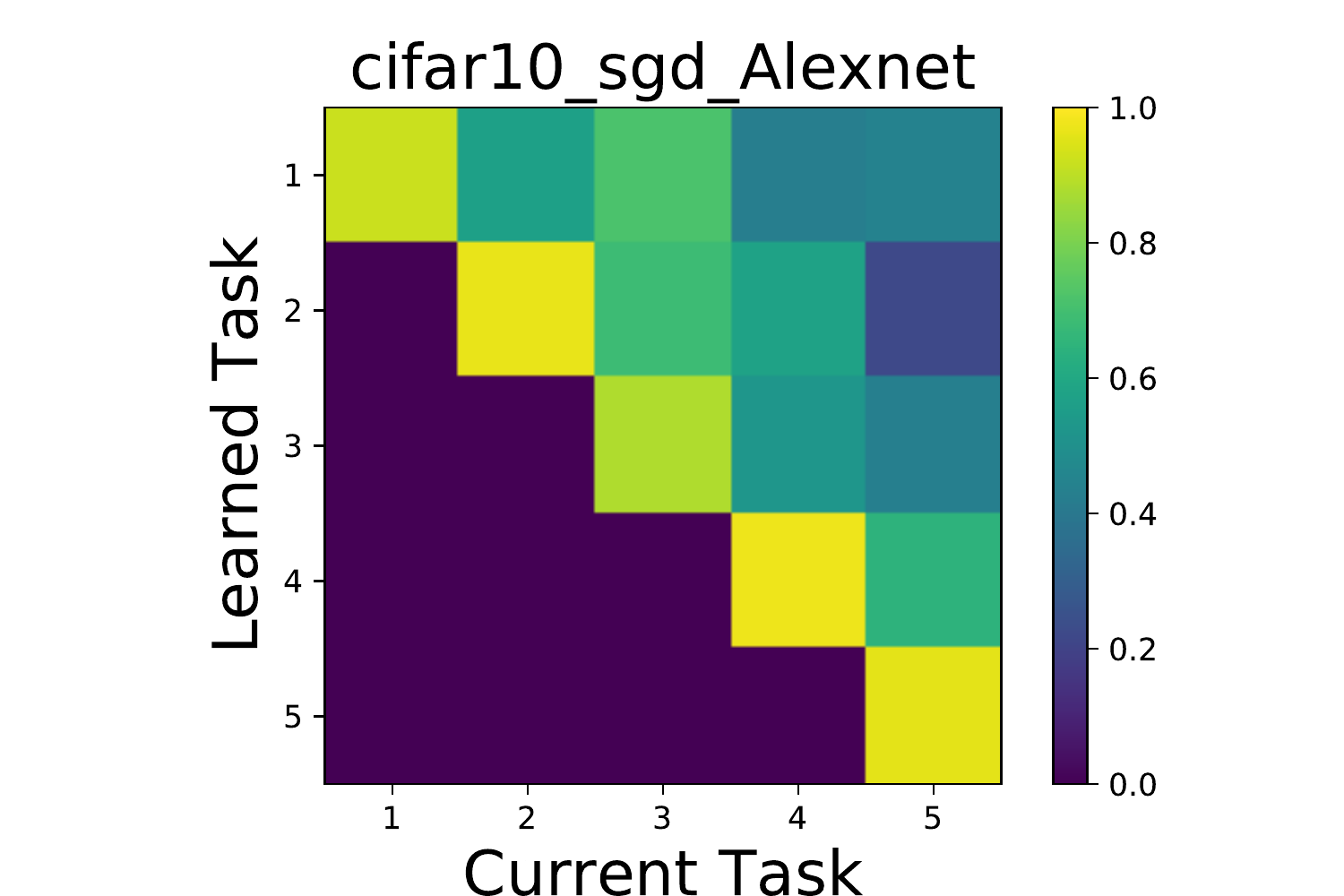}
    \end{minipage}
    \begin{minipage}[]{0.2\linewidth}
    \centering
    \includegraphics[width=1\linewidth]{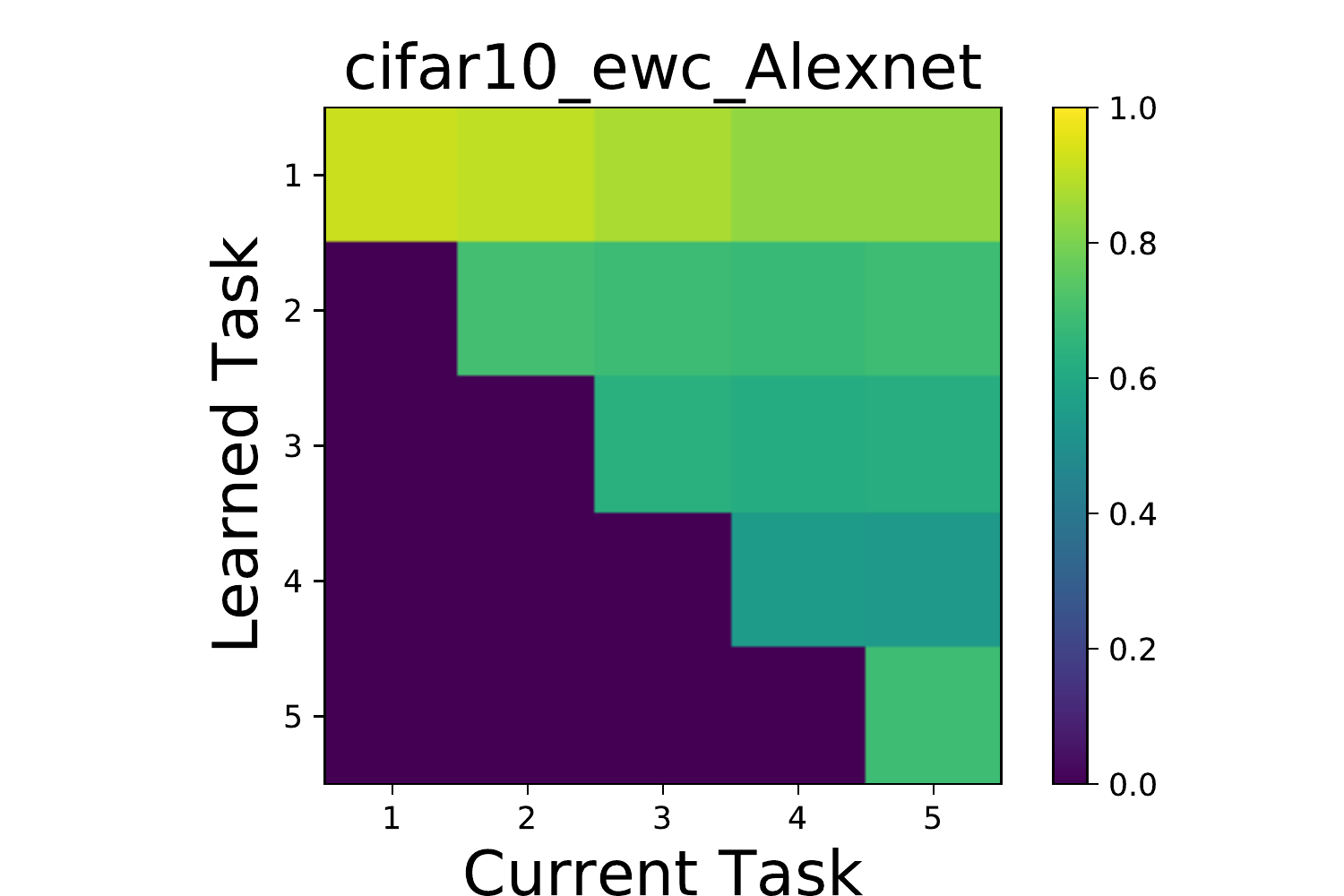}
    \end{minipage}%
    \begin{minipage}[]{0.2\linewidth}
    \centering
    \includegraphics[width=1\linewidth]{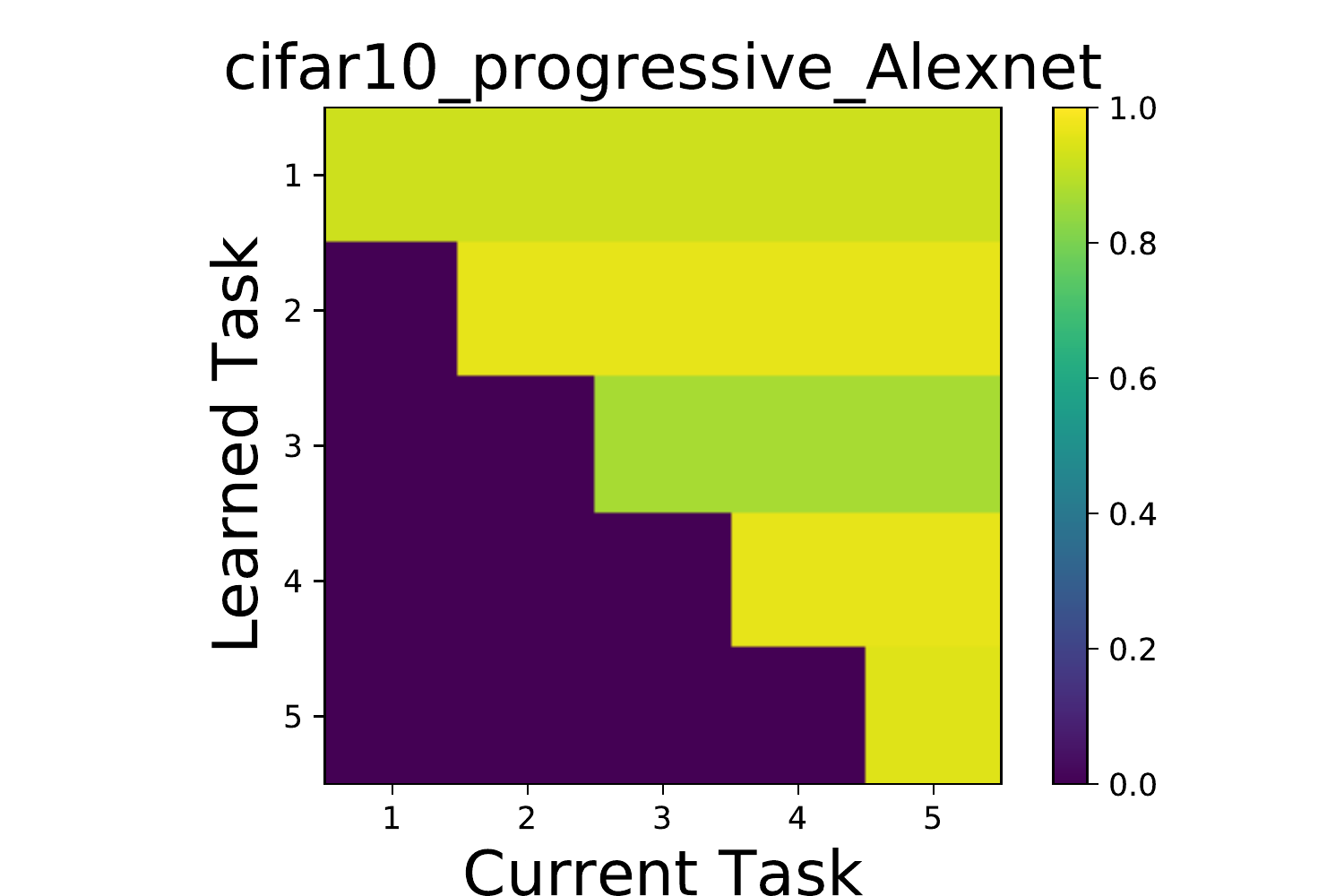}
    \end{minipage}
    \begin{minipage}[]{0.2\linewidth}
    \centering
    \includegraphics[width=1\linewidth]{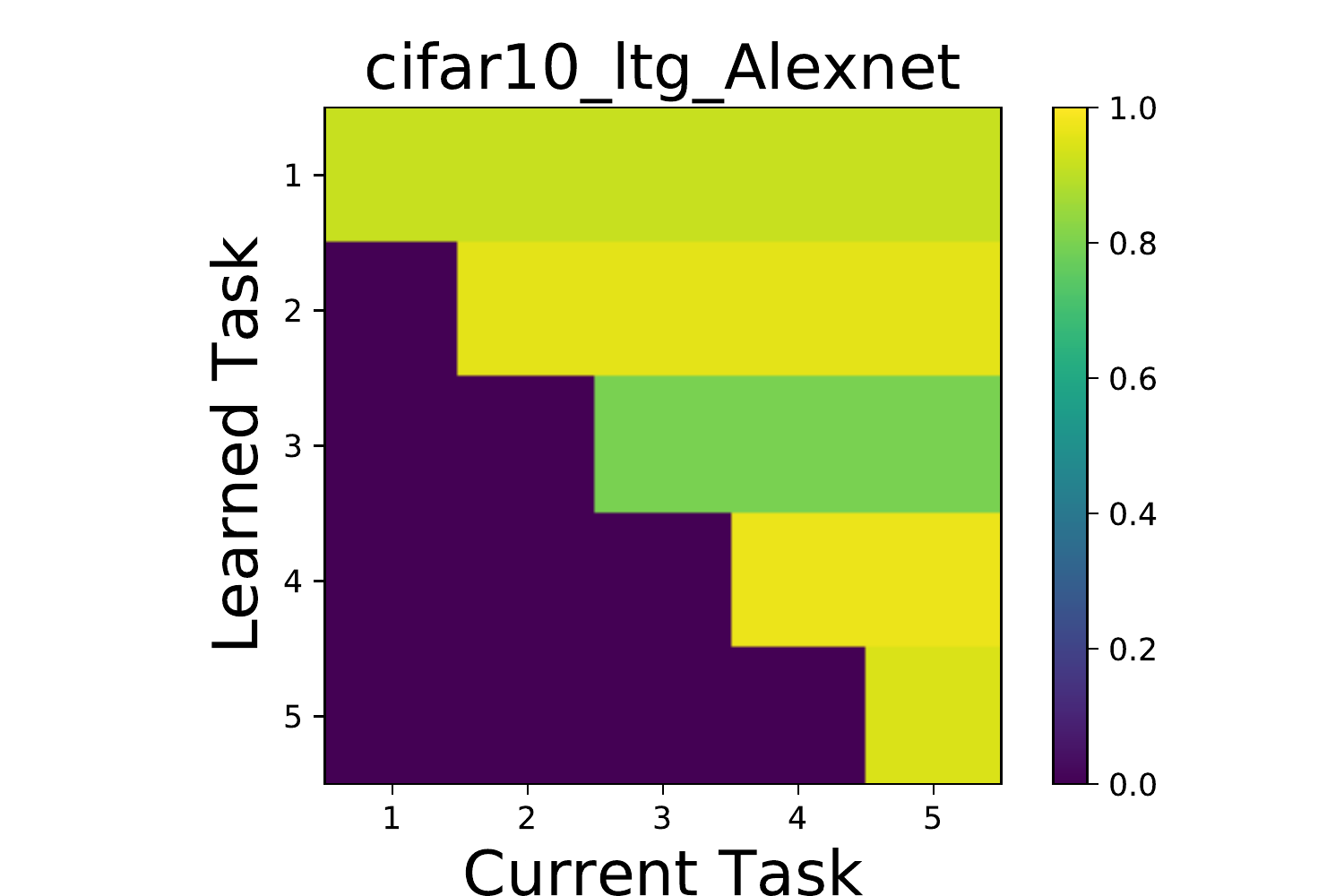}
    \end{minipage}
    \begin{minipage}[]{0.2\linewidth}
    \centering
    \includegraphics[width=1\linewidth]{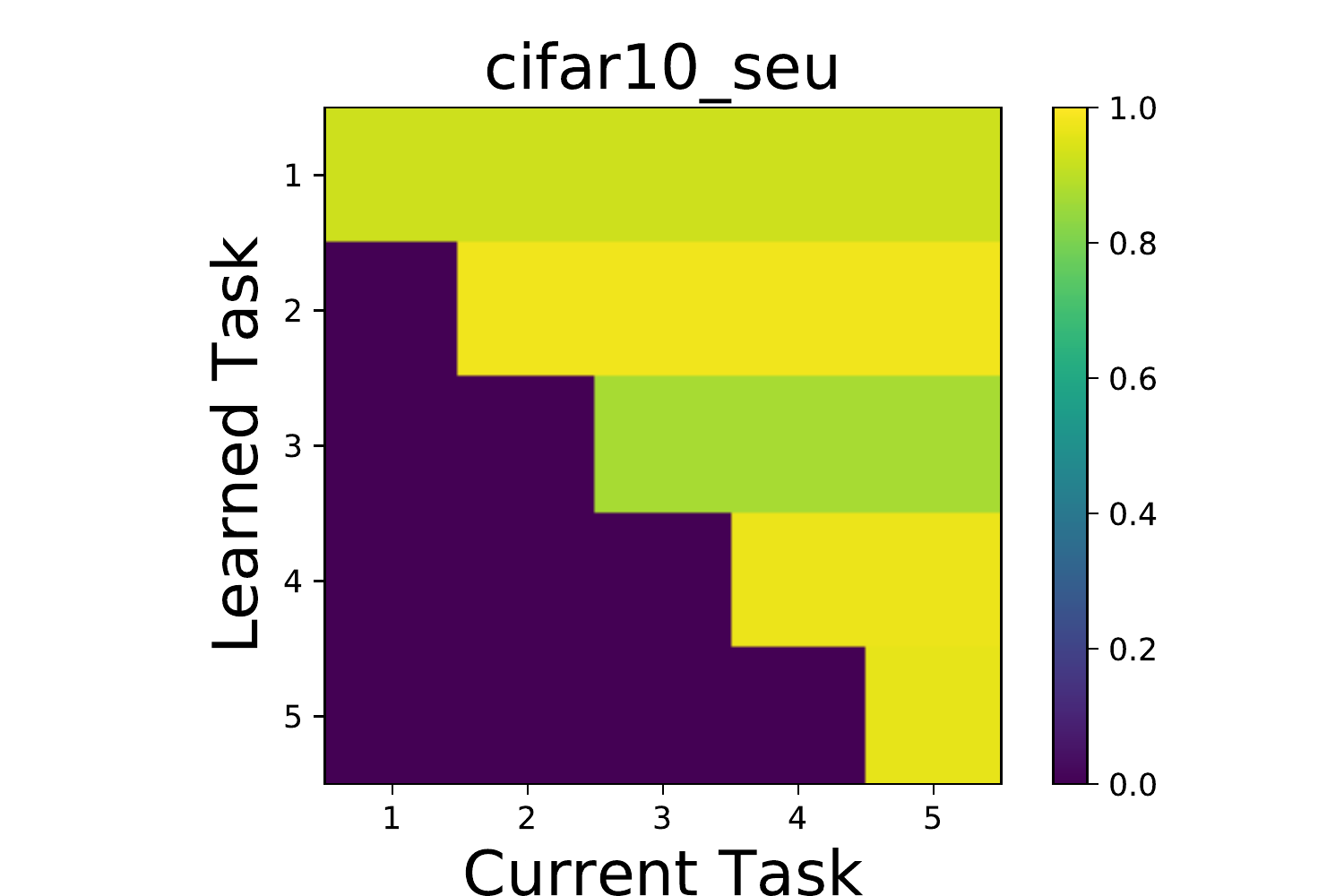}
    \end{minipage}
    \label{exp_acc_heat_cifar10}
    }

    \subfloat[]{
    \begin{minipage}[]{0.2\linewidth}
    \centering
    \includegraphics[width=1\linewidth]{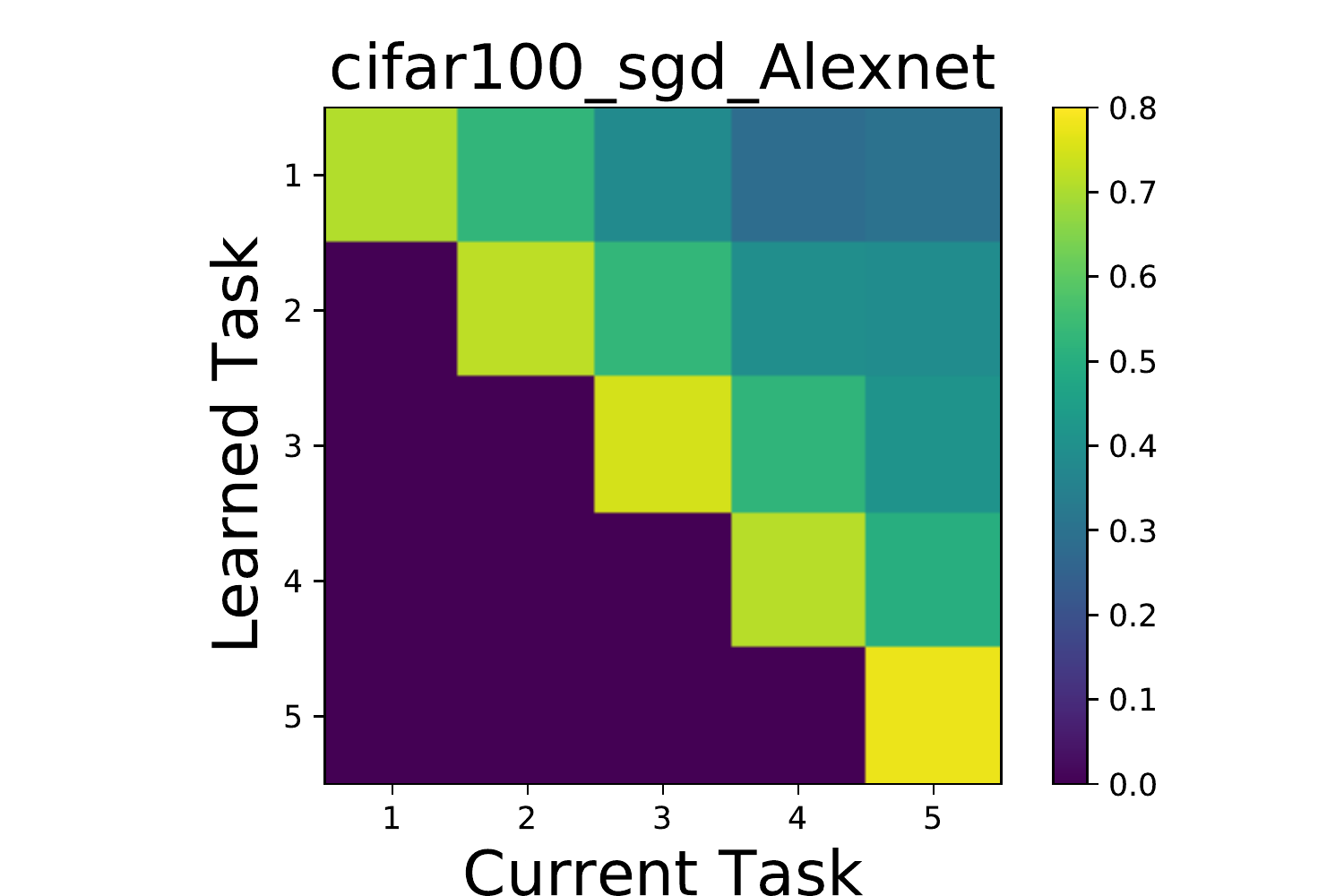}
    \end{minipage}
    \begin{minipage}[]{0.2\linewidth}
    \centering
    \includegraphics[width=1\linewidth]{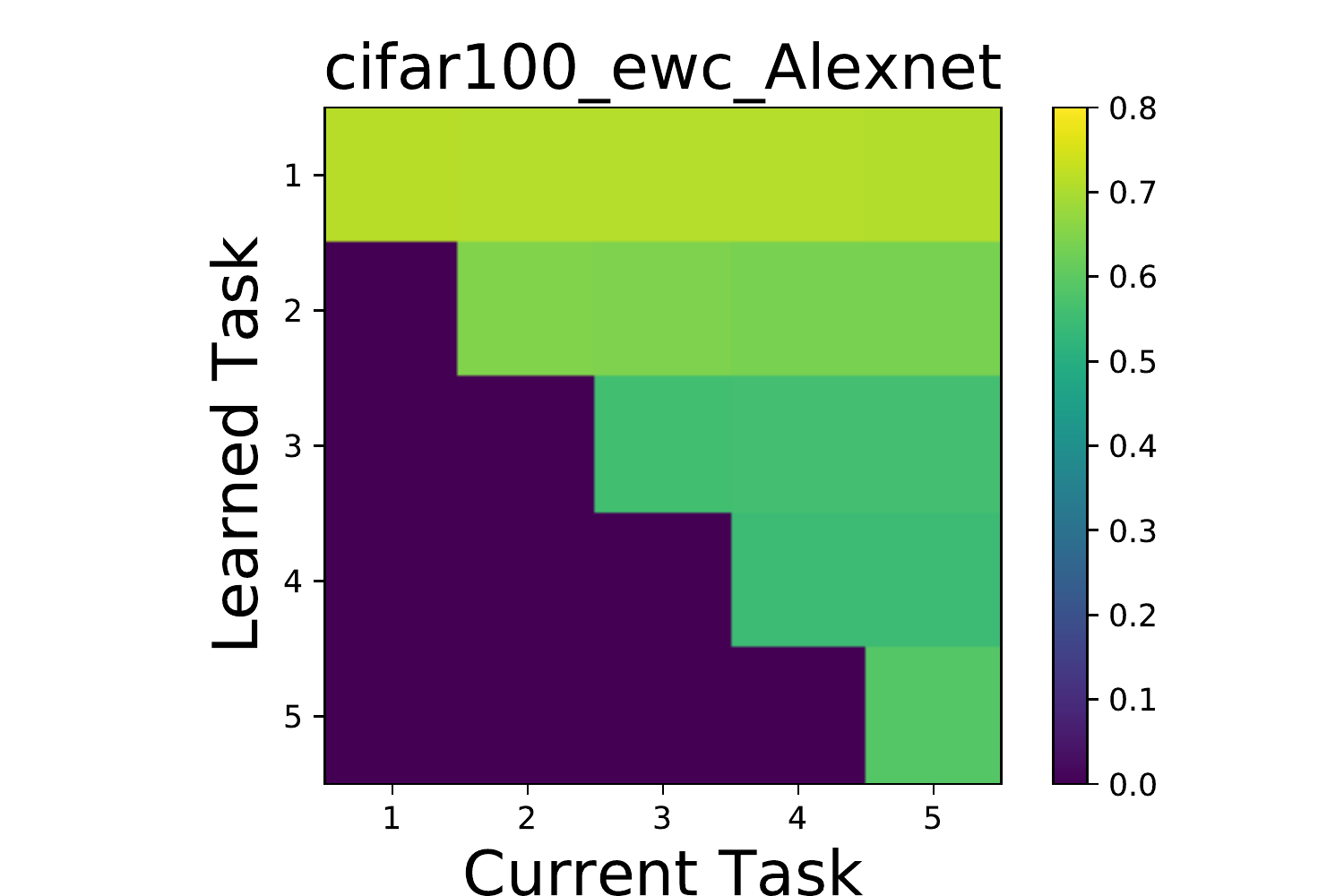}
    \end{minipage}%
    \begin{minipage}[]{0.2\linewidth}
    \centering
    \includegraphics[width=1\linewidth]{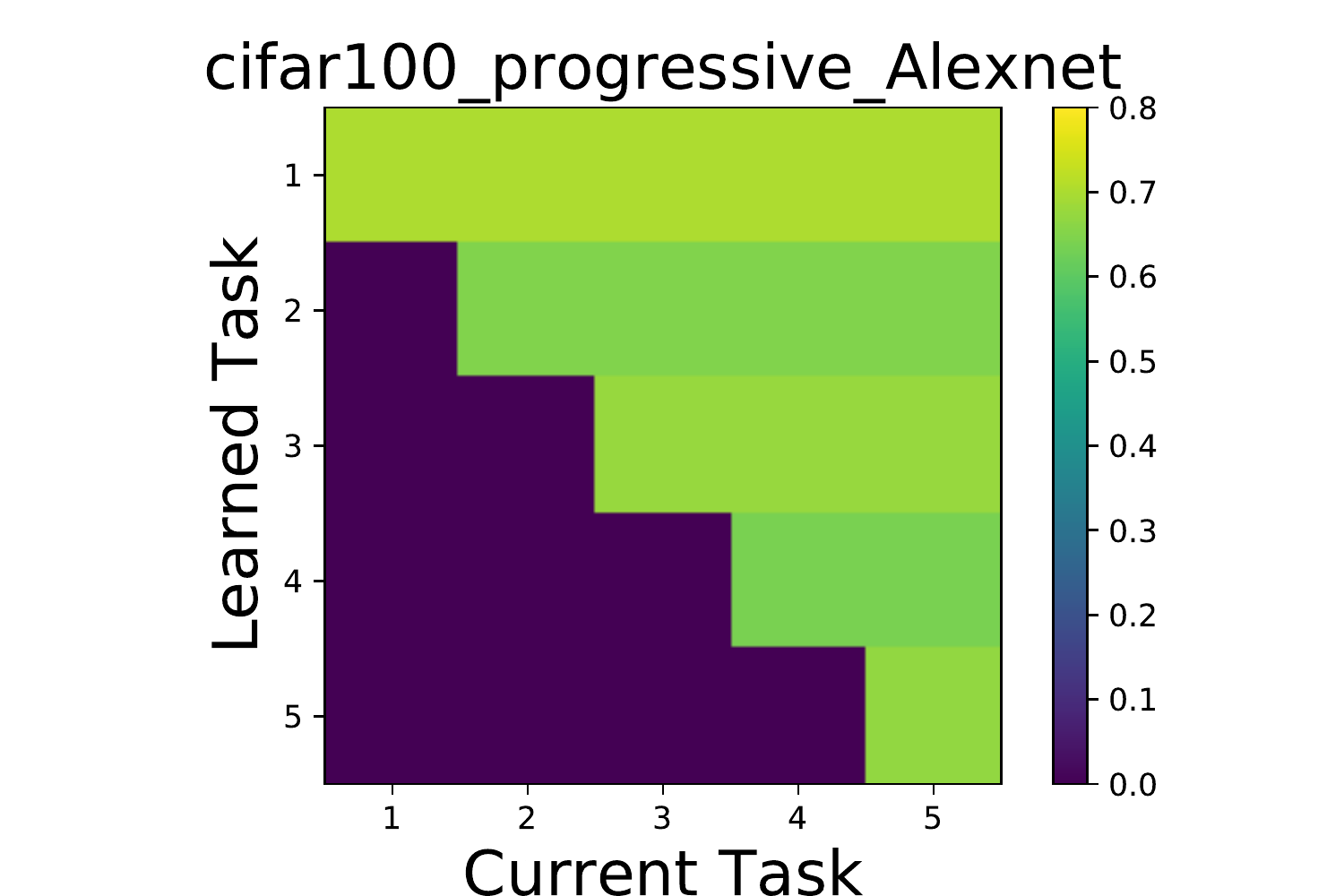}
    \end{minipage}
    \begin{minipage}[]{0.2\linewidth}
    \centering
    \includegraphics[width=1\linewidth]{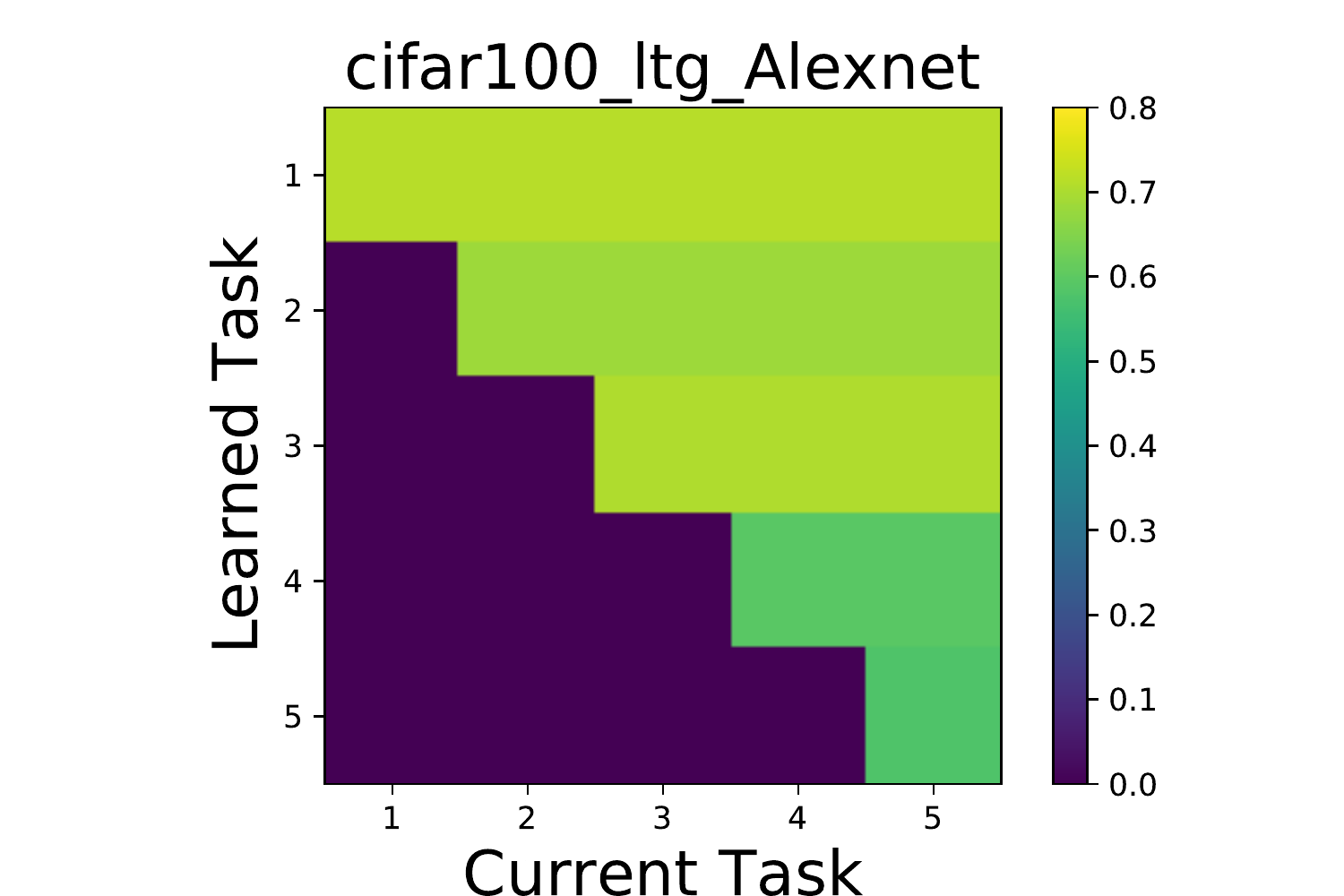}
    \end{minipage}
    \begin{minipage}[]{0.2\linewidth}
    \centering
    \includegraphics[width=1\linewidth]{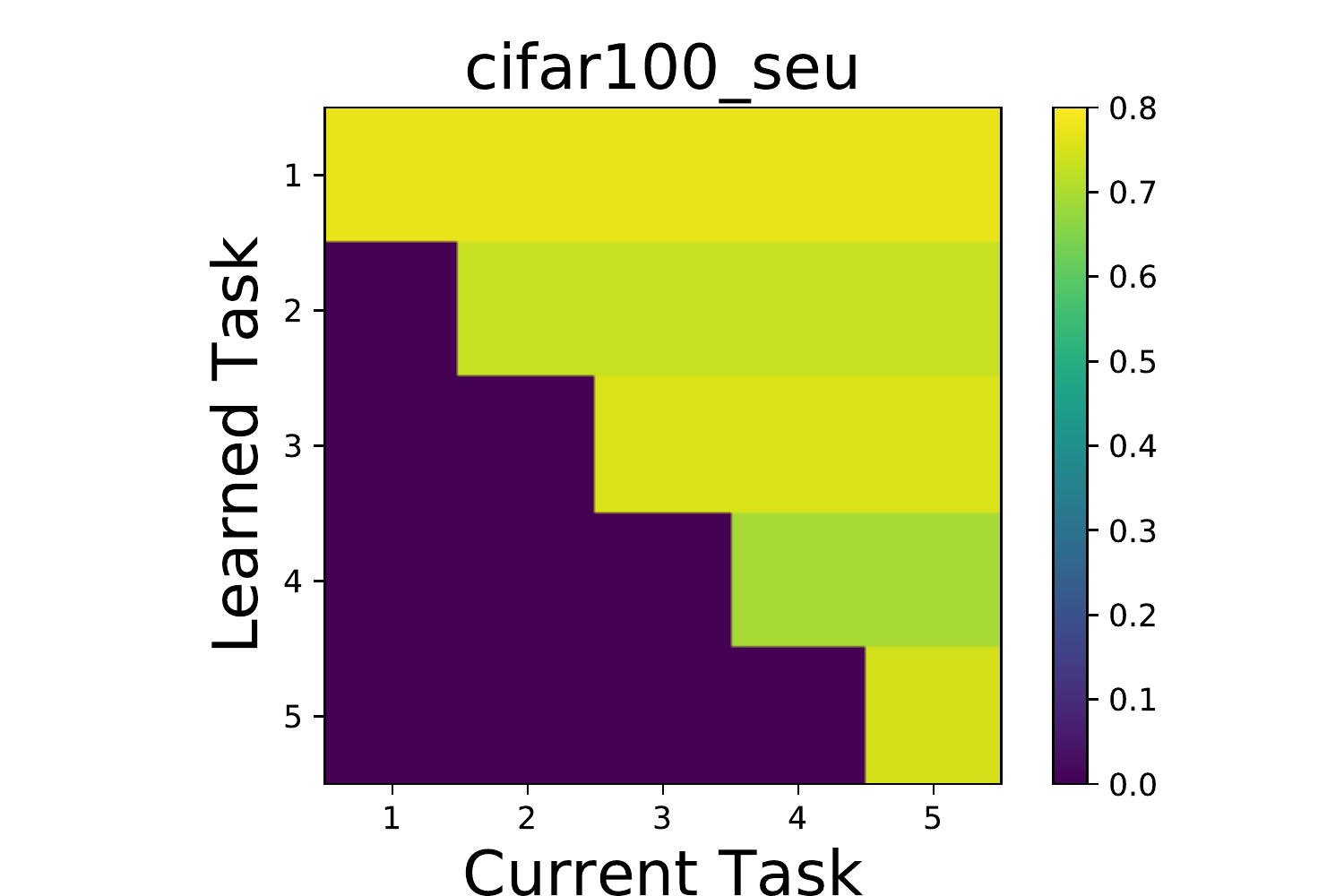}
    \end{minipage}
    \label{exp_acc_heat_cifar100}
    }

    \subfloat[]{
    \begin{minipage}[]{0.2\linewidth}
    \centering
    \includegraphics[width=1\linewidth]{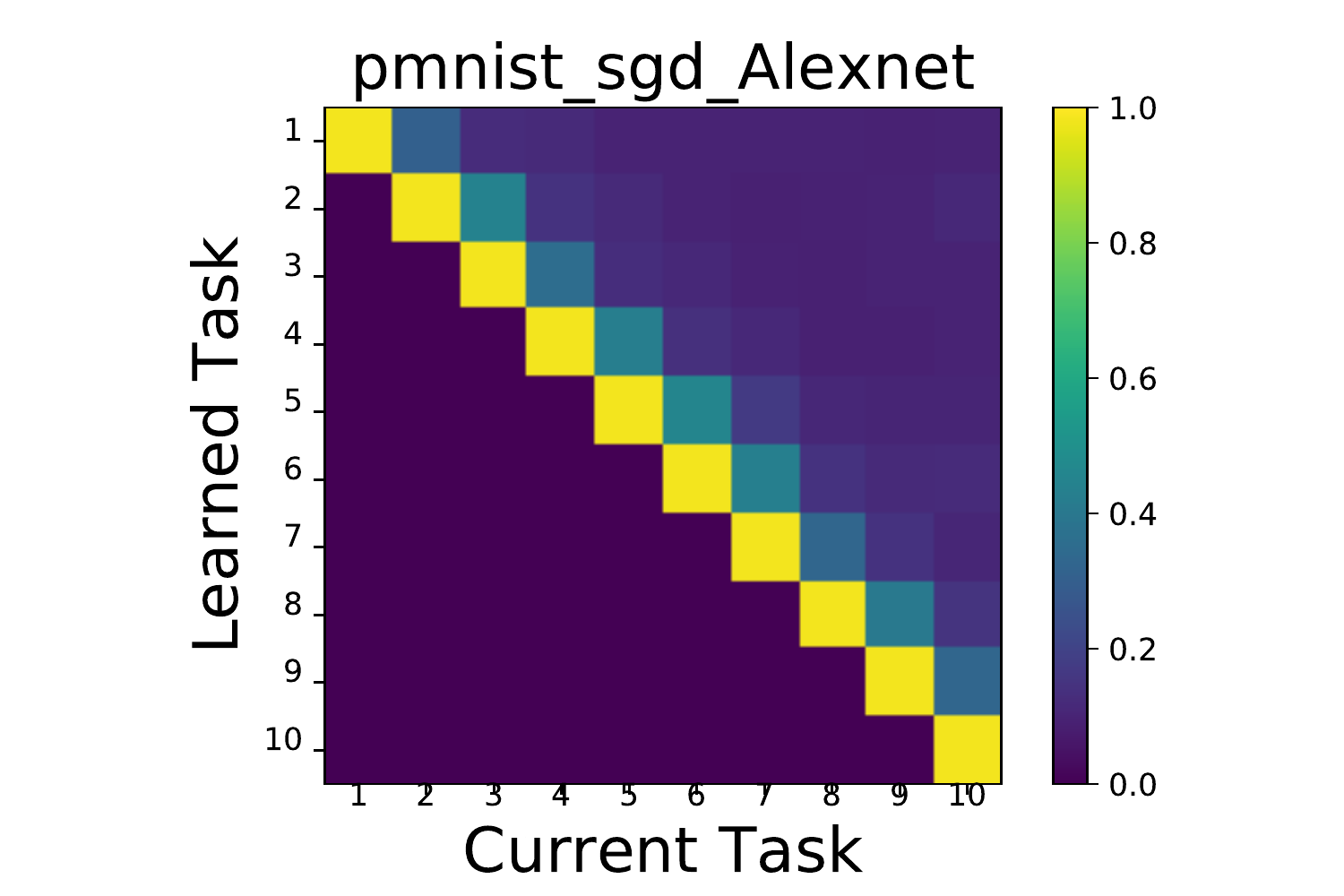}
    \end{minipage}
    \begin{minipage}[]{0.2\linewidth}
    \centering
    \includegraphics[width=1\linewidth]{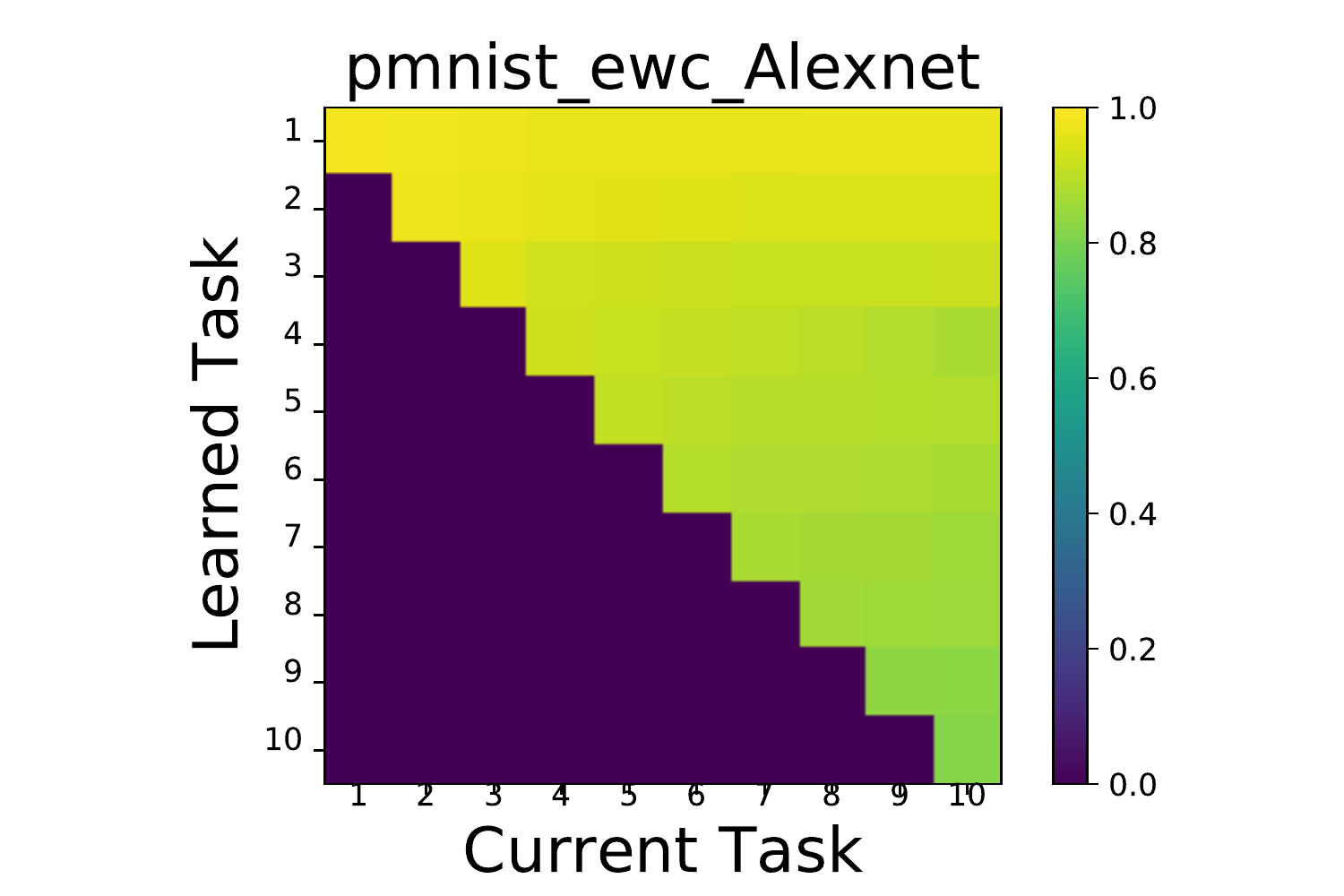}
    \end{minipage}%
    \begin{minipage}[]{0.2\linewidth}
    \centering
    \includegraphics[width=1\linewidth]{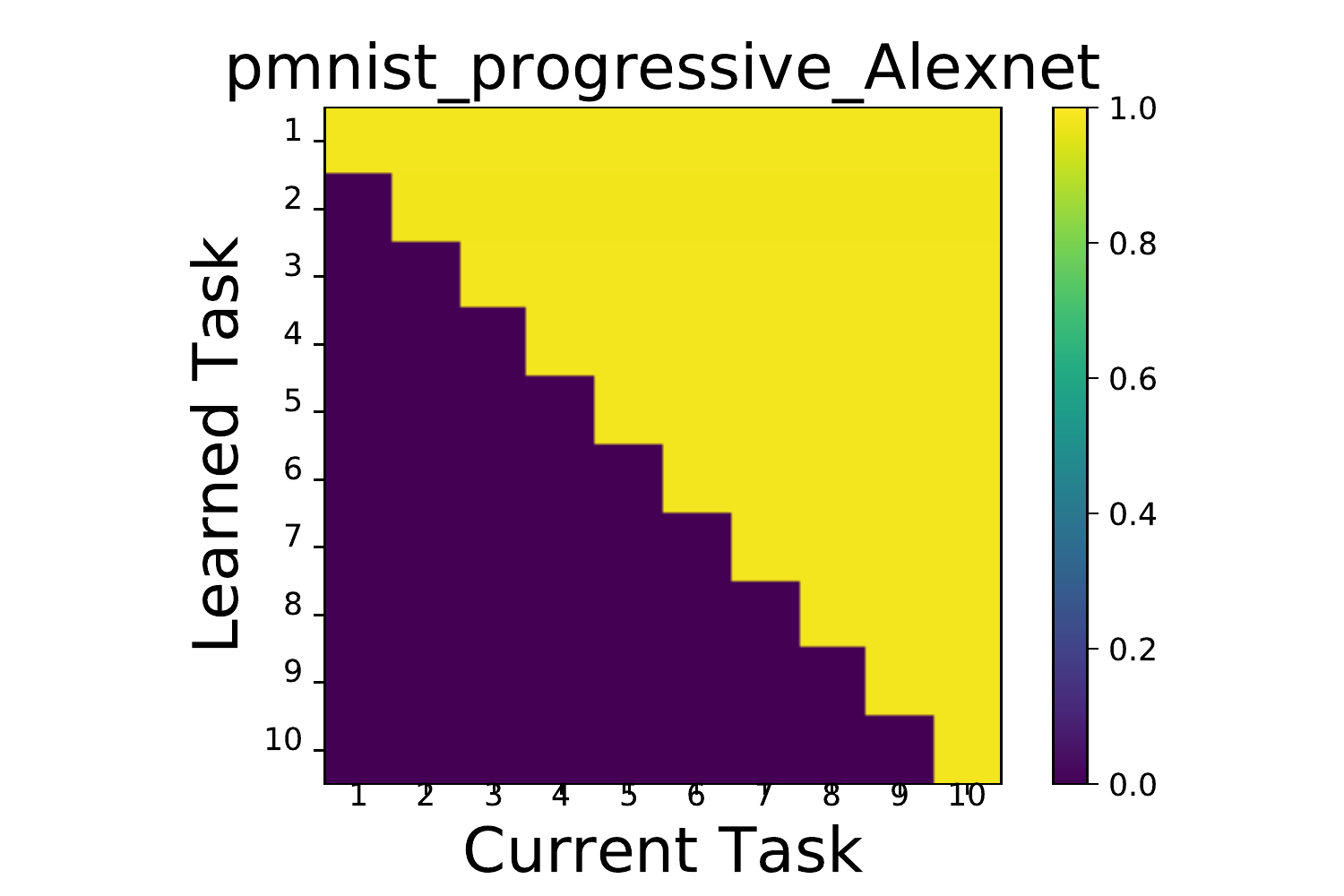}
    \end{minipage}
    \begin{minipage}[]{0.2\linewidth}
    \centering
    \includegraphics[width=1\linewidth]{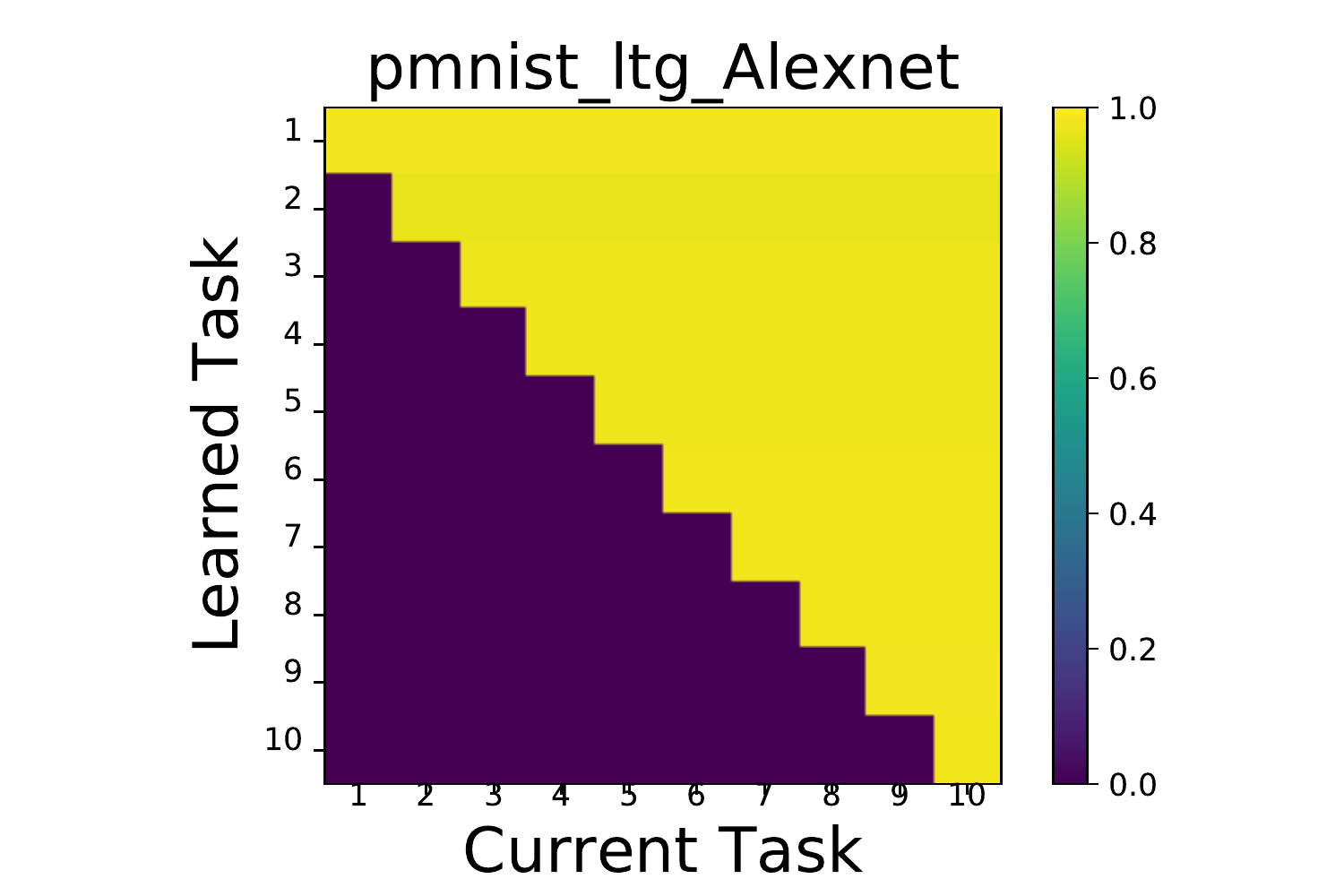}
    \end{minipage}
    \begin{minipage}[]{0.2\linewidth}
    \centering
    \includegraphics[width=1\linewidth]{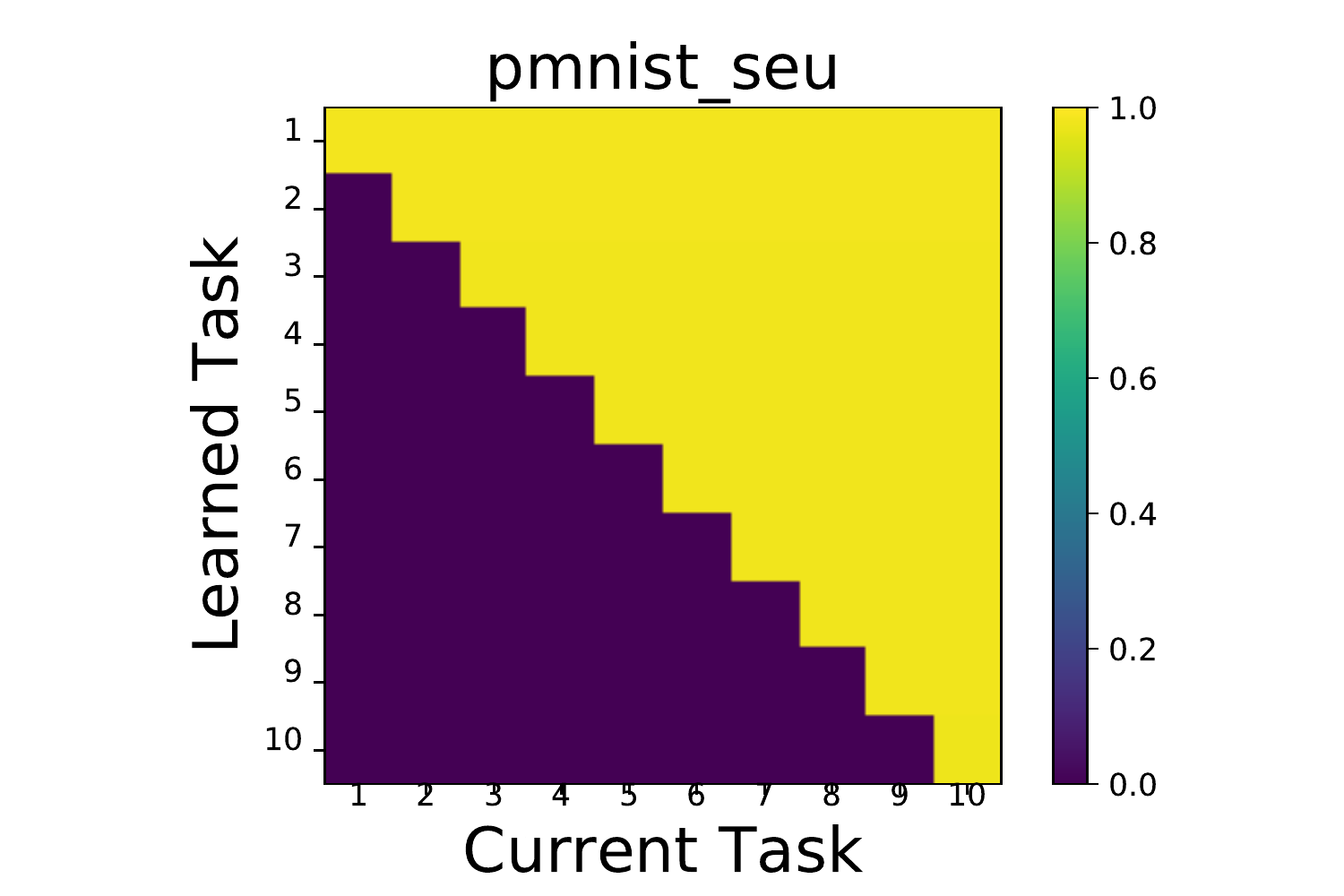}
    \end{minipage}
    \label{exp_acc_heat_pmnist}
    }

    \subfloat[]{
    \begin{minipage}[]{0.2\linewidth}
    \centering
    \includegraphics[width=1\linewidth]{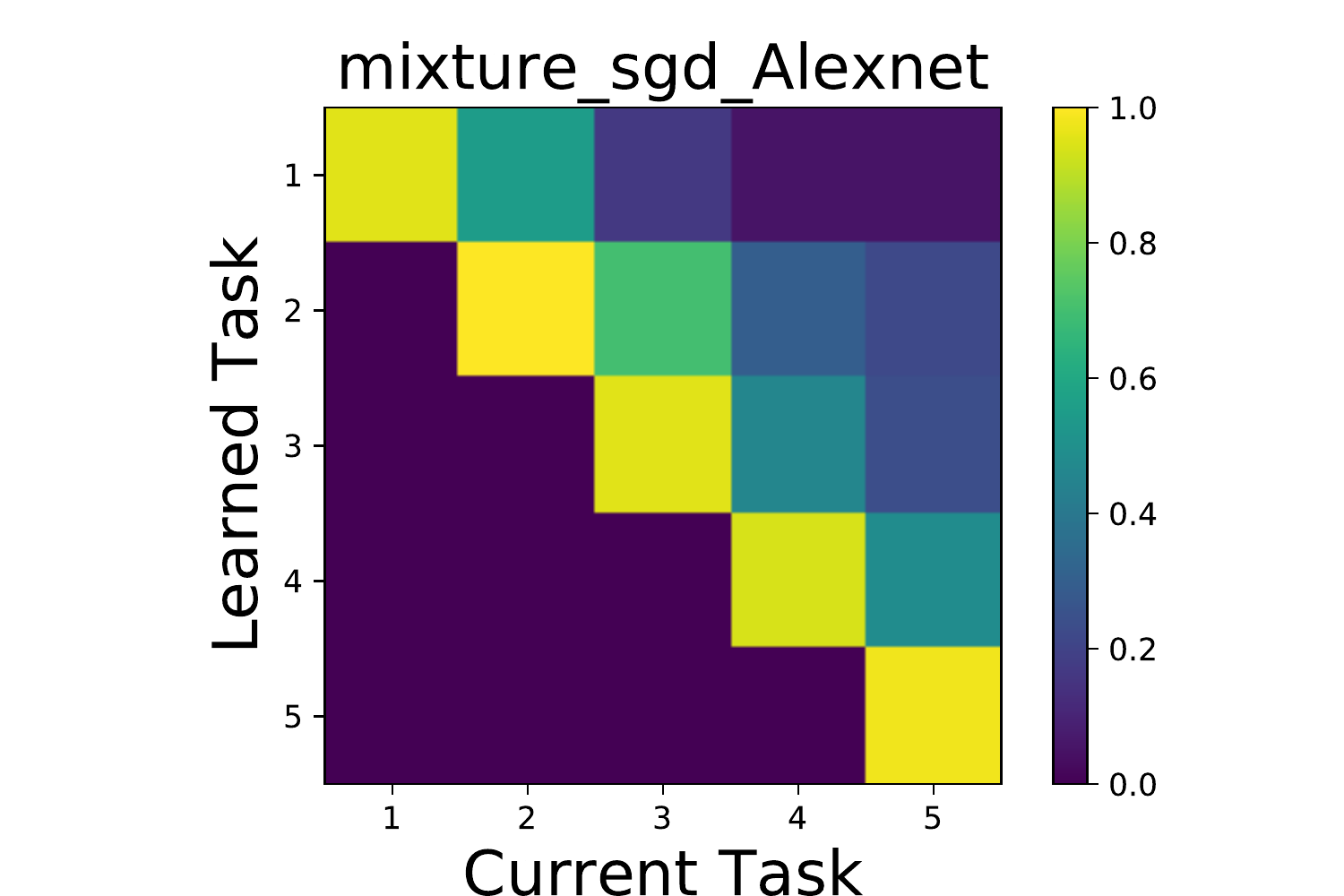}
    \end{minipage}
    \begin{minipage}[]{0.2\linewidth}
    \centering
    \includegraphics[width=1\linewidth]{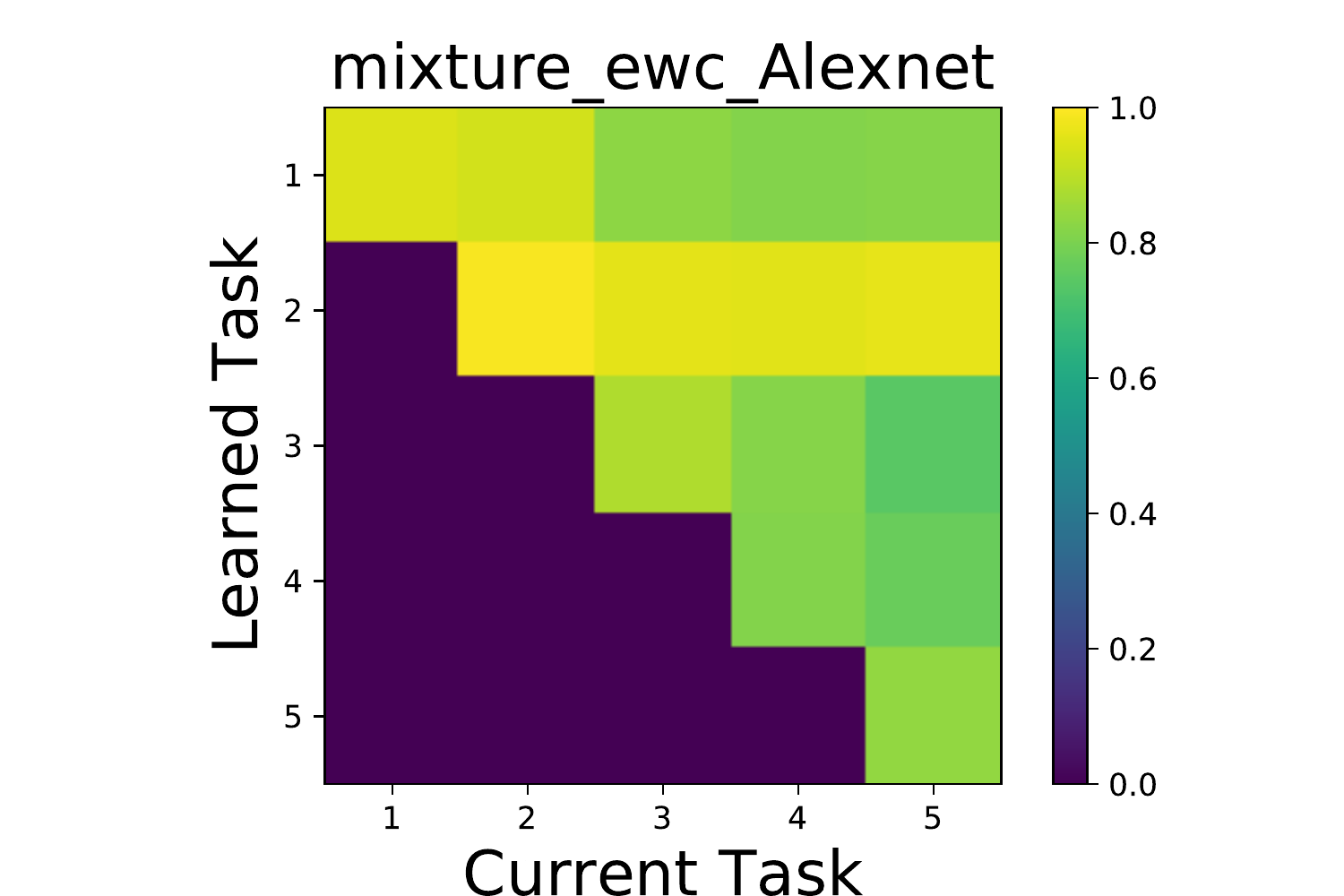}
    \end{minipage}%
    \begin{minipage}[]{0.2\linewidth}
    \centering
    \includegraphics[width=1\linewidth]{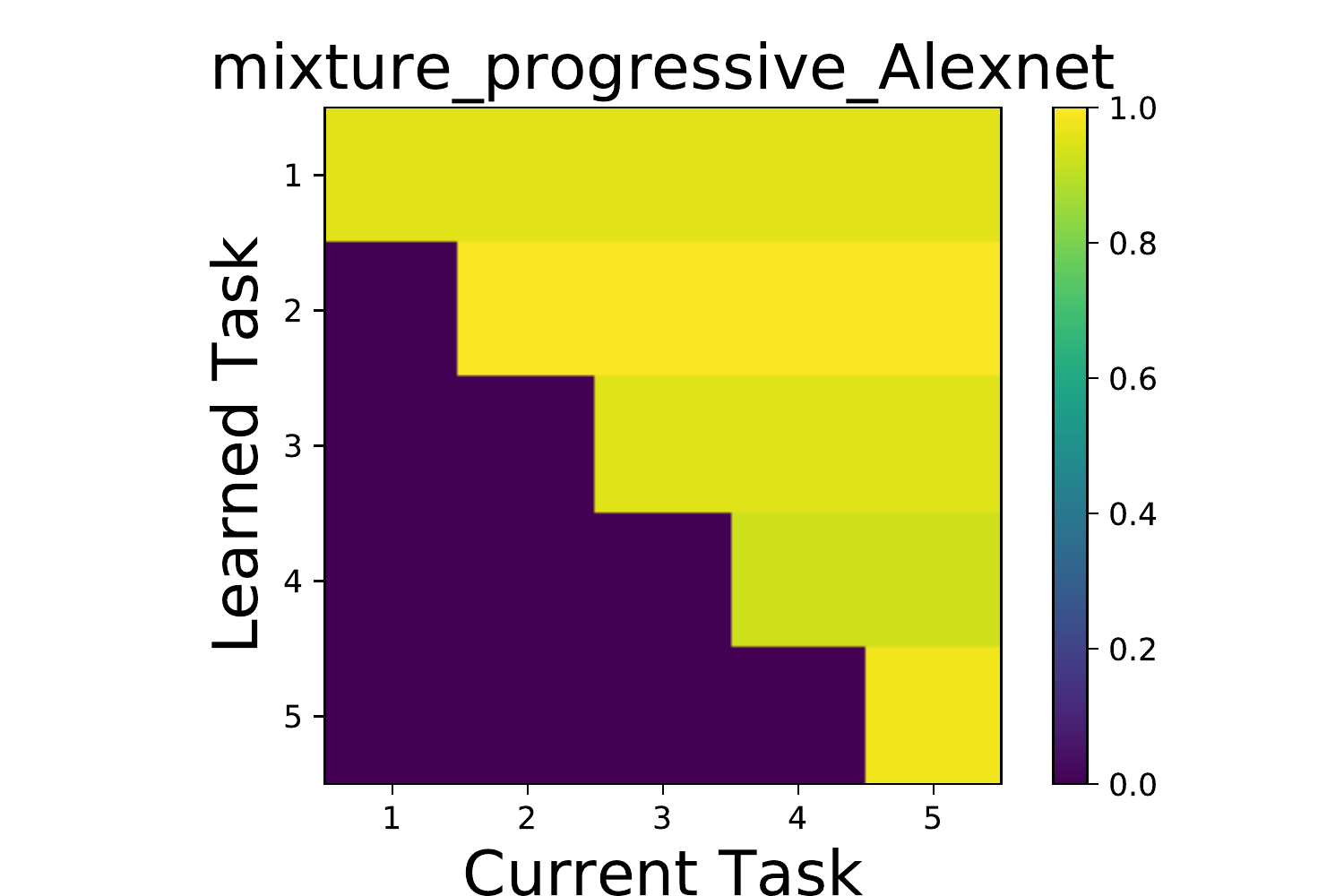}
    \end{minipage}
    \begin{minipage}[]{0.2\linewidth}
    \centering
    \includegraphics[width=1\linewidth]{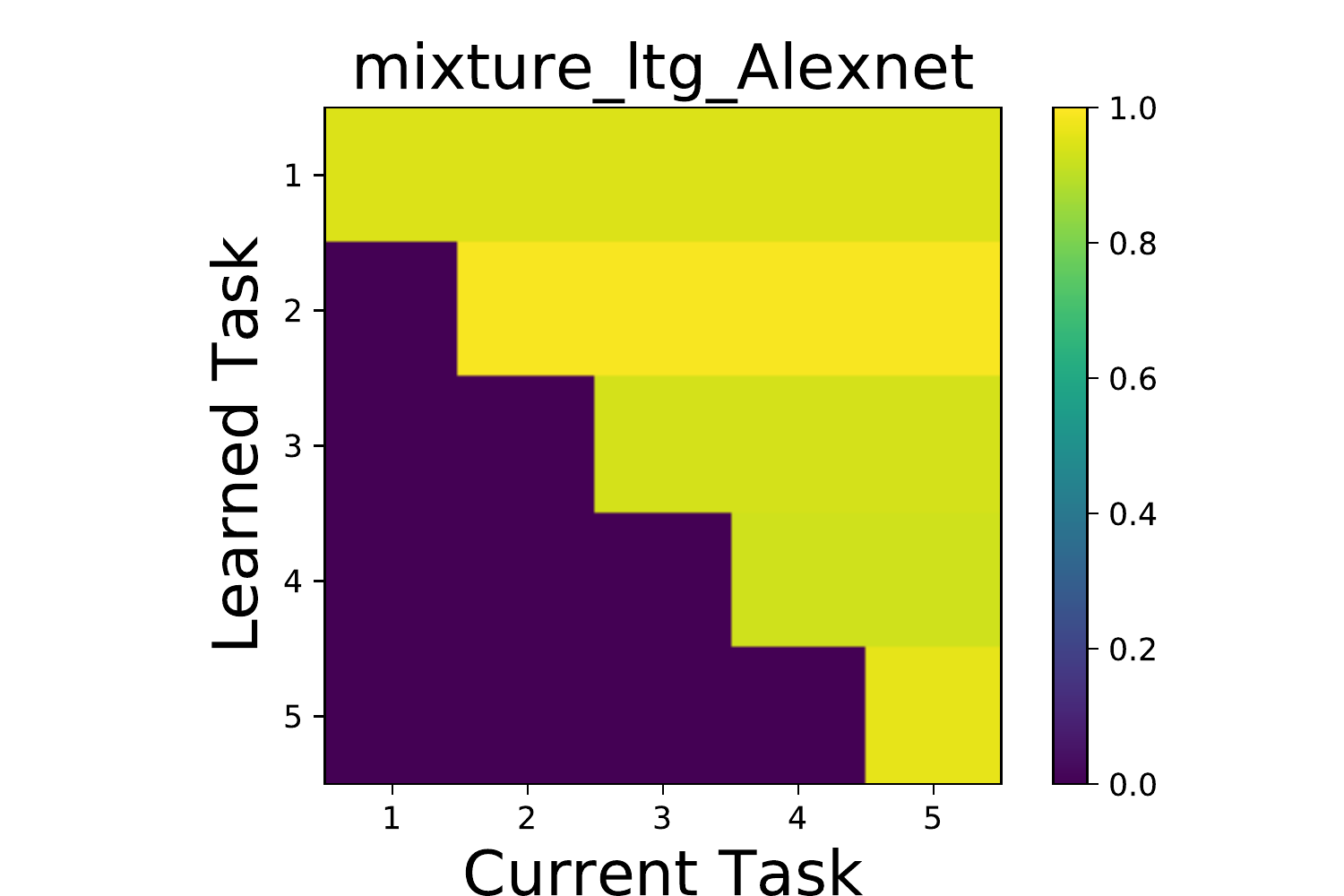}
    \end{minipage}
    \begin{minipage}[]{0.2\linewidth}
    \centering
    \includegraphics[width=1\linewidth]{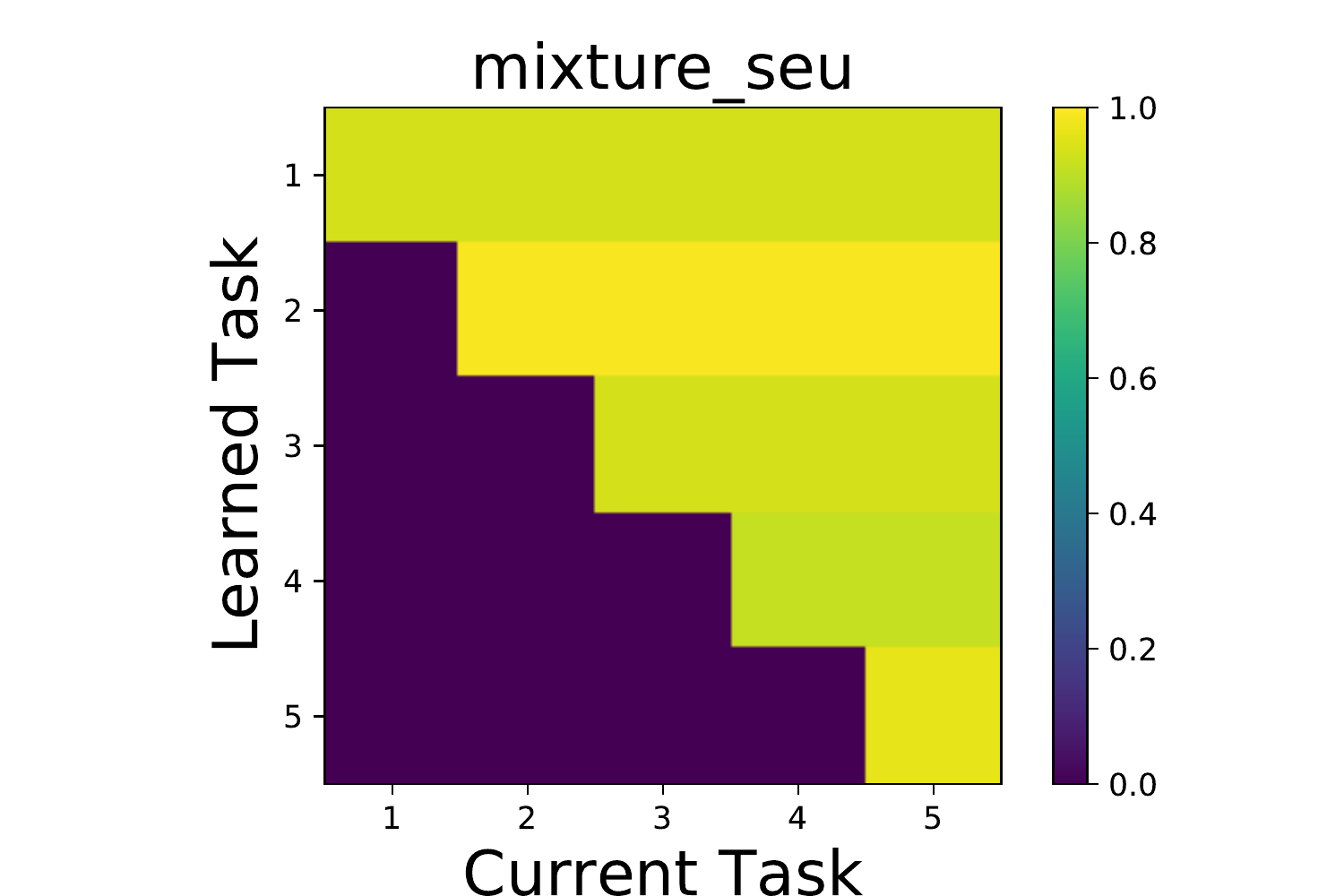}
    \end{minipage}
    \label{exp_acc_heat_mixture}
    }

    \centering
    \caption{Heat Map: The accuracy of each task learned after learning current task. \textbf{Note that only the upper triangle of the image makes sense.} (a) split CIFAR10 (b) split CIFAR100 (c) PMNIST (d) Mixture}
    \label{exp_acc_heat}
\end{figure*}

The purpose of lifelong learning is to learn a series of tasks and get a great performance on each task. So, after learning each task, we compare the average accuracy of all tasks that have been learned. We first compare them on the PMNIST (see \figurename{\ref{exp_avg_acc_pmnist}}). The average accuracy curve of SGD drops rapidly with the arrival of the new task and the average accuracy drops from 98.12\% to 22.09\%. Although the rate of decline slowed, the average accuracy of EWC declines as the new task is learned and drops from 98.16\% to 88.20\%. The curves of all three methods with a growth model are relatively flat. All average accuracies are greater than 97\% from start to finish. After learning all tasks, Progressive Network achieves the first performances (98.16\%). Our method  
achieves second place (97.91\%) with fewer parameters (see in Section \ref{sec_exp_acc_size}). The third place is taken by LTG (97.72\%). On the split CIFAR10 (see \figurename{\ref{exp_avg_acc_cifar10}}), our method wins the first place (94.13\%) and the second place is Progressive Network (93.32\%). The third place is still taken by LTG (91.86\%). On the split CIFAR100 (see \figurename{\ref{exp_avg_acc_cifar100}}), our method wins the first place (74.05\%). The second is Progressive (66.76\%) and the third is LTG (65.53\%). On the Mixture (see \figurename{\ref{exp_avg_acc_mixture}}), our method achieves third place (94.65\%).

In order to further analyze the overall performance of different methods, we show the accuracy of each task that learned after learning current task in the form of a heat map (see in \figurename{\ref{exp_acc_heat}}). Note that only the upper triangle of the image makes sense. The image of SGD is only brighter on the diagonal which means it performs well only on the task it has just learned, and poorly on previous tasks. The color of the image of EWC is more even than SGD which means it performs better on previous tasks. But obviously, the diagonal of the image of EWC is darker than SGD. Therefore, in order to guarantee the performance on previous tasks, EWC limits the ability of the model to learn new tasks. Images of methods with a growth model look better and the color of the image of our method is more even and brighter especially on the split CIFAR100 (see \figurename{\ref{exp_acc_heat_cifar100}}).

\subsection{Accuracy vs Number of Parameters}
\label{sec_exp_acc_size}
\begin{figure}[h]
\centering
\includegraphics[width=1\linewidth]{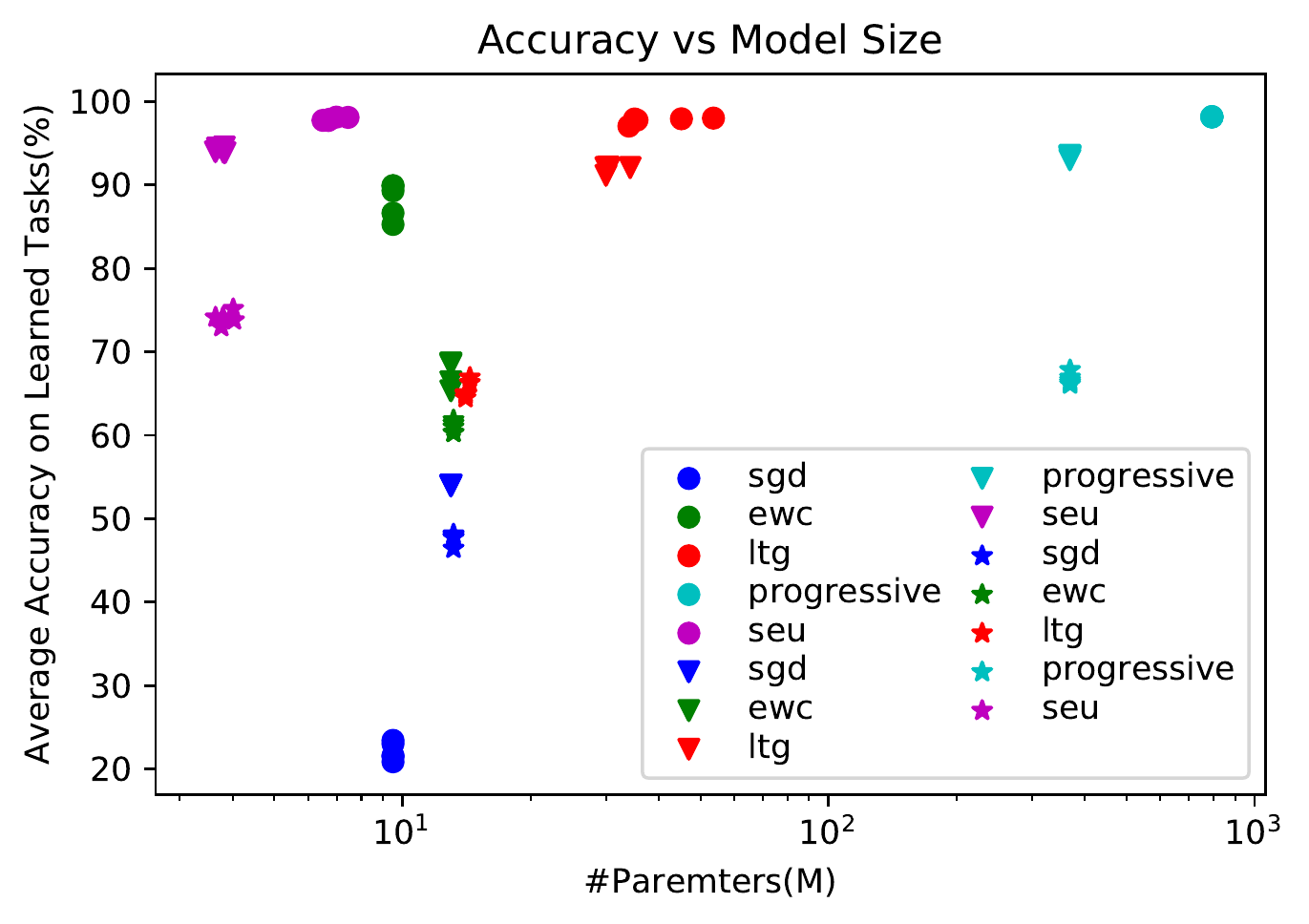}
\caption{Accuracy vs Number of Parameters. A circle in figure indicates that the experiment is conducted on the PMNIST. The triangle corresponds to the split CIFAR10 and the pentagram corresponds to the split CIAFR100. The higher the top left corner, the higher the model utilization.}
\label{exp_acc_size}
\end{figure}

In lifelong learning, the accuracy of and the number of parameters of the model can better reflect whether the model is fully utilized. We show the relationship between the accuracy and the number of parameters of models obtained by different methods on the PMNIST, the split CIAFR10, and the CIAFR100 (see \figurename{\ref{exp_acc_size}}). For quantitative comparsion purposes, a new metric $f(a, n)$ considers both accuracy and number of parameters is needed, where $a$ is the average accuracy($\%$) for all learned tasks and $n$ is the number of parameters of the model. And it should obey the following properties:
\begin{itemize}
    \item $0\le f(a,n)\le 1,f(0,n)=0,f(a,+\infty)\rightarrow 0$.
    \item For any $n$, $f(a,n)$ is a monotonically increasing function of $a$.
    \item For any $a$, $f(a,n)$ is a monotonically decreasing function of $n$.
    \item $f(a,n)=1\Leftrightarrow a=100, n=0$.
\end{itemize}
We denote a mixture score of accuracy and number of parameters (MS) as follows and our method achieves the highest scores on all four datasets (see in Table \ref{table_mixed_score}).
\begin{equation}
    \text{MS}(a,n)=\frac{\sqrt{a}}{10}\frac{\cos[\pi(1-e^{-\lg(n+1)/10})]+1}{2}
\end{equation} 

In general, methods with a growth model have a larger number of parameters. However, since both EU and Task Model for each task are specifically searched, the model of our method is even more compact than methods with a fixed model. Although the Learn to Grow also tries to search for a better expansion operation, its expansion module always has a fixed architecture that is not compact enough. So, this experimental result proves empirically that searching for a suitable EU and Task Model for each new task is useful for a more compact model. Our method can achieve higher accuracy with fewer model parameters and achieve the highest mixed score.

\begin{table*}[t]
    \caption{Mixed Score}
    \label{table_mixed_score}
    \centering
    \begin{tabular}{|c|lll|lll|lll|lll|}
        \hline
        &\multicolumn{3}{|c|}{split CIFAR10} 
        &\multicolumn{3}{|c|}{split CIFAR100}
        &\multicolumn{3}{|c|}{PMNIST}
        &\multicolumn{3}{|c|}{Mixture}\\
        \hline
        \bfseries Method &\bfseries Acc(\%) &\bfseries Num(M) &\bfseries MS &\bfseries Acc(\%) &\bfseries Num(M) &\bfseries MS &\bfseries Acc(\%) &\bfseries Num(M) &\bfseries MS &\bfseries Acc(\%) &\bfseries Num(M) &\bfseries MS\\
        \hline
        \bfseries SGD
        &54.00&12.98 &0.357  
        &47.62&13.16 &0.335 
        &22.09&9.49 &0.233 
        &39.54&13.10 &0.305 \\
        \bfseries EWC
        &67.49 &12.98 &0.399
        &60.98 &13.16 &0.379
        &88.20 &9.49 &0.466
        &82.61 &13.10 &0.441 \\
        \bfseries LTG
        &91.86 &31.00 &0.438
        &65.53 &14.18 &0.391
        &97.73 &40.66 &0.444
        &95.33 &60.00 &0.426 \\
        \bfseries PN 
        &93.32 &368.55 &0.370
        &66.76 &368.88 &0.313
        &\bfseries 98.16 &794.18 &0.358
        &\bfseries 96.12 &368.73 &0.375 \\
        \bfseries SEU 
        &\bfseries 94.13 &\bfseries 3.75 &\bfseries 0.513
        &\bfseries 74.05 &\bfseries 3.84 &\bfseries 0.454
        &97.91 &\bfseries 6.87 &\bfseries 0.502
        &94.65 &\bfseries 3.44 &\bfseries 0.517 \\
        \hline
    \end{tabular}
\end{table*}

\section{Related Work}
\label{sec_related}
Lifelong learning is remaining a challenging task in the field of artificial intelligence\cite{lifelong,parisi_continual_2019}. It focuses on the ability of intelligent agents to learn consecutive tasks without forgetting the knowledge learned in the past. A major problem with lifelong learning is catastrophic forgetting, which is still a big problem in deep learning \cite{french_catastrophic_1999}. Catastrophic forgetting occurs when a model that has been trained on some other tasks is trained on a new task. The weights which are important for previous tasks are changed to adapt to the new task. Up to now, many new methods have emerged to solve this problem.

One way to solve the problem directly is to use a memory system to alleviate forgetting, inspired by the complementary learning systems (CLS). The memory system stores previous data and replay these old samples when learning the new task\cite{rebuffi_icarl:_2017, lopez-paz_gradient_2017}. One obvious drawback to these methods is the need to store old information. So, \cite{deep_generative_replay} proposed a dual-model architecture consisting of a task solver and a generative model. In place of storing real samples of previous tasks, the memory system only needs to retrain the generative model after learning a new task. 

In addition to retaining information about previous tasks by storing old examples, an agent can also try to store the knowledge it has learned from previous tasks. For neural networks, whose knowledge is stored in model weights, an intuitive approach to retain the knowledge of previous tasks is adding a quadratic penalty on the difference between the weights for previous tasks and the new task\cite{ewc, Synaptic_Intelligence,memory_aware_synapses}. By modifying the penalty with fish information matrix, \cite{ewc} proposed the Elastic weight consolidation (EWC) model and further considered the importance of different weights for previous tasks. \cite{Synaptic_Intelligence} estimated the importance of a parameter by the sensitivity of the loss to a parameter change. \cite{memory_aware_synapses} proposed the Memory Aware Synapses method and estimated the importance of a parameter by the sensitivity of the learned function to a parameter change.

Another way to keep knowledge of previous tasks is by selecting different sub-network from a single fixed network for different tasks. Concretely speaking, when learning a new task, the sub-networks of previous tasks are protected and the network selects a sub-network for the new task from the parts that have never been used. The PathNet, proposed by \cite{pathnet:_2017}, selects a path for the current new task by a tournament selection genetic algorithm. When learning the new task, the weights in paths of previous tasks are frozen so that the knowledge can be preserved. \cite{hard_attention} brought up a concept named hard attention which implies the importance of weights for tasks. Then, the paths for tasks are selected by hard attention. When learning a new task, the weights whose hard attention to previous tasks is large will be protected to hold the knowledge. \cite{adaption_2019} considered splitting the parameters of the model into two parts. One of the two is shared by all tasks and the other is adapted to different samples. 

As the number of tasks increases, a fixed network with a limited capacity will get into trouble. \cite{progressive_2016} proposed Progressive Network to assign additional fixed model resources to each task. When learning a new task, all of the weights of the model will be frozen and a list of new model modules will be trained. The new task can also reuse the previous knowledge by lateral connections between new and old modules. To avoid wasting resources, \cite{DEN_2018} proposed Dynamically Expandable Networks (DEN) and this method can remove unnecessary new modules by sparse regularization. Moreover, \cite{RCL_2018} assigned different new modules for different tasks by a controller trained by a reinforcement learning algorithm.

Except for the model weights, \cite{ltg} attempted to take the model architecture into account and proposed the Learn to Grow algorithm. At each layer of a model, this approach allows the new task to reuse the weights of previous tasks or retrain the module. Furthermore, it allows the new task to get a new module by adapting the used module. However, this method should select a original model at first which is suitable for all possible tasks. But this is hard to guarantee in practice and such a model is always large. Moreover, such a predefined structure ignores the fact that different tasks might fit different model structures or even different micro-architectures.

Our method Searchable Extension Units (SEU), as one of the methods with dynamic expansion, breaks the limitation of a predefined model and can search for specific extension units for different tasks.




\section{Conclusion}
\label{sec_conclusion}
In this paper, we propose a new framework for lifelong learning named Searchable Extension Units. Our method can search for suitable EU and Task Model for different tasks and break the limitation of a predefined original model. So the model in our method is more compact than methods with a predefined original model and fixed architectures of new resources. We validate our method on 4 datasets and the experimental results empirically prove that our method can achieve higher accuracy with fewer model parameters than other methods. In the future, it will make sense to find more efficient tools for Extension Unit Searcher and Task Solver Creator in SEU.


%



\section*{Acknowledgment}

The authors would like to thank the anonymous reviewers for their insightful comments on this paper. 

\ifCLASSOPTIONcaptionsoff
  \newpage
\fi


%
\bibliographystyle{IEEEtran}
\bibliography{IEEEabrv, ref}

\end{document}